%% file: main.tex
\documentclass[runningheads]{llncs}

\usepackage{eccv}

\input{sec/configs.tex}

\begin{document}

\title{Is user feedback always informative? \\Retrieval Latent Defending for Semi-Supervised Domain Adaptation without Source Data} 

\titlerunning{NBF \& RLD}

\input{sec/authors.tex}

\maketitle

\begin{abstract}
  \input{sec/0abstract}
\end{abstract}

\vspace{-1.5em}
\section{Introduction} \label{sec:intro}
\vspace{-.5em}
\input{sec/1introduction}

\vspace{-2.em}
\section{Related Work} \label{sec:relatedwork}
\vspace{-.5em}
\input{sec/2relatedwork}

\vspace{-.5em}
\section{Negatively Biased Feedback} \label{sec:setup}
\vspace{-.3em}
\input{sec/3problem_definition}

\vspace{-1.em}
\section{Approach} \label{sec:balancing}
\vspace{-.5em}
\input{sec/4approach}

\vspace{-1.em}
\section{Experiments} \label{sec:results}
\input{sec/5experimental_setups}

\input{sec/6main_results}
\input{sec/7ablations}

\vspace{-1.em}
\section{Conclusion \& Discussion} \label{sec:conclusion}
\vspace{-.5em}
\input{sec/8conclusion}

\bibliographystyle{splncs04}
\bibliography{egbib}

\clearpage
\input{sec/appendix}

\end{document}

%% file: sec/configs.tex
\usepackage{eccvabbrv}
\usepackage[accsupp]{axessibility}  % Improves PDF readability for those with disabilities.

\usepackage[pagebackref,breaklinks,colorlinks,citecolor=eccvblue]{hyperref}
\usepackage{orcidlink}

\usepackage{graphicx}
\usepackage{amsmath}
\usepackage{amssymb}
\usepackage{pifont}
\usepackage{booktabs}
\usepackage{array}
\usepackage{graphicx, amsmath, multirow, overpic, textpos} % caption, subcaption
\usepackage{colortbl}
\usepackage[linesnumbered,ruled,vlined]{algorithm2e}
\SetKwInOut{Parameter}{Parameters}
\usepackage{listings}
\usepackage{tikz}
\usepackage{bm}
\usepackage{arydshln}
\usepackage{hhline}
\usepackage{enumitem}

\DeclareMathOperator*{\argmax}{argmax}

\definecolor{cvprblue}{rgb}{0.21,0.49,0.74}
\definecolor{darkred}{rgb}{0.8,0.02,0.02}
\definecolor{darkredd}{rgb}{0.7,0.01,0.01}
\definecolor{darkblue}{rgb}{0.12,0.25,0.4}
\definecolor{blackpink}{rgb}{0.6,0,0.6}
\definecolor{highlight}{HTML}{FFF0C3}
\hypersetup{
  colorlinks   = true
}
\newcommand{\cmark}{\ding{51}}%
\newcommand{\xmark}{\ding{55}}%

\newlength\savewidth
\newcommand\shline{\noalign{\global\savewidth\arrayrulewidth
  \global\arrayrulewidth 1.25pt}\hline\noalign{\global\arrayrulewidth\savewidth}}
\newcommand\shlinesmall{\noalign{\global\savewidth\arrayrulewidth
  \global\arrayrulewidth 0.75pt}\hline\noalign{\global\arrayrulewidth\savewidth}}
\newcommand{\tablestyle}[2]{\setlength{\tabcolsep}{#1}\renewcommand{\arraystretch}{#2}\centering\scriptsize}
\renewcommand{\paragraph}[1]{\vspace{1.25mm}\noindent\textbf{#1}}
\newcolumntype{x}[1]{>{\centering\arraybackslash}p{#1pt}}
\newcolumntype{y}[1]{>{\raggedright\arraybackslash}p{#1pt}}
\newcolumntype{z}[1]{>{\raggedleft\arraybackslash}p{#1pt}}
\newcommand{\app}{\raise.17ex\hbox{$\scriptstyle\sim$}}

\definecolor{deemph}{gray}{0.6}

\definecolor{baselinecolor}{gray}{.9}
\newcommand{\baseline}[1]{\cellcolor{baselinecolor}{#1}}
\newcommand{\gr}{\rowcolor[gray]{.9}}
\newcommand{\grgr}{\rowcolor[gray]{.9}}
\newcommand{\figref}[1]{Figure~\ref{#1}}
\newcommand{\tabref}[1]{Table~\ref{#1}}
\newcommand{\equref}[1]{Eq.~(\ref{#1})}

\newcommand{\secref}[1]{Section~\ref{#1}}

\newcommand\blfootnote[1]{%
  \begingroup
  \renewcommand\thefootnote{}\footnote{#1}%
  \addtocounter{footnote}{-1}%
  \endgroup
}

\newcommand*\circled[1]{\tikz[baseline=(char.base)]{
            \node[shape=circle,draw,inner sep=0.5pt] (char) {\textbf{#1}};}}

%% file: sec/authors.tex
\newcommand*{\affmark}[1][*]{\textsuperscript{#1}}
% \newcommand*{\email}[1]{\texttt{#1}}

% \author{
% {Junha Song}\affmark[1,2]\thanks{Work done during an internship at Lunit Global.} \,,\,\;
% {Tae Soo Kim}\affmark[1],\,\;
% {Gunhee Nam}\affmark[1],\,\;
% {Junha Kim}\affmark[1],\,\;
% {Thijs Kooi}\affmark[1],\,\;
% {Jaegul Choo}\affmark[2]
% \\
% \affmark[1]Lunit$^\ddag$,\;\;
% \affmark[2]KAIST\\
% }

%%% Original CVPR23 templete
% \author{Junha Song\\
% Lunit, KAIST\\
% {\tt\small junhas@lunit.io}
% \and

% \thanks{Work done during an internship at Lunit Inc.}
\author{Junha Song\inst{1,2}\orcidlink{0000-0003-2424-6198} \and
Tae Soo Kim\inst{2}\orcidlink{0009-0007-0726-9597} \and
Junha Kim\inst{2}\orcidlink{0009-0001-6811-2805} \and
Gunhee Nam\inst{2}\orcidlink{0000-0001-6724-1771} \and
\\
Thijs Kooi\inst{2}\orcidlink{0000-0001-7701-7837} \and
Jaegul Choo\inst{1}\orcidlink{0000-0003-1071-4835} 
}

\authorrunning{J. Song et al.}

% \institute{
% \affmark[1]~~KAIST, ~~~\affmark[2]~~Lunit Inc. \\
% \email{sb020518@kaist.ac.kr, \{taesoo.kim, junha.kim, ghnam, tkooi\}@lunit.io, jchoo@kaist.ac.kr}}
% \{sb020518, jchoo\}@kaist.ac.kr} \and
% Lunit \email{lncs@springer.com} \url{http://www.springer.com/gp/computer-science/lncs}}

\institute{
KAIST \email{\{sb020518, jchoo\}@kaist.ac.kr} \and
Lunit Inc. \email{\{junha.kim, taesoo.kim, ghnam, tkooi\}@lunit.io}
}

%% file: sec/0abstract.tex
This paper aims to adapt the source model to the target environment, leveraging small user feedback (\ie, labeled target data) readily available in real-world applications. 
We find that existing semi-supervised domain adaptation\,(SemiSDA) methods often suffer from poorly improved adaptation performance when directly utilizing such feedback data, as shown in \figref{fig:figure1}.
We analyze this phenomenon via a novel concept called \textit{Negatively Biased Feedback}\,(NBF), which stems from the observation that user feedback is more likely for data points where the model produces incorrect predictions.
To leverage this feedback while avoiding the issue, we propose a scalable adapting approach, \textit{Retrieval Latent Defending}.
This approach helps existing SemiSDA methods to adapt the model with a balanced supervised signal by utilizing latent defending samples throughout the adaptation process.
We demonstrate the problem caused by NBF and the efficacy of our approach across various benchmarks, including image classification, semantic segmentation, and a real-world medical imaging application. 
Our extensive experiments reveal that integrating our approach with multiple state-of-the-art SemiSDA methods leads to significant performance improvements.

\keywords{Rethinking user-provided feedback 
\and Semi-supervised \& \\Source-free domain adaptation  
\and Medical image diagnosis}

%% file: sec/1introduction.tex
While deep neural networks have demonstrated remarkable performance in the development domain\,(\ie, source domain)~\cite{resnet, vit}, they often suffer from performance degradation in the deployed domain\,(\ie, target domain) due to domain shift~\cite{ganin2015unsupervised, wang2018deep, Tsai_adaptseg_2018}. 
To mitigate this issue, domain adaptation (DA) techniques have been introduced~\cite{sun2016deep, shot, mme}. 
The most common DA tasks include semi-supervised domain adaptation (SemiSDA) and source-free domain adaptation (SFDA). SemiSDA aims to adapt the model given a small amount of labeled target data along with massive unlabeled target data~\cite{mme, adamatch, zhang2021semi, fixmatch}. SFDA conducts adaptation with only target data without accessing source data considering data privacy or memory constraints in edge devices~\cite{shot, gen_sfda, ecotta}.

Despite such advances in DA, adapting the model with \textit{user feedback} still remains an open area for further research, even though practical machine learning (ML) products often allow users to provide feedback in order to further improve the model in the target environment. 
For example, facial recognition or medical image diagnosis applications enable users to give feedback correcting \textit{wrong} model predictions, as depicted in \figref{fig:figure1}\,(a). 
Since feedback can be modeled in this case as a small amount of labeled target data, it is anticipated that previous SemiSDA methods assuming the same setup would yield promising results.  
However, we observe that they show inferior adaptation performance on multiple DA benchmarks when using such user feedback in practice, as shown in the dark-gray bar\,\colorbox[HTML]{7F7F7F}{\makebox(4,4){\strut\textcolor{black}{~~}}} in \figref{fig:figure1}\,(b).

\input{fig/figure1}

We introduce a novel concept called \textit{Negatively Biased Feedback} (NBF) to explain this phenomenon. NBF is based on the observation that user feedback is more likely to be derived from \textit{incorrect} model predictions.
For example, a radiologist might log a misdiagnosed chest X-ray by the model, as its accuracy directly impacts the patient's survival. 
Interestingly, our observation aligns with findings from cognitive psychology literature~\cite{bad_stronger, negativity_bias} that proves that humans are more likely to react and provide feedback to negative events (\ie, wrong model predictions). 
Since such an NBF scenario is feasible, we analyze its unexpected impact on SemiSDA observed above. We identify that a biased distribution of NBF within the overall data distribution leads to sub-optimal adaptation results, particularly compared to Random Feedback\,(RF). RF represents the classical SemiSDA setup, where labeled data is randomly selected from the target data.

To address the problem caused by NBF, we present a \textit{scalable} approach named \textit{Retrieval Latent Defending}, which can be seamlessly integrated with existing SemiSDA methods. 
Our approach allows them to adapt the model without a strong dependence on the biasedly distributed labeled data. 
Specifically, we balance the supervised adapting signal by appending latent defending samples to the mini-batch and help to keep the model's balanced class discriminability throughout adapting iterations. 
We evaluate the unexpected influence of NBF using various benchmarks, including image classification, semantic segmentation, and medical image diagnosis. Building upon these evaluations, we demonstrate that our approach not only complements, but significantly enhances the performance of multiple SemiSDA methods.

\input{fig/customization_setup}

\vspace{.2em}
The contributions of the paper are as follows:
\vspace{-.6em}
\begin{itemize}[leftmargin=2em, label=$\circ$]
    \setlength\itemsep{-.05em}
    \item We introduce the novel concept called \textit{Negatively Biased Feedback} and uncover that it can lead to sub-optimal adaptation performance of existing SemiSDA methods.
    \item We analyze this problem and present a scalable solution, \textit{Retrieval Latent Defending}, that combines with SemiSDA methods and allows them to avoid the unexpected effect of NBF.
    \item We show that our approach generalizes through diverse DA benchmarks and improves adaptation results of state-of-the-art SemiSDA methods.
    \item We publicly release the code on \url{https://github.com/junha1125/RLD-SemiSDA}.
\end{itemize}

%% file: fig/figure1.tex
\begin{figure}[t]\centering
\includegraphics[width=0.99\linewidth]{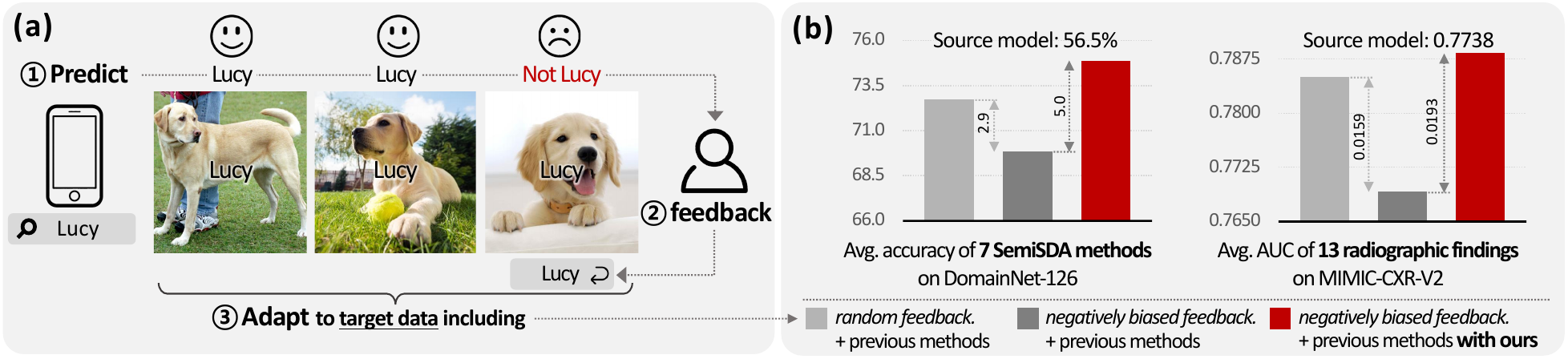}
\vspace{-.7em}
% \caption{\textbf{(a) User feedback for model adaptation.} Users can provide feedback while interacting with an ML product. Our assumption is that user feedback is available in limited number and is biased towards misclassified samples which we define as \textit{Negatively Biased Feedback}.
% Our approach addresses the issue, improving adaptation performance.
\caption{\textbf{(a) User feedback.} Users can provide feedback while interacting with an ML product, where feedback is likely to be biased towards misclassified samples, which we define as \textit{Negatively Biased Feedback}\,(\textit{NBF}).
\textbf{(b) Adaptation results.} We adapt the source model with small user feedback and large unlabeled target data using previous semi-supervised domain adaptation (SemiSDA) algorithms. Compared to \textit{random feedback}, which is the classical SemiSDA setup where labeled data is a random subset of target data, model adaptation with NBF leads to subpar performance. This paper analyzes this problem and introduces a scalable solution.}
\label{fig:figure1}
\vspace{-2.em}
\end{figure}

%% file: fig/customization_setup.tex
\begin{figure*}[t]\centering
\includegraphics[width=0.99\linewidth]{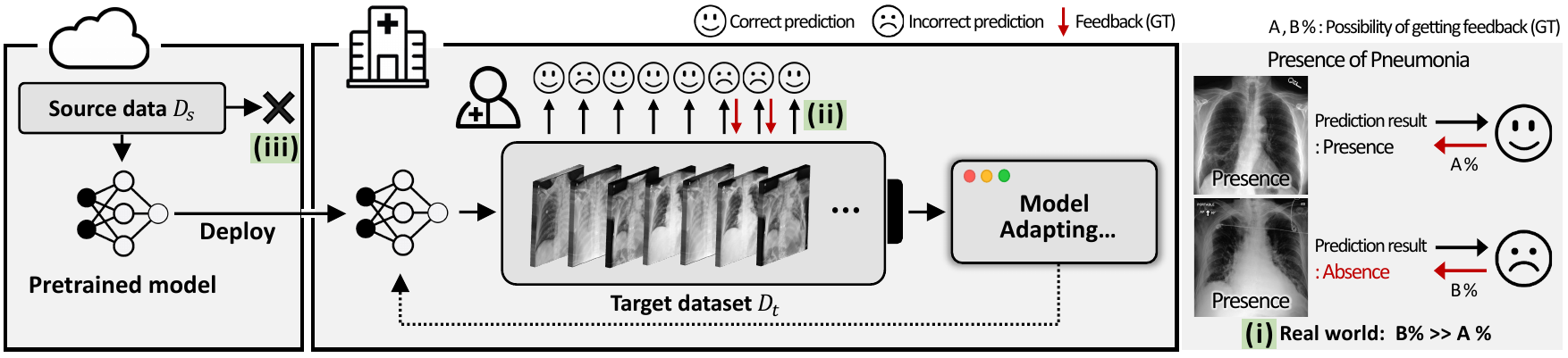}
\vspace{-.7em}
\caption{\textbf{Adaptation with user feedback} can be effective in alleviating performance degradation caused by domain shift. However, there are some challenges: 
\colorbox[HTML]{D2DEC6}{\makebox(7,6){\strut\textcolor{black}{(i)}}} user feedback may be a biased sampling of the true target distribution due to the nature of feedback, 
 % in comparison to the target data
\colorbox[HTML]{D2DEC6}{\makebox(9,6){\strut\textcolor{black}{(ii)}}} the amount of the ground truths {\scriptsize (GT)} labels obtained through feedback is small, and 
\colorbox[HTML]{D2DEC6}{\makebox(11,6){\strut\textcolor{black}{(iii)}}} only unlabeled target data is typically available, not source data.}
\label{fig:custom_setup}
\vspace{-1.5em}
\end{figure*}

% or in changing model behavior to fit user's preferences

%% file: sec/2relatedwork.tex
\paragraph{Adaptation in the deployment environment.} 
Real-world ML products often encounter performance degradation caused by gaps between the source and target environment~\cite{ganin2015unsupervised}.
One solution is to adapt the model using unlabeled data observed in the target domain, 
referred to as \textbf{unsupervised domain adaptation} (UDA)~\cite{Tsai_adaptseg_2018, saito2018maximum, liu2022deep}.
Works on UDA use both source and target data to improve the target performance by using methods such as domain discrepancy minimization by adversarial training~\cite{long2017deep, sun2016deep, advent, ganin2016domain, tsai2018learning, Tsai_adaptseg_2018, saito2018maximum}, and self-training with pseudo labels~\cite{mei2020instance, zhang2018collaborative, pan2019transferrable, zhang2023making}.
\textbf{Source-free DA} (SFDA) builds on UDA and imposes an additional constraint that the source data can not be accessed during domain adaptation. This has practical implications for addressing data privacy concerns or barriers in data transmission to edge devices~\cite{shot, liu2021source, wang2020tent, yu2023comprehensive}. The majority of recent SFDA works rely on strategies like domain clustering~\cite{shot}, nearest neighbors~\cite{gen_sfda, nrc, yang2022attracting}, and contrastive learning~\cite{contrastivetta, guidingps, zhang2023class}. Nevertheless, SFDA does not consider the availability of small labeled data, which may be available in practical ML systems.
\textbf{Semi-supervised DA} (SemiSDA) works mainly demonstrate that permitting small labeled data in the target domain can substantially enhance adaptation performance compared to traditional UDA~\cite{mme}. Their primary strategy is to use domain alignment~\cite{mme, cdac, harada2023cluster, sla}, multi-view consistency~\cite{cdac, adamatch, basak2023semi, yan2022multi}, and asymmetric co-training~\cite{liu2022act, yang2021deep}.

\paragraph{Active domain adaptation}\,(ActiveDA)~\cite{clue, sdm, diana} envisions a scenario in which the machine selects specific target samples and instructs annotators to label them. The primary objective of ActiveDA is to strategically identify and select the most informative samples for annotation. These chosen samples (\ie, labeled target data) are subsequently utilized to update the source model using SemiSDA methods~\cite{mme, cdac}, and the effectiveness of ActiveDA is assessed by evaluating the target performance of the adapted model.

\paragraph{Semi-supervised learning} (SemiSL) aims to reduce expensive human annotations, and propose methods to train a model from scratch using massive unlabeled data along with limited amounts of labeled data~\cite{van2020survey, madani2018semi}.
The majority of SemiSL methods depend on consistency regularization~\cite{sajjadi2016regularization, fixmatch, uda, mixmatch, berthelot2019remixmatch, fini2023semi}, which helps the model to make similar predictions for augmented versions of the same image. 
Moreover, adaptive thresholding~\cite{fixmatch, freematch, higuchi2021momentum, xu2021dash, chen2023boosting, chen2023softmatch, zhang2021semi} is also popularly utilized to produce reliable pseudo labels from unlabeled data. 

SemiSDA and SemiSL setups mimic small labeled datasets by randomly selecting subsets of the target dataset, whereas ActiveDA involves selections instructed by the machine.
In contrast, this paper posits that in real-world applications, labeled data is typically acquired through user intervention. Additionally, users often provide feedback on samples misclassified by the model (\ie, negatively biased feedback), a process detailed in the following section.

\input{table/relatedworks}

The table above summarizes the comparison of relevant studies to our setup. In the table, adaptation means fine-tuning the source pre-trained model {\small (}as opposed to training from scratch{\small )}; feedback represents a small number of labeled target samples. Appendix~\ref{sec:addrelated} provides further comparisons with settings like class-imbalanced SemiSDA and test-time adaptation (TTA).

%% file: table/relatedworks.tex
\begin{table}[h]
\tablestyle{3pt}{1.15}
\vspace{-2.0em}
\resizebox{0.99\textwidth}{!}{
\begin{tabular}{y{40}              x{40}                    x{40}                     x{70}                     x{65}                     x{65}                     x{65}                                             }
                                   & UDA                    & SFDA                    & ActiveDA                        & SemiSDA                  & SemiSL                & \baseline{Our setup}                               \\
\shline
Adaptation                         & $\circ$                & $\circ$                 & $\circ$                         & $\circ$                 & {\scriptsize $\times$}  & \baseline{$\circ$}                            \\
Source-free                        & {\scriptsize $\times$} & $\circ$                 & {\scriptsize $\times$}          & {\scriptsize $\times$}  & -                       & \baseline{$\circ$}                            \\
\textbf{Feedback}                           & {\scriptsize $\times$} & {\scriptsize $\times$}  & {\scriptsize machine-instructed}  & {\scriptsize randomly selected}    & {\scriptsize randomly selected}    & \baseline {\scriptsize \textbf{user-provided}}     \\

\end{tabular}}
\vspace{-1.em}
% \caption{\textbf{Comparison of related setups.} We compare our setup to 
% unsupervised domain adaptation\,{\small (UDA)}, 
% source-free UDA\,{\small (SFDA)},
% and semi-supervised DA\,{\small (SemiSDA)}, and semi-supervised learning\,{\small (SemiSL)}.
% Adaptation means fine-tuning the source pre-trained model, as opposed to training from scratch. Feedback represents a small amount of annotations obtained by users. Source-free represents adapting without source data.}
\label{tab:related} 
\vspace{-2.em}
\end{table}

% \begin{table}[t]
% \tablestyle{3pt}{1.15}
% \begin{tabular}{y{40} x{22} x{32} x{32} x{22} x{50} }
%                                    & UDA                     & SemiSDA                 & SemiSL                 & SFDA                    & \baseline{Ours}                               \\
% \shline
% Adaptation                         & {\scriptsize \cmark}                 & {\scriptsize \cmark}                 & {\scriptsize \xmark} & {\scriptsize \cmark}                 & \baseline{{\scriptsize \cmark}}                            \\
% Feedback                           & {\scriptsize \xmark}  & {\scriptsize random}    & {\scriptsize random}   & {\scriptsize \xmark}  & \baseline {\scriptsize negatively biased}     \\
% Source-free                        & {\scriptsize \xmark}  & {\scriptsize \xmark}  & -                      & {\scriptsize \cmark}                 & \baseline{{\scriptsize \cmark}}                            \\

% \end{tabular}
% \vspace{-1.em}
% \caption{\textbf{Comparison of related setups.} We compare our setup to 
% unsupervised domain adaptation\,{\small (UDA)}, 
% and semi-supervised DA\,{\small (SemiSDA)},
% semi-supervised learning\,{\small (SemiSL)}, 
% source-free DA\,{\small (SFDA)}.
% Adaptation means fine-tuning the source pre-trained model, as opposed to training from scratch. Feedback represents a small amount of annotations obtained by users. Source-free represents adapting without source data.}
% \label{tab:related} \vspace{-2.em}
% \end{table}

%% file: sec/3problem_definition.tex
\subsection{Adaptation with user feedback.} \vspace{-.5em}
Our adaptation setup is illustrated in \figref{fig:custom_setup}. A model is pre-trained on the source data $D_s$. 
Next, the model is deployed to the target domain, such as a smartphone or a hospital, where we assume the transfer of $D_s$ is prohibited due to data privacy regulations or resource constraints (same setup as SFDA\,\cite{shot}). 
While users utilize ML products on the target domain, the model provides prediction results for data observed in the target domain $D_t$ and occasionally obtains user feedback in the form of annotations $y$.
We represent the target data as $D_t=X^{lb}_{t}\cup X^{ulb}_{t}$, where $X^{lb}_{t}=\{(x^{n}_{lb}, y^{n}_{lb}):n\in[1..\,N_{lb}]\}, X^{ulb}_{t}=\{(x^{n}_{ulb}):n\in[1..\,N_{ulb}]\}$, $x_{lb}$ and $x_{ulb}$ denote labeled and unlabeled data and $N_{lb}$ and $N_{ulb}$ is their number of data. 
Lastly, the model can utilize $D_t$ and SemiSDA algorithms for adaptation during its inactive phase (\eg, when users do not use the product, like at nighttime) in order to alleviate performance degradation due to domain shift or to personalize the model based on user feedback.

\input{fig/toyexperiment}

\paragraph{Rethinking user-provided feedback.}
Classical SemiSDA works simply assume that a random subset in target data $D_t$ is labeled by users when building $X^{lb}_{t}$.
However, as illustrated in \figref{fig:custom_setup}\,\colorbox[HTML]{D2DEC6}{\makebox(7,7){\strut\textcolor{black}{(i)}}}, we suggest that users are more likely to provide feedback on misclassified samples by the source model, named negatively biased feedback (NBF). This behavior can be understood from two perspectives: (a) users generally expect their feedback to be used as a basis of model improvement, motivating them to provide NBF, and (b) humans tend to react more strongly to negative experiences, such as receiving incorrect predictions, as observed in psychological studies~\cite{bad_stronger, negativity_bias}. 
We note that the NBF assumption holds more strongly for the \textit{medical} application: it is reasonable to imagine that the user (\ie, radiologist) logs the \textit{mistakes} of the model while diagnosing a chest X-ray exam because the diagnostic accuracy of the model is directly related to the patient's chances of survival. 
Furthermore, applications beyond the medical domain can also exhibit NBF. For instance, 
users in self-driving cars can report errors, such as object detection failures or navigation mistakes, to enhance the car's driving capabilities.

\vspace{-.6em}
\subsection{Influence of NBF on SemiSDA} \label{sec:nbf_effect}
\vspace{-.5em}
\paragraph{Simulation study.} As shown in \figref{fig:toy_exper}, we conduct a simulation study to understand the effect of NBF on SemiSDA. We first use the blobs dataset~\cite{scikit-learn} and construct the source and target data so that domain shift exists between them (left sub-figures). We pre-train a source model on the source data and compute the accuracy in the target domain, where the performance drop due to domain shift is observed (98.5\%→76.4\%). Next, we simulate two types of feedback (\ie, labeled data): random feedback and negatively biased feedback following a previous SemiSDA setup and our setup, respectively. Specifically, NBF is randomly selected among misclassified samples by the source model. We find that random feedback (RF) points are \textit{evenly} distributed, while NBF points are \textit{biasedly} positioned across each class cluster (refer to blue points in the dashed circle in the center sub-figures).

To alleviate the performance drop caused by domain shift, we adapt the model using the target data and a semi-supervised method, Pseudo-labeling~\cite{arazo2020pseudo}. This method iteratively optimizes the model by the cross-entropy loss computed by the ground truth of labeled data and pseudo labels of unlabeled data in a mini-batch {\small (}pseudo labels are predicted by the \textit{current} adapting model so they can be changed according to an updating decision boundary. 
Further comprehension can be achieved by referring Appendix~\ref{sec:addsimulation}.{\small )}. The SemiSDA results are shown in the right sub-figures, where we make two interesting observations: (i) the distribution of labeled data can contribute significantly to a decision boundary of the adapted model (red arrows in the figure), and (ii) the adapted model under NBF has poorly improved performance compared with one under RF (76.4\%→88.1\% with NBF, but 76.4\%→91.7\% with RF). 

\paragraph{\underline{Unexpected influence of NBF.}} 
Our intuitive reasoning probably suggests that NBF provides more information than RF by correcting more source model deficiencies, and thus leads to better adaptation performance. However, we empirically show that NBF can result in inferior adaptation performance due to its biased distribution across each class cluster, as illustrated in \figref{fig:toy_exper}. 
Surprisingly, we also show that this problem persists, even with other state-of-the-art SemiSDA methods and large datasets for various DA benchmarks, including image classification, semantic segmentation, and medical image diagnosis.
Our work highlights the importance of careful design when using user feedback in real-world scenarios and, to the best of our knowledge, is the first study to uncover and analyze this phenomenon.

%% file: fig/toyexperiment.tex
\begin{figure}[t]\centering
% \vspace{-2.em}
\includegraphics[width=0.90\linewidth]{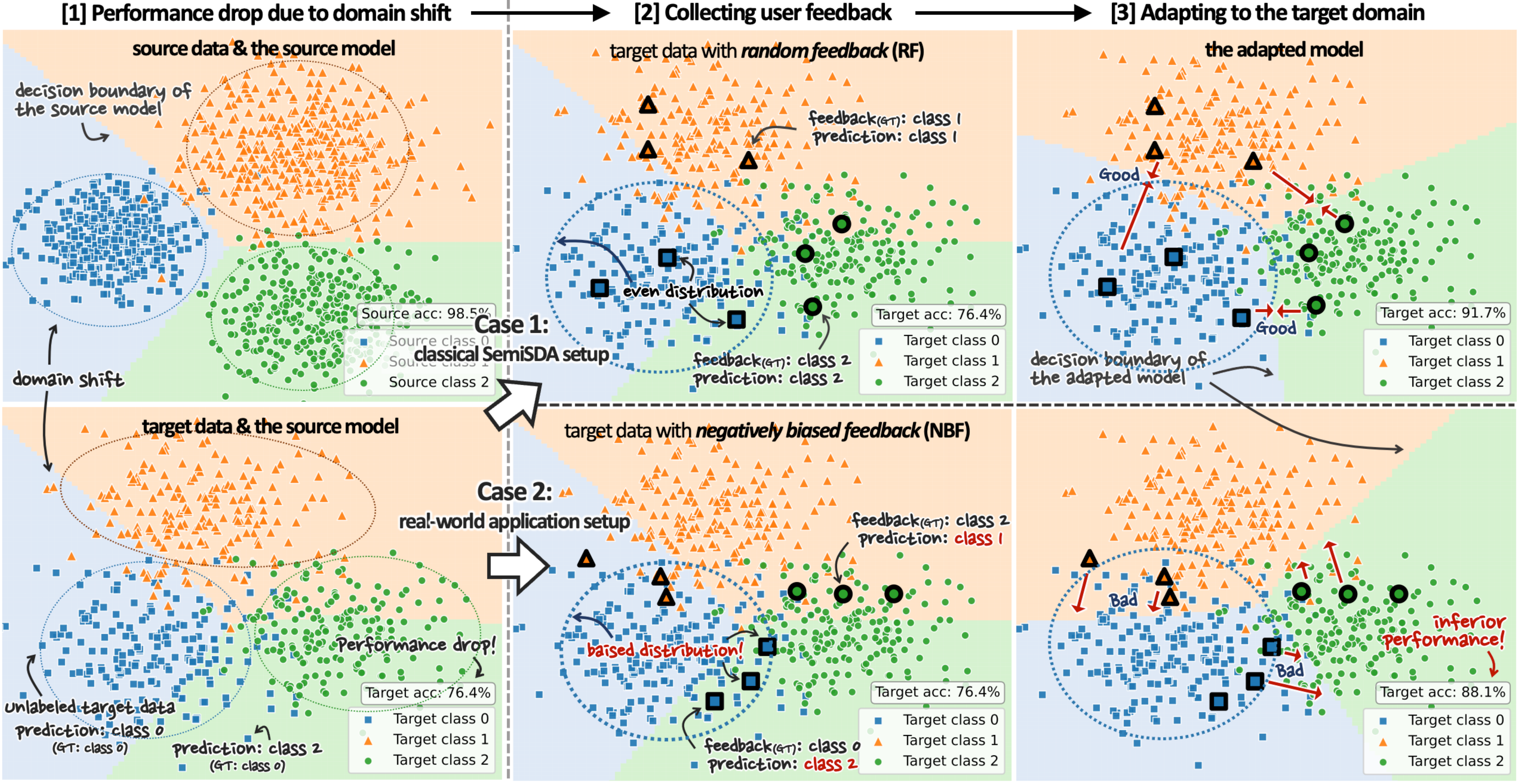}
\vspace{-1.em}
\caption{\textbf{Effect of negatively biased feedback.} Our novel observations are that \textbf{(\textit{a})}~user-provided feedback in practice has a biased distribution in each class cluster (the bottom center sub-figure) which is in contrast to random feedback, \textbf{(\textit{b})}~Existing SemiSDA methods adapt the model by dominating the labeled data points (the right sub-figures) even though they are biasedly positioned, and \textbf{(\textit{c})}~NBF prevents the model from having a decision boundary for true class clusters and leads to inferior adaptation performance (the bottom right sub-figure).}
\label{fig:toy_exper}
\vspace{-1.5em}
\end{figure}
% %The adaptation results on the target domain is presented. NBF is biasedly positioned, resulting in inferior adaptation performance in comparison to RF. }

%% file: sec/4approach.tex
\subsection{Prerequisite: Previous SemiSDA method} \label{sec:prerequ}\vspace{-.3em}
Previous SemiSDA and SemiSL works typically construct a mini-batch with labeled data {\small $\{(x^{b}_{lb}, y^{b}_{lb}):b\in[1..\,B]\}$}, and unlabeled data whose size is $\mu$ times larger than labeled ones {\small $\{(x^{b}_{ulb}):b\in[1..\,\mu 
{\cdot}B]\}$}, where $B$ is the mini-batch size for labeled data. 
To adapt the model iteratively, they compute the cross-entropy loss $\mathcal{H}(\cdot, \cdot)$ with labeled data and the consistency regularization to multi-view of unlabeled data, which are formulated as the following:
\vspace{-.7em}
\begin{equation}\label{equ:unsuploss}\footnotesize
    \mathcal{L}_{sup} = \frac{1}{B}\sum^{B}_{b=1} \mathcal{H}(y_{lb}^{b},f_{\theta}(x_{lb}^{b})), ~~ \mathcal{L}_{unsup} = \frac{1}{\mu \cdot B}\sum_{b=1}^{\mu \cdot B} \mathcal{H}(\hat{y}_{ulb}^{b}, f_{\theta}(\Omega(x_{ulb}^{b}))),
    \vspace{-.5em}
\end{equation}
where $f_{\theta}(\cdot)$ is the output probability from the model, {\small $\hat{y}_{ulb}$} denotes a pseudo label obtained from {\small $f_{\theta}(\omega(x_{ulb}))$}, and $\omega(\cdot)$ and $\Omega(\cdot)$ represent weak and strong image augmentation, respectively.
While sharing the core framework, each SemiSDA method employs distinct adapting strategies, especially to enhance the effectiveness of the use of \textit{unlabeled} data rather than \textit{labeled} data~\cite{adamatch, flexmatch, freematch}.

\paragraph{Problem of previous works.} 
Since previous SemiSDA methods have overlooked the unexpected impact of NBF, they often suffer from sub-optimal performance under the NBF assumption {\small (}shown in \secref{sec:results}{\small )}.
To address this problem, we focus on developing a \textit{scalable} solution that {\small (i)} can easily combine with existing DA methods without modifying their core adapting strategies and {\small (ii)} can be applied to a wide range of benchmarks, including medical image diagnosis.

\input{fig/approach}

\vspace{-.5em}
\subsection{Retrieval Latent Defending} \label{sec:rld}
\vspace{-.4em}

Based on the observations in \figref{fig:toy_exper}, we illustrate the unintended effect of NBF when using an existing SemiSDA method in \figref{fig:approach}\,(top center), where
NBF is likely to exhibit a biased distribution, leading to undesirable adaptation results. 
To alleviate this issue, we propose \textit{Retrieval Latent Defending} as depicted in \figref{fig:approach}\,(bottom). 
{\scriptsize \circled{1}} 
Prior to each epoch, we generate a candidate bank of data points, denoted as {\small $x_{LD}$}.
{\scriptsize \circled{2}}{\small $\sim$}{\scriptsize \circled{4}} 
For each adapting iteration, we balance the mini-batch by retrieving latent defending samples {\small $x_{LD}$} from the bank. 
% For each adapting iteration, we balance a mini-batch by selecting defending samples {\small $x_{LD}$} from the bank. 
{\scriptsize \circled{5}}{\small $\sim$}{\scriptsize \circled{6}} 
The model is then adapted using the reconfigured mini-batch and following the baseline SemiSDA approach.
We hypothesize that the latent space progressively created by the $x_{LD}$ candidates throughout the adaptation process (bold dashed circle in \figref{fig:approach}~(top right)) mitigates the issue caused by NBF, thereby allowing the SemiSDA baseline to achieve robust adaptation against NBF.

\paragraph{Candidate bank generation.} 
The candidate bank serves as a repository of pseudo labels $\hat{Y}^{ulb}_{t}$ for a subset of the target unlabeled data $X^{ulb}_{t}$.
Before each epoch, we freeze the model and use it to generate pseudo labels $\hat{Y}^{ulb}_{t}{=}\{(\hat{y}^{n}_{ulb}):n\in[1..\,N_{ulb}]\}$, 
where $\hat{y}^{n}_{ulb}$ is assigned to $x^{n}_{ulb}$ as the predicted class with the highest softmax probability: {\small $\hat{y}^{n}_{ulb} = \argmax_{c} \left[f_{\theta}(x^{n}_{ulb})\right]_c$}.
We then retain only samples with the top $p$\% highest probabilities within each class. 
This filtering step helps mitigate the inclusion of data with potentially inaccurate pseudo labels, as the model's predictions on $X^{ulb}_{t}$ might not always be perfect.

\paragraph{Defending sample selection.} 
We select $k$ latent defending samples $x_{LD}$ from the bank at random for each labeled data {\small $(x^{b}_{lb}, y^{b}_{lb})$}.
These selected samples share the \textit{same} pseudo label as the ground-truth label of their associated counterparts~(\ie, $\hat{y}_{LD}\,{=}\,y^{b}_{lb}$).
By incorporating these defending samples, we balance the data distribution within the current mini-batch.  
For example, consider $x^{1}_{lb}$ and $x^{2}_{lb}$ in \figref{fig:approach}\,(top right). 
As these labeled samples are included in the current mini-batch alongside the selected defending samples $x^{1}_{LD}$ and $x^{2}_{LD}$, we expect to prevent the supervised adapting signal from becoming overly dependent on the labeled samples.
We imagine the effect of the defending samples \textit{throughout} the adaptation process and depict the latent space formed gradually by the $x_{LD}$ candidates as bold dashed circles in \figref{fig:approach}\,(top right).

Consequently, the overall loss consists of the sum of losses in \equref{equ:unsuploss} and a loss from our proposed method as,
\vspace{-.7em}
\begin{equation}\label{equ:overallloss}\scriptsize
    \mathcal{L}_{total} = \underbrace{\mathcal{L}_{sup} + \mathcal{L}_{unsup}}_{\text{baseline}} + \underbrace{\frac{1}{k\cdot B}\sum_{b=1}^{k\cdot B} \mathcal{H}(\hat{y}_{LD}^{b},f_{\theta}(x_{LD}^{b}))}_{\text{retrieval latent defending}}.
    \vspace{-.5em}
\end{equation}
\vspace{-.7em}

\paragraph{\underline{Importance of our method.}}
Understanding the impact of NBF on adaptation performance is crucial. For example, naively adapting a model for a medical application using radiologist-provided feedback can actually cause performance degradation (shown in Table~\ref{tab:mimic}), potentially posing significant risks to patients. 
We propose a \textit{scalable} and \textit{simple} approach to solve the problem caused by NBF, which can not be addressed by existing methods. 
Given the practicality of the NBF problem and the scalability of our solution, we believe our work holds considerable potential for real-world applications.

%% file: fig/approach.tex
\begin{figure*}[t]\centering
\includegraphics[width=0.93\linewidth]{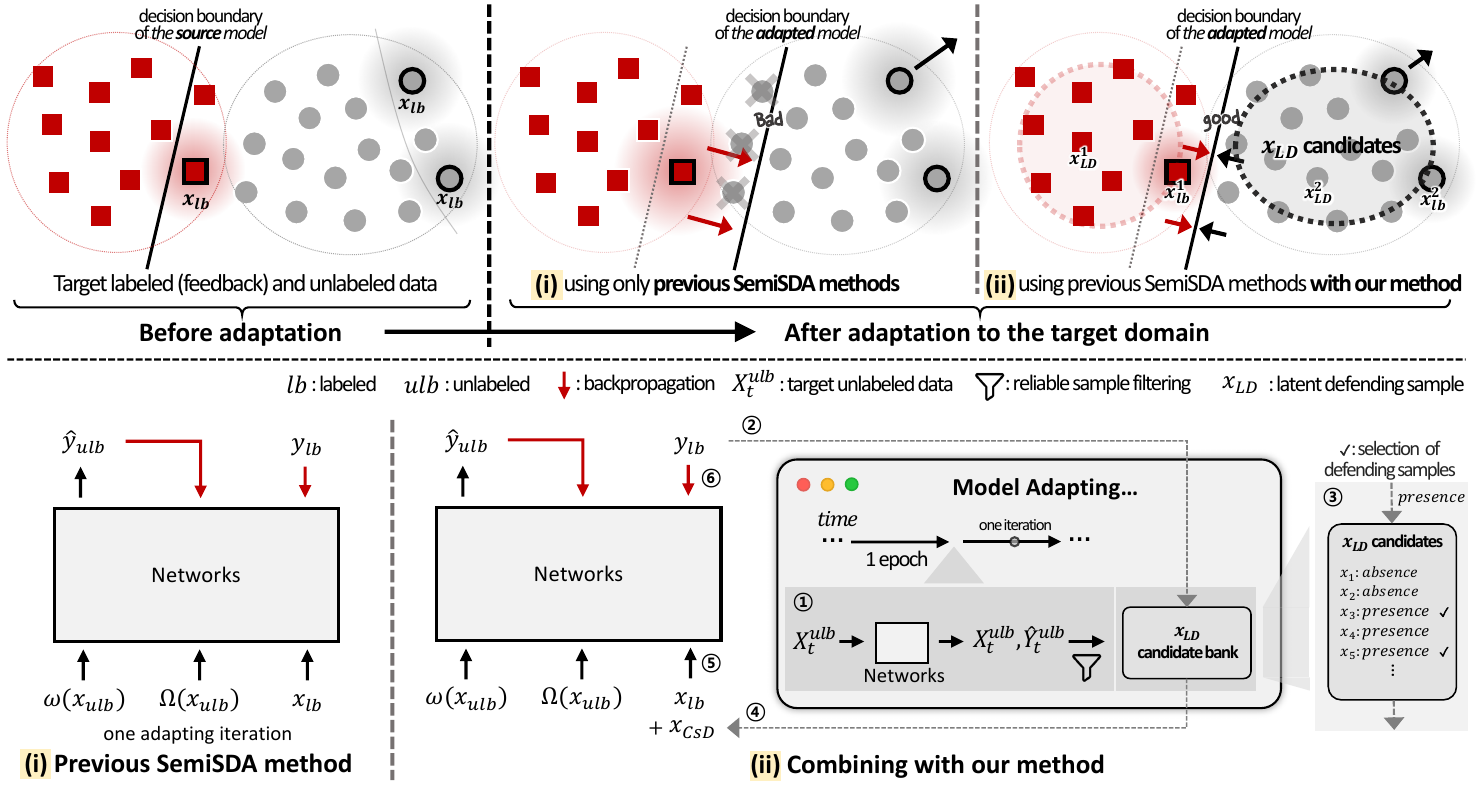}
\vspace{-1.em}
\caption{Even though labeled data {\small $(x_{lb}, y_{lb})$} is biasedly positioned, the model needs to be adapted with balanced class discriminability (\ie, decision boundary). 
\colorbox{highlight!85}{\makebox(8,6){\strut\textcolor{black}{\textbf{(i)}}}} However, previous SemiSDA methods have overlooked this fact and used the labeled data naively by applying a cross-entropy loss, leading to inadequate adaptation performance.
\colorbox{highlight!85}{\makebox(9,6){\strut\textcolor{black}{\textbf{(ii)}}}} To alleviate this problem, we propose a scalable adapting approach, retrieval latent defending, which allows the model to adjust the balance of a mini-batch on each iteration by using latent defending samples {\small $x_{LD}$} together with labeled data {\small $x_{lb}$}.
}
\label{fig:approach}
\vspace{-1.7em}
\end{figure*}
% Labeled {\small $x_{lb}$} and unlabeled {\small $x_{ulb}$} target data in a mini-batch are used at each iteration. Two cross-entropy losses are computed with the ground truth {\small $y_{lb}$} (\ie, user-provided feedback) and the pseudo label {\small $\hat{y}_{ulb}$} using weak {\small $\omega$} and strong $\Omega$ augmentation. (ii) We generate the pseudo label bank prior to each epoch and balance a mini-batch by randomly selecting class-aware samples {\small $x_{Cab}$} with the same label as $y_{lb}$ from the bank.
% In addition, tha majority of SemiSDA works depend on consistency regularization for unlabeled {\small $x_{ulb}$}, where 

%% file: sec/5experimental_setups.tex
\vspace{-.7em}
\subsection{Experimental Setups}
\vspace{-.3em}
Our approach is simple enough to seamlessly combine with existing SemiSDA algorithms and also be applied to diverse benchmarks. This section describes our experimental setup for natural image classification tasks and a real-world medical application. 
Details for semantic segmentation experiments are in Appendix~\ref{sec:adddetail}.

\vspace{0.2em}\noindent
\textbf{Baselines.}
We validate our approach by combining various state-of-the-art algorithms for SemiSDA~\cite{mme} {\small (}\eg, CDAC~\cite{cdac} and AdaMatch~\cite{adamatch}{\small )} and SemiSL~\cite{fixmatch, uda} {\small (}\eg, FlexMatch~\cite{flexmatch} and FreeMatch~\cite{freematch}{\small )}.
Note that the SemiSL methods have been demonstrated to be strong \textit{SemiSDA learners}~\cite{zhang2021semi}, so we can consider them as SemiSDA methods. For medical experiments, we use Pseudo-labeling~\cite{arazo2020pseudo} as a baseline since it is easily applicable to medical image adaptation. 

\vspace{0.2em}\noindent
\textbf{Datasets.}
We utilize natural image datasets containing multiple kinds of domains\,(\eg, real and painting).
The datasets include DomainNet-126~\cite{domainnet, mme} with 142k images of 126 classes, and OfficeHome~\cite{officehome} with 15K images of 65 classes.

\vspace{-0.1em}
To conduct medical experiments, we present a practical medical setting. 
We adopt the MIMIC-CXR-V2 dataset~\cite{mimic}. It assumes a multi-finding binary classification setup, where multiple radiographic findings, like Pneumonia and Atelectasis, can coexist in a single chest X-ray\,(CXR) sample. Thus, the model predicts the presence or absence (binary classes) of \textit{each individual} finding.
We simulate domain shift by using Posterior-Anterior\,(PA)-view data as the source and AP-view data as the target, capturing real-world variations in data acquisition.
Typically, patients requiring an AP X-ray are those facing positioning challenges that prevent them from undergoing a PA X-ray.
Therefore, this setup can be seen as a scenario where the target environment is the intensive care unit, which hospitalizes critically ill patients.

\vspace{-0.1em}
Following the recent SemiSDA~\cite{sla} and SFDA~\cite{contrastivetta} setups, we assume the model is pre-trained in the source domain and deployed in the target domain.
Since the datasets above were not initially divided into training and test sets, we performed a random 8:2 split within each domain, designating them respectively for training and testing. The training set is used to adapt the model, while the test set is used to report the top-1 accuracy.

\input{table/domainnet_semis_1}
\input{table/officehome_semis_1}

\vspace{0.2em}\noindent
\textbf{User feedback.}
Feedback given by users is modeled as annotations $\{y^{n}_{lb}:n\in[1..\,N_{lb}]\}$ on a small subset of the target's training set $D_t^{train}$, while the remaining of them are used as unlabeled target data.
In our experiments, we take into account two types of feedback: random feedback (RF) and negatively biased feedback (NBF).
RF is the same setup of classical SemiSDA and SemiSL, where randomly selected samples from $D_t^{train}$ are used as small labeled set $X_{t}^{lb}$.
For NBF, we randomly select samples that are incorrectly predicted in $D_t^{train}$ by the source model~(\ie, the pre-trained model before adaptation). 
Note that we focus on the impact of a biased label distribution within the \textit{same} class, as shown in \figref{fig:toy_exper}, and thus take the same number of feedback for each class. Further discussion about the imbalance in the number of feedback \textit{between classes} presented in \cite{wei2021crest, oh2022daso, lee2021abc} is provided in Appendix~\ref{sec:classim}.

\vspace{0.2em}\noindent
\textbf{Network architectures.}
We adopt commonly used networks, ResNet \cite{resnet} and ViT~\cite{vit} for natural image tasks and DenseNet~\cite{densenet} for a medical task.
We employ ResNet-50 with the last classification layer comprising a weight normalization layer and a bottleneck layer, following previous works~\cite{shot, contrastivetta} and use the ViT-Small (\ie, ViT-S) introduced in \cite{usb}. The DenseNet-121 is used, provided in TorchXrayVision~\cite{torchxray}, like existing medical works~\cite{lenga2020continual, mahapatra2022unsupervised}.

\vspace{0.2em}\noindent
\textbf{Implementation details.}
We implement our framework by extending the publicly available USB~\cite{usb} repository.
Both pre-training and adaptation are conducted with a mini-batch size of 128 and the SGD optimizer. 
Diverse baselines for SemiSDA and SemiSL are used to compute the losses in \equref{equ:unsuploss}.
The hyper-parameters for each baseline simply follow USB~\cite{usb} or public code~\cite{mme, cdac}. 
For all experiments, our approach uses the same hyper-parameters of the appended defending samples\,$k$ and reliable filtering rate\,$p$ as 3 and 0.4, respectively.

\input{table/domainnet_semis_2}

\input{table/officehome_semis_2}

%% file: table/domainnet_semis_1.tex
\begin{table*}[t]
\tablestyle{3.5pt}{1.2}
\resizebox{0.93\textwidth}{!}{
\begin{tabular}{x{5} y{50} x{35} x{50} ;{0.5pt/1.pt} x{35} x{35} x{35} x{35} x{35} x{35} x{35}} 
& method    & feedback & average & r{\scriptsize →}c & r{\scriptsize →}p & p{\scriptsize →}c & c{\scriptsize →}s & s{\scriptsize →}p & r{\scriptsize →}s & p{\scriptsize →}r \\
\shline
% &FreeMatch~\cite{freematch}  & {\scriptsize RF}  & 73.8 & 76.6 & 74.2 & 75.5 & 67.7 & 73.5 & 65.1 & 84.0  \\
% &          &         {\scriptsize NBF} & 72.0\,{\tiny(-1.8)} & 75.5 & 72.9 & 74.6 & 65.0 & 72.3 & 62.0 & 81.7  \\
% \gr
% \multirow{-3}{*}{\cellcolor{white}\rotatebox[origin=c]{90}{{\tiny ResNet-50}}}&w/\,ours     & {\scriptsize NBF} & \textbf{74.8}\,{\tiny(+2.8)} & 78.1 & 74.5 & 77.1 & 68.8 & 72.4 & 67.3 & 85.0  \\
% % \cdashline{2-11}
&AdaMatch~\cite{adamatch}  & {\scriptsize RF}  & 67.6 & 66.6 & 68.5 & 68.5 & 60.3 & 69.2 & 58.7 & 81.5  \\
&          & {\scriptsize NBF} & 64.5\,{\tiny(-3.1)} & 64.3 & 66.1 & 65.6 & 56.9 & 65.6 & 54.2 & 78.9  \\
\gr
\multirow{-3}{*}{\cellcolor{white}\rotatebox[origin=c]{90}{{\tiny ResNet}}}&w/\,ours     & {\scriptsize NBF} & \textbf{72.0}\,{\tiny(+7.5)} & 74.5 & 72.7 & 73.9 & 65.5 & 70.0 & 64.3 & 83.2  \\
\shlinesmall
% & FreeMatch~\cite{freematch}  & {\scriptsize RF}  & 74.9 & 75.3 & 76.8 & 74.5 & 68.1 & 76.5 & 67.0 & 86.0  \\
% &          & {\scriptsize NBF} & 73.9\,{\tiny(-1.0)} & 74.6 & 76.4 & 75.0 & 66.0 & 74.5 & 66.5 & 84.1  \\
% \gr
% \multirow{-3}{*}{\cellcolor{white}\rotatebox[origin=c]{90}{{\tiny ViT-S}}}& w/\,ours     & {\scriptsize NBF} & \textbf{75.7}\,{\tiny(+1.8)} & 76.9 & 77.5 & 77.9 & 68.1 & 76.7 & 67.8 & 85.2  \\
% % \cdashline{2-11}
& AdaMatch~\cite{adamatch}  & {\scriptsize RF}  & 74.7 & 75.3 & 76.9 & 73.8 & 68.0 & 76.3 & 67.1 & 85.5  \\
&          & {\scriptsize NBF} & 73.7\,{\tiny(-1.0)} & 74.7 & 76.2 & 74.7 & 65.7 & 74.0 & 66.8 & 84.0  \\
\gr
\multirow{-3}{*}{\cellcolor{white}\rotatebox[origin=c]{90}{{\tiny ViT}}}& w/\,ours     & {\scriptsize NBF} & \textbf{75.9}\,{\tiny(+2.2)} & 76.9 & 77.8 & 77.8 & 68.5 & 76.6 & 68.3 & 85.1 
\end{tabular}}
\vspace{+.1em}
\caption{\textbf{Adaptation results on DomainNet-126.} We simulate seven domain-shift scenarios (\ie, source\,→\,target). The model is pre-trained on the source domain and then adapted to a training set of the target domain. The results on the test set of the target domain are reported as the top-1 accuracy (\%). DomainNet-126~\cite{domainnet, mme} dataset includes \underline{r}eal, \underline{p}ainting, \underline{s}ketch, and \underline{c}lip-art domains. In this experiment, we assume that the 378 feedback samples (\ie, 3 labeled data per class) are obtained from users. A state-of-the-art SemiSDA method, AdaMatch~\cite{adamatch}, is used as a baseline.}
\label{tab:domainnet1} 
\end{table*}
% in seven domain-shift scenarios 
 % to compute the losses in \equref{equ:unsuploss}.

%% file: table/officehome_semis_1.tex
\begin{table*}[t]
\tablestyle{4pt}{1.15}
\vspace*{-2.5em}
\resizebox{0.93\textwidth}{!}{
\begin{tabular}{y{44} x{35} x{38} ;{0.5pt/1.pt} x{18} x{18} x{18} x{18} x{18} x{18} x{18} x{18} x{18} x{18} x{18} x{18} x{18}} 
method    & feedback & average  & a\,→\,c & a\,→\,p & a\,→\,r & c\,→\,a & c\,→\,p & c\,→\,r & p\,→\,a & p\,→\,c & p\,→\,r & r\,→\,a & r\,→\,c & r\,→\,p \\
\shline
% FreeMatch~\cite{freematch} & {\scriptsize RF}  & 74.0 & 58.5 & 85.0 & 79.4 & 68.2 & 84.7 & 79.2 & 68.4 & 62.5 & 80.4 & 71.0 & 63.7 & 87.0  \\
%           & {\scriptsize NBF} & 72.2\,{\tiny(-1.8)} & 56.4 & 79.3 & 77.7 & 67.7 & 83.4 & 78.5 & 67.3 & 60.5 & 79.1 & 69.2 & 61.0 & 86.9  \\
% \gr
% w/\,ours     & {\scriptsize NBF} & \textbf{74.8}\,{\tiny(+2.6)} & 60.6 & 81.4 & 81.5 & 70.8 & 86.7 & 80.0 & 68.6 & 61.6 & 81.7 & 69.8 & 66.2 & 89.2  \\
AdaMatch~\cite{adamatch}   & {\scriptsize RF}  & 70.9 & 55.4 & 80.4 & 75.9 & 65.7 & 81.5 & 74.6 & 65.9 & 58.7 & 78.4 & 68.8 & 61.5 & 84.3  \\
          & {\scriptsize NBF} & 69.3\,{\tiny(-1.6)} & 54.2 & 76.6 & 75.3 & 65.9 & 79.3 & 75.5 & 63.7 & 57.4 & 75.9 & 66.7 & 56.8 & 84.2  \\
\gr
w/\,ours     & {\scriptsize NBF} & \textbf{73.8}\,{\tiny(+4.5)} & 62.2 & 81.0 & 79.7 & 68.8 & 85.4 & 78.6 & 67.7 & 61.7 & 79.5 & 69.0 & 64.1 & 88.2 
\end{tabular}}
\vspace{+.1em}
\caption{\textbf{Adaptation results on OfficeHome.} OfficeHome~\cite{officehome} dataset includes \underline{r}eal, \underline{p}roduct, \underline{a}rt, and \underline{c}lip-art domain. We assume that the 195 feedback samples (\ie, 3 labeled data per class) are obtained. AdaMatch~\cite{adamatch} and ResNet-50~\cite{resnet} are used.}
\label{tab:officehome1} 
\vspace{-3.5em}
\end{table*}

%% file: table/domainnet_semis_2.tex
\begin{table}[t]
\tablestyle{2pt}{1.1}
\newcolumntype{g}{>{\columncolor[gray]{0.9}} x{45}}
\resizebox{0.7\textwidth}{!}{
\begin{tabular}{x{6} y{53} ;{0.2pt/1.5pt} x{30} x{45} g ;{0.2pt/1.5pt} x{30} x{45} g}
&\multicolumn{1}{l}{feed.\,amount} & \multicolumn{3}{c}{378~{\tiny (3\,labeled data\,per\,class)}}                                     & \multicolumn{3}{c}{630~{\tiny (5\,labeled data\,per\,class)}}                                      \\
&method & {\scriptsize RF} & {\scriptsize NBF} & w/\,ours & {\scriptsize RF} & {\scriptsize NBF} & w/\,ours  \\
\shline
&\multicolumn{1}{l}{Source model}     & \multicolumn{6}{c}{56.5} \\
\cdashline{2-7}
% &FixMatch~\cite{fixmatch}  & 67.6 & 63.4\,{\tiny(-4.2)} & \textbf{73.2}\,{\tiny(+9.8)} & 71.5 & 66.1 & \textbf{75.1}  \\  
% &UDA~\cite{uda}       & 69.2 & 64.9 & \textbf{73.4} & 72.9 & 68.8 & \textbf{75.3}  \\
% &FlexMatch~\cite{flexmatch} & 73.3 & 71.4 & \textbf{74.7} & 75.3 & 73.9 & \textbf{76.0}  \\
% &FreeMatch~\cite{freematch} & 73.8 & 72.0 & \textbf{74.8} & 75.6 & 74.4 & \textbf{76.1}  \\
% \cdashline{2-7}
% & MME~\cite{mme}       & 69.5 & 68.4 & \textbf{70.8} & 71.2 & 70.1 & \textbf{72.5}  \\
% &CDAC~\cite{cdac}      & 68.3 & 64.6 & \textbf{73.2} & 71.7 & 68.1 & \textbf{74.9}  \\
% &AdaMatch~\cite{adamatch}  & 67.6 & 64.5 & \textbf{72.0} & 70.9 & 67.7 & \textbf{74.3}  \\
% \cdashline{2-7}
% &SHOT~\cite{shot}      & 69.6 & 70.7 & \textbf{71.5} & 71.1 & 72.3 & \textbf{73.0}  \\
% &NRC~\cite{nrc}        & 66.3 & 64.9 & \textbf{69.3} & 68.5 & 66.4 & \textbf{69.6}  \\
% &ContraTTA~\cite{contrastivetta} & 68.6 & 69.2 & \textbf{71.6} & 70.1 & 70.5 & \textbf{72.4}  \\
% &GuidingSP~\cite{guidingps}  & 69.7 & 70.2 & \textbf{71.8} & 70.5 & 71.0 & \textbf{73.2} \\
&MME~\cite{mme}&69.5&68.4\,{\tiny(-1.1)}&{70.8\,{\tiny(+2.4)}}&71.2&70.1\,{\tiny(-1.1)}&{72.5\,{\tiny(+2.4)}}\\
&CDAC~\cite{cdac}&68.3&64.6\,{\tiny(-3.7)}&{73.2\,{\tiny(+8.6)}}&71.7&68.1\,{\tiny(-3.6)}&{74.9\,{\tiny(+6.8)}}\\
&AdaMatch~\cite{adamatch}&67.6&64.5\,{\tiny(-3.1)}&{72.0\,{\tiny(+7.5)}}&70.9&67.7\,{\tiny(-3.2)}&{74.3\,{\tiny(+6.6)}}\\
\cdashline{2-7}
&FixMatch\cite{fixmatch}&67.6&63.4\,{\tiny(-4.2)}&{73.2\,{\tiny(+9.8)}}&71.5&66.1\,{\tiny(-5.4)}&{75.1\,{\tiny(+9.0)}}\\
&UDA~\cite{uda}&69.2&64.9\,{\tiny(-4.3)}&{73.4\,{\tiny(+8.5)}}&72.9&68.8\,{\tiny(-4.1)}&{75.3\,{\tiny(+6.5)}}\\
&FlexMatch~\cite{flexmatch}&73.3&71.4\,{\tiny(-1.9)}&{74.7\,{\tiny(+3.3)}}&75.3&73.9\,{\tiny(-1.4)}&{76.0\,{\tiny(+2.1)}}\\
&FreeMatch~\cite{freematch}&73.8&72.0\,{\tiny(-1.8)}&{\textbf{74.8}\,{\tiny(+2.8)}}&75.6&74.4\,{\tiny(-1.2)}&{\textbf{76.1}\,{\tiny(+1.7)}}\\
% \cdashline{2-7}
% &SHOT~\cite{shot}&69.6&70.7\,{\tiny(+1.1)}&{71.5\,{\tiny(+0.8)}}&71.1&72.3\,{\tiny(+1.2)}&{73.0\,{\tiny(+0.7)}}\\
% &NRC~\cite{nrc}&66.3&64.9\,{\tiny(-1.4)}&{69.3\,{\tiny(+4.4)}}&68.5&66.4\,{\tiny(-2.1)}&{69.6\,{\tiny(+3.2)}}\\
% &ContraTTA~\cite{contrastivetta}&68.6&69.2\,{\tiny(+0.6)}&{71.6\,{\tiny(+2.4)}}&70.1&70.5\,{\tiny(+0.4)}&{72.4\,{\tiny(+1.9)}}\\
% &GuidingSP~\cite{guidingps}&69.7&70.2\,{\tiny(+0.5)}&{71.8\,{\tiny(+1.6)}}&70.5&71.0\,{\tiny(+0.5)}&{73.2\,{\tiny(+2.2)}}\\

\cdashline{2-7}
\multirow{-9}{*}{\rotatebox[origin=c]{90}{ResNet-50~\cite{resnet}}} & \multicolumn{1}{l}{Fully supervised} & \multicolumn{6}{c}{83.6}  \\
% \cdashline{2-7}
% \\[-.5em]
\shlinesmall
&\multicolumn{1}{l}{Source model}    & \multicolumn{6}{c}{64.5} \\
\cdashline{2-7}
% &FixMatch~\cite{fixmatch}  & 74.6 & 73.0 & \textbf{75.6} & 75.7 & 74.3 & \textbf{76.5}  \\
% &UDA~\cite{uda}       & 74.8 & 73.3 & \textbf{75.8} & 75.9 & 74.5 & \textbf{76.7}  \\
% &FlexMatch~\cite{flexmatch} & 74.9 & 73.9 & \textbf{75.8} & 76.0 & 75.1 & \textbf{76.9}  \\
% &FreeMatch~\cite{freematch} & 74.9 & 73.9 & \textbf{75.7} & 76.0 & 75.1 & \textbf{76.8}  \\
% \cdashline{2-7}
% &MME~\cite{mme}       & 73.2 & 72.7 & \textbf{74.1} & 74.5 & 74.0 & \textbf{75.2}  \\
% &CDAC~\cite{cdac}      & 74.2 & 72.8 & \textbf{75.4} & 75.4 & 74.1 & \textbf{76.2}  \\
% &AdaMatch~\cite{adamatch}  & 74.7 & 73.7 & \textbf{75.9} & 75.9 & 75.1 & \textbf{76.7}  \\
% \cdashline{2-7}
% &SHOT~\cite{shot}                & 73.4 & 73.7 & \textbf{74.1} & 74.4 & 74.8 & \textbf{75.4} \\
% &NRC~\cite{nrc}                  & 72.2 & 71.9 & \textbf{72.9} & 73.9 & 73.7 & \textbf{74.6} \\
% &ContraTTA~\cite{contrastivetta} & 72.8 & 73.4 & \textbf{74.9} & 73.9 & 74.8 & \textbf{76.4} \\
% &GuidingSP~\cite{guidingps}      & 73.3 & 73.7 & \textbf{75.1} & 74.1 & 74.9 & \textbf{76.5} \\
&MME~\cite{mme}&73.2&72.7\,{\tiny(-0.5)}&{74.1\,{\tiny(+1.4)}}&74.5&74.0\,{\tiny(-0.5)}&{75.2\,{\tiny(+1.2)}}\\
&CDAC~\cite{cdac}&74.2&72.8\,{\tiny(-1.4)}&{75.4\,{\tiny(+2.6)}}        &75.4 &74.1\,{\tiny(-1.3)}&{76.2\,{\tiny(+2.1)}}\\
&AdaMatch~\cite{adamatch}&74.7&73.7\,{\tiny(-1.0)}&{\textbf{75.9}\,{\tiny(+2.2)}}&75.9 &75.1\,{\tiny(-0.8)}&{76.7\,{\tiny(+1.6)}}\\
\cdashline{2-7}
&FixMatch~\cite{fixmatch}&74.6&73.0\,{\tiny(-1.6)}&{75.6\,{\tiny(+2.6)}}&75.7&74.3\,{\tiny(-1.4)}&{76.5\,{\tiny(+2.2)}}\\
&UDA~\cite{uda}&74.8&73.3\,{\tiny(-1.5)}&{75.8\,{\tiny(+2.5)}}&75.9&74.5\,{\tiny(-1.4)}&{76.7\,{\tiny(+2.2)}}\\
&FlexMatch~\cite{flexmatch}&74.9&73.9\,{\tiny(-1.0)}&{75.8\,{\tiny(+1.9)}}&76.0&75.1\,{\tiny(-0.9)}&{\textbf{76.9}\,{\tiny(+1.8)}}\\
&FreeMatch~\cite{freematch}&74.9&73.9\,{\tiny(-1.0)}&{75.7\,{\tiny(+1.8)}}&76.0&75.1\,{\tiny(-0.9)}&{76.8\,{\tiny(+1.7)}}\\

% \cdashline{2-7}
% &SHOT~\cite{shot}&73.4&73.7\,{\tiny(+0.3)}&{74.1\,{\tiny(+0.4)}}&74.4&74.8\,{\tiny(+0.4)}&{75.4\,{\tiny(+0.6)}}\\
% &NRC~\cite{nrc}&72.2&71.9\,{\tiny(-0.3)}&{72.9\,{\tiny(+1.0)}}&73.9&73.7\,{\tiny(-0.2)}&{74.6\,{\tiny(+0.9)}}\\
% &ContraTTA~\cite{contrastivetta}&72.8&73.4\,{\tiny(+0.6)}&{74.9\,{\tiny(+1.5)}}&73.9&74.8\,{\tiny(+0.9)}&{76.4\,{\tiny(+1.6)}}\\
% &GuidingSP~\cite{guidingps}&73.3&73.7\,{\tiny(+0.4)}&{75.1\,{\tiny(+1.4)}}&74.1&74.9\,{\tiny(+0.8)}&{76.5\,{\tiny(+1.6)}}\\
\cdashline{2-7}
\multirow{-9}{*}{\rotatebox[origin=c]{90}{ViT-S~\cite{vit}}} &\multicolumn{1}{l}{Fully supervised} & \multicolumn{6}{c}{85.4}                                \\
% \hline
\end{tabular}}
\vspace{-.2em}
\caption{\textbf{Comparisons on DomainNet-126.} 
We evaluate our method by integrating it with \textit{SemiSDA} and \textit{SemiSL} methods. The average accuracy of seven domain-shift scenarios in \tabref{tab:domainnet1} is reported. Source model represents the pre-trained model without adaptation. Fully supervised means the model is adapted with fully labeled target data.}
\label{tab:domainnet2} 
\end{table}

%% file: table/officehome_semis_2.tex
\begin{table}[t]
\tablestyle{2pt}{1.1}
\newcolumntype{g}{>{\columncolor[gray]{0.9}} x{45}}
\vspace{-2.2em}
\resizebox{0.7\textwidth}{!}{
\begin{tabular}{y{53} ;{0.2pt/1.5pt} x{30} x{45} g ;{0.2pt/1.5pt} x{30} x{45} g}
\multicolumn{1}{l}{feed.\,amount} & \multicolumn{3}{c}{195~{\tiny (3\,labeled data\,per\,class)}}  & \multicolumn{3}{c}{325~{\tiny (5\,labeled data\,per\,class)}} \\
method & {\scriptsize RF} & {\scriptsize NBF} & w/\,ours & {\scriptsize RF} & {\scriptsize NBF} & w/\,ours  \\
\shline
\multicolumn{1}{l}{Source model}    & \multicolumn{6}{c}{57.6} \\
\hdashline
% FixMatch~\cite{fixmatch}   & 71.4 & 68.6 & 73.7 & 73.9 & 72.2 & 75.3 \\
% UDA~\cite{uda}        & 72.2 & 69.5 & 74.1 & 74.4 & 73.0 & 76.0 \\
% FlexMatch~\cite{flexmatch}  & 73.7 & 72.1 & 74.7 & 75.9 & 74.9 & 76.6 \\
% FreeMatch~\cite{freematch}  & 74.0 & 72.7 & \textbf{74.8} & 75.8 & 75.0 & \textbf{76.6} \\
% \hdashline
% MME~\cite{mme}        & 71.2 & 70.2 & 73.4 & 73.5 & 73.1 & 75.6 \\
% CDAC~\cite{cdac}       & 71.2 & 69.0 & 74.3 & 73.5 & 72.3 & 75.7 \\
% AdaMatch~\cite{adamatch}   & 70.9 & 69.3 & 73.8 & 73.4 & 72.7 & 75.5 \\
MME~\cite{mme}&71.2&70.2\,{\tiny(-1.0)}&73.4\,{\tiny(+3.2)}&73.5&73.1\,{\tiny(-0.4)}&75.6\,{\tiny(+2.5)}\\
CDAC~\cite{cdac}&71.2&69.0\,{\tiny(-2.2)}&74.3\,{\tiny(+5.3)}&73.5&72.3\,{\tiny(-1.2)}&75.7\,{\tiny(+3.4)}\\
AdaMatch~\cite{adamatch}&70.9&69.3\,{\tiny(-1.6)}&73.8\,{\tiny(+4.5)}&73.4&72.7\,{\tiny(-0.7)}&75.5\,{\tiny(+2.8)}\\
\hdashline
FixMatch~\cite{fixmatch}&71.4&68.6\,{\tiny(-2.8)}&73.7\,{\tiny(+5.1)}&73.9&72.2\,{\tiny(-1.7)}&75.3\,{\tiny(+3.1)}\\
UDA~\cite{uda}&72.2&69.5\,{\tiny(-2.7)}&74.1\,{\tiny(+4.6)}&74.4&73.0\,{\tiny(-1.4)}&76.0\,{\tiny(+3.0)}\\
FlexMatch~\cite{flexmatch}&73.7&72.1\,{\tiny(-1.6)}&74.7\,{\tiny(+2.6)}&75.9&74.9\,{\tiny(-1.0)}&\textbf{76.6}\,{\tiny(+1.7)}\\
FreeMatch~\cite{freematch}&74.0&72.7\,{\tiny(-1.3)}&{\textbf{74.8}\,{\tiny(+2.1)}}&75.8&75.0\,{\tiny(-0.8)}&{\textbf{76.6}\,{\tiny(+1.6)}}\\
\hdashline
\multicolumn{1}{l}{Fully supervised} & \multicolumn{6}{c}{87.4}                                
\end{tabular}}
\vspace{-.2em}
\caption{\textbf{Comparisons on OfficeHome.} The average accuracy of twelve domain-shift scenarios in \tabref{tab:officehome1} is reported. ResNet-50 is used.}
\label{tab:officehome2} 
\vspace{-3.em}
\end{table}
% All results of each domain-shift scenario are in the Appendix.

%% file: sec/6main_results.tex
\vspace{-.5em}
\subsection{Main Results}
\vspace{-.5em}\noindent
\textbf{Natural image classification.}
Following recent DA works~\cite{contrastivetta, sla}, we conduct experiments on seven and twelve domain shift scenarios provided with the DomainNet-126 and OfficeHome datasets, respectively. 
\tabref{tab:domainnet1} and \tabref{tab:officehome1} show the results, where AdaMatch~\cite{adamatch} is used as the baseline.
We observe consistent results with \figref{fig:toy_exper} \textit{even on} large natural datasets: when simply applying the baseline under the NBF assumption, the adapted model shows inferior performance for most domain shifts than applying it under RF, \eg, $64.5{<}67.6$. 
Combining our approach with the baseline mitigates this issue and achieves a performance increase, \eg, $64.5$→$72.0$. 

We also use other promising baselines and report the average accuracy of all domain shifts in \tabref{tab:domainnet2} and \tabref{tab:officehome2} (all results can be found in Appendix~\ref{sec:allshift}). 
While both feedback types bring performance improvement from the source model, lower performance is observed with NBF. 
Our method enables the baselines to not only address this problem but surpass performance under RF. 
The above results suggest that the biased distribution of labeled samples, which has been overlooked in previous SemiSDA works, is actually problematic, and our retrieval latent defending approach is effective.

\input{table/mimic_semis}
\input{table/segmen_semis}

\vspace{.3em}\noindent
\textbf{Medical image diagnosis.}
\tabref{tab:mimic} shows the results (bottom) and also depicts the effect of NBF (top center).
We report the AUROC~\cite{aucscore} for each finding following standard practice for measuring computer-aided-diagnosis model evaluation~\cite{lenga2020continual, mahapatra2022unsupervised}.
The baseline SemiSDA method under NBF exhibits inferior performance compared to one under RF, but this issue can be mitigated by combining our approach.

\vspace{-0.2em}
In addition, we propose an interesting and practical scenario named NBF with more confident errors (\textbf{NBF-CE}). In this scenario, we assume that a radiologist is likely to give feedback when the model makes \textit{confidently wrong} predictions. 
Imagine that the model predicts a 1\% likelihood of cancer in a CXR image, but the person actually has cancer.
Such failure to detect potential patients early on can significantly reduce the patient's chances of survival, so a radiologist may provide feedback to the model. 
To simulate NBF-CE, we select samples where the source model most confidently predicts a finding to be absent ($\hat{y}{\approx}0$) although it is clearly visible in the radiograph ($y{=}1$), and vice versa, \ie, samples of $\hat{y}{\approx}1$ but $y{=}0$. 
\tabref{tab:mimic} also shows the results under an NBF-CE scenario, where the model's adaptation performance is further reduced compared with NBF (0.7691 for NBF\,→\,0.7639 for NBF-CE).
By combining our method, we observe performance improvements for both NBF variants, \eg, 0.7639 for NBF-CE\,→\,0.7875 with ours. We illustrate the hypothesized impact of our method in \tabref{tab:mimic}.

\vspace{.3em}\noindent
\textbf{Semantic segmentation.} We evaluate the influence of NBF and our approach on a semantic segmentation task. We utilize the most common adaptation benchmark of GTA5~\cite{gta5} to Cityscapes~\cite{cityscapes}. The baseline DA algorithms are used as IAST~\cite{iast,labor} and RIPU~\cite{ripu} in a source-free scenario. We regard Pixel-based Annotation\,(PA) in which we assume 40 pixels per image like LabOR~\cite{labor}. \tabref{tab:cityscape} shows results similar to those we observed in the classification and medical imaging tasks. The baselines under NBF exhibit inferior performance compared to those under RF\,(54.5 for NBF${<}$57.6 for RF), but this issue is addressed by combining our approach with them (+3.5 mIoU). Although out of our scope (refer to Appendix~\ref{sec:activeda}), we validate one active labeling strategy ENT\,\cite{shen2017entropy}, which assigns highly uncertain (\ie, \textit{probably misclassified}) pixels as feedback. Consequently, the feedback instructed by ENT is \textit{biasedly} distributed in a manner similar to NBF. ENT also causes unexpected results\,(54.6 for ENT${<}$57.6 for RF), and our approach alleviates this issue (+3.1 mIoU). 

%% file: table/mimic_semis.tex
\begin{figure*}[t]\centering
\includegraphics[width=0.93\linewidth]{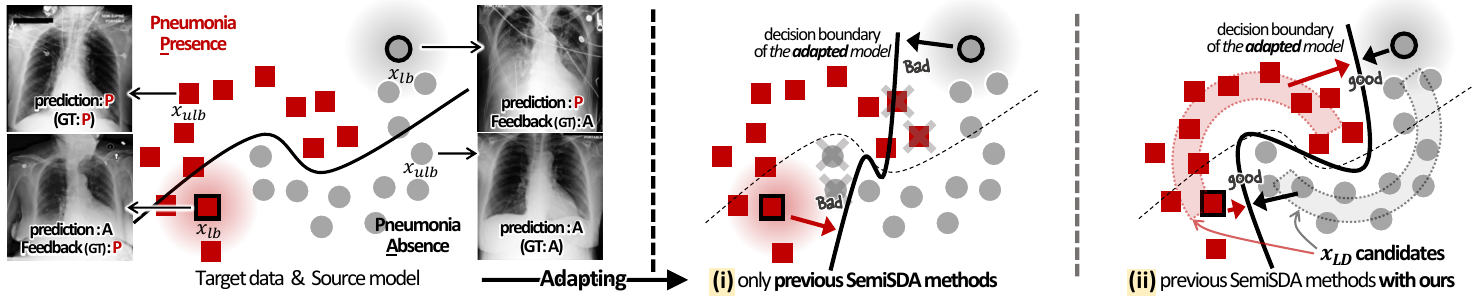}
\vspace{-1.5em}
\end{figure*}

\begin{table*}[t]
\tablestyle{1pt}{1.15}
\resizebox{\textwidth}{!}{
\begin{tabular}{y{42} x{35} x{42}  ;{.5pt/1.pt} x{27} x{27} x{27} x{27} x{27} x{27} x{27} x{27} x{27} x{27} x{27} x{27} x{27} x{27}}
method           & feedback & average & \rotatebox[origin=c]{40}{atelectasis}  & \rotatebox[origin=c]{40}{cardiomegaly} & \rotatebox[origin=c]{40}{consolidation}  & \rotatebox[origin=c]{40}{edema}    & \rotatebox[origin=c]{40}{enlarged cardio.}  & \rotatebox[origin=c]{40}{fracture}  & \rotatebox[origin=c]{40}{lung lesion}  & \rotatebox[origin=c]{40}{lung opacity} & \rotatebox[origin=c]{40}{effusion} & \rotatebox[origin=c]{40}{pleural}  & \rotatebox[origin=c]{40}{pneumonia}  & \rotatebox[origin=c]{40}{pneumothorax}  & \rotatebox[origin=c]{40}{support device}  \\
\shline
Source mo.           &          & .7738  & .7784  & .7919  & .8236  & .8500  & .7646  & .6642  & .7555  & .7818  & .8271  & .8288  & .7535  & .6894  & .7500  \\
\hdashline
PseudoL~\cite{arazo2020pseudo}    & RF       & .7850  & .7828  & .7965  & .8453  & .8615  & .7639  & .6832  & .7598  & .7947  & .8333  & .8565  & .7702  & .6957  & .7622  \\
\hdashline
                                        & NBF      & .7691  & .7719  & .7851  & .8202  & .8468  & .7403  & .6934  & .7446  & .7809  & .8070  & .8260  & .7521  & .6979  & .7324  \\
 & gap & \textcolor{darkredd}{-.0159} & -.0109 & -.0114 & -.0252 & -.0147 & -.0236 & +.0102  & -.0152 & -.0138 & -.0262 & -.0304 & -.0181 & +.0022  & -.0298 \\
\gr
w/\,ours           &    NBF      & \textbf{.7884}  & .7895  & .7956  & .8515  & .8606  & .7730  & .6821  & .7599  & .7973  & .8445  & .8611  & .7753  & .6851  & .7736  \\
\gr
& gain  & \textcolor{darkblue}{+.0193} & +.0176 & +.0105 & +.0313 & +.0138 & +.0326 & -.0113 & +.0153 & +.0164 & +.0375 & +.0351 & +.0232 & -.0128 & +.0412 \\
\hdashline
   & NBF-CE   & .7639  & .7682  & .7834  & .8124  & .8418  & .7403  & .6808  & .7472  & .7744  & .8005  & .8199  & .7469  & .6879  & .7277  \\
 & gap & \textcolor{darkredd}{-.0211} & -.0146 & -.0131 & -.0330 & -.0198 & -.0236 & -.0024 & -.0126 & -.0203 & -.0328 & -.0366 & -.0233 & -.0079 & -.0344 \\
\gr
w/\,ours           &  NBF-CE        & .7875  & .7895  & .7956  & .8515  & .8606  & .7730  & .6731  & .7599  & .7973  & .8445  & .8611  & .7753  & .6831  & .7736  \\
\gr
& gain  & \textcolor{darkblue}{+.0236} & +.0213 & +.0122 & +.0391 & +.0189 & +.0327 & -.0077 & +.0126 & +.0229 & +.0440 & +.0412 & +.0284 & -.0048 & +.0459 \\
\hdashline
Fully super. &          & .8117  & .8150  & .8277  & .8758  & .8820  & .7984  & .6949  & .7750  & .8200  & .8725  & .8441  & .8044  & .7398  & .8025 
\end{tabular}}
\vspace{+.1em}
\caption{\textbf{Adaptation in a medical application.} We use samples with PA-view as the source data and samples with AP-view as the target data in MIMIC-CXR-V2 dataset~\cite{mimic}. 
% The experimental setup is the same as in \tabref{tab:domainnet1}.
% The target adaptation results of thirteen findings are reported as the AUC score. 
NBF-CE represents a scenario when NBF is composed of cases with confident errors.
%means the feedback set obtained by radiologists when a model makes seriously wrong predictions (details are in \secref{sec:medicalimaging}). 
We use DensNet-121~\cite{densenet, torchxray} and assume the 20 feedback for the absence and presence per finding.}
\label{tab:mimic} 
\vspace{-3.5em}
\end{table*}

%% file: table/segmen_semis.tex
% {\scriptsize $\times$} $\circ$
\begin{table*}[t]
\tablestyle{1.5pt}{1.15}
\newcolumntype{g}{>{\columncolor[gray]{0.9}} x{55}}
\resizebox{0.99\textwidth}{!}{
\begin{tabular}{y{60} x{60} ;{1.5pt/1.pt} x{45} ;{1.5pt/1.pt} x{45} g | x{45} x{55} }
labeling type & feed. amount & RF & NBF & NBF w/\,ours & ENT~\cite{shen2017entropy} & ENT w/\,ours \\
\shlinesmall
IAST~\cite{iast, labor} & PA, 40 points & 55.3 & 53.0\,{\tiny(-2.3)} & \textbf{56.3}\,{\tiny(+3.3)} & 53.5 & 56.0\,{\tiny(+2.5)}  \\
RIPU~\cite{ripu} & PA, 40 points & 57.6 & 54.5\,{\tiny(-3.1)} & \textbf{58.0}\,{\tiny(+3.5)} & 54.6 & 57.7\,{\tiny(+3.1)}  \\
\end{tabular}}
\vspace{+.1em}
\caption{\textbf{Adaptation on semantic segmentation.} The GTA5\,\cite{gta5} → Cityscapes\,\cite{cityscapes} setup is used~\cite{Tsai_adaptseg_2018}. The target performance of the source model is 36.6 mIoU.}
\label{tab:cityscape} 
\vspace{-3.3em}
\end{table*}
% We also report the performance of the baseline without our approach (rightmost).

%% file: sec/7ablations.tex
\vspace{-.5em}
\subsection{Ablation Study}\blfootnote{If not specified, we use ResNet-50 and report the average accuracy (\%) of seven domain shift scenarios in \tabref{tab:domainnet1} for ablation studies.\vspace{-2.em}}
\label{sec:ablation}

\vspace{-1.5em}
\paragraph{Positive vs. Negative feedback.} 
We study the role of feedback on the adaptation results by varying feedback configurations.
Let positively-provided feedback (PF) be obtained from samples that the source model \textit{correctly} predicts, as opposed to negatively-provided feedback (NF).
We adjust the ratio of PF:NF while keeping the total number of labeled samples constant, as shown in \figref{fig:feedbacktype}.

When using only FreeMatch (gray dot-dashed line), both biased feedback types (\ie, NBF and PBF) result in worse adaptation performance compared to balanced feedback for the baseline, \eg, 72.6 in 378:0{\small \,(PBF)}\,${<}$\,73.3 in 252:126. 
In contrast, when our method is applied (red line), NBF yields the best performance. 
PBF and NBF can be respectively regarded as contributing previously \textit{known} knowledge of the model and \textit{new} knowledge that complements model deficiencies. Hence, it may be natural that NBF, which actually encodes the model's mistakes, contributes to favorable adaptation results.

\vspace{.3em}\noindent
\textbf{Number of unlabeled samples in a mini-batch.}
Existing SemiSDA methods~\cite{adamatch, freematch} typically set the ratio $\mu$ between labeled and unlabeled samples in a mini-batch to 1:7. However, we observe that adhering to this ratio is not optimal for our approach, as shown in \tabref{tab:curriculum}.
Our method shows better performance when the ratio is varied to 1:4, \ie, decreasing unlabeled sample sizes.
This finding contradicts observations in several TTA works~\cite{niu2023towards,khurana2021sita,ecotta}, where adaptation performance tends to increase with larger batch sizes.
We speculate that it is beneficial to prioritize more reliable information, which refers to labeled data and our defending samples selected from the \textit{filtering}-applied bank, during the adapting process.
This result may be aligned with previous works for curriculum learning~\cite{zhang2017curriculum, liu2020open} and adaptive thresholding~\cite{flexmatch}.

\input{fig/feedbacktypes}
\input{table/ablation_curriculum}

\vspace{.3em}\noindent
\textbf{Number of labeled data.}
We measure the impact of feedback size (number of labeled samples) in \figref{fig:num_label}.
The results show that the inferior performance on NBF persists even with an increased amount of feedback (gray\,→\,black line); however, our approach mitigates it and improves performance (black\,→\,red line). 
We make an interesting observation that the performance gap between black and red lines becomes larger as the number of available feedback decreases. Since obtaining large feedback may be challenging in real-world applications, our method is expected to be more helpful in this practical case.

\input{fig/num_label}
\input{table/ablation_selection}

\vspace{.3em}\noindent
\textbf{Data selecting strategy.}
We explore various strategies for selecting defending samples to balance the mini-batch, as shown in Table~\ref{tab:selection}\,(top).
The strategies include: in the $x_{LD}$ candidate bank, 
{\small (i)} random selection regardless of the class of the labeled data, 
{\small (ii)} random selection in the same class as the labeled data (\ie, class-aware), 
{\small (iii)} selecting samples close to the cluster center obtained by k means clustering~\cite{kmeans} and
{\small (iv)} selecting samples with embedded features distant from the labeled data where cosine distance is used. 
While our approach consistently outperforms the baseline regardless of the chosen strategy, we empirically find that strategy\,{\small (ii)} achieves the best performance. Therefore, we adopt this strategy for our proposed method.

Further studies, such as extension to a TTA scenario, combining with SFDA methods and different feedback configurations, are presented in Appendix~\ref{sec:addabl}.

%% file: fig/feedbacktypes.tex
\begin{figure}[t]\centering
\includegraphics[width=0.55\linewidth]{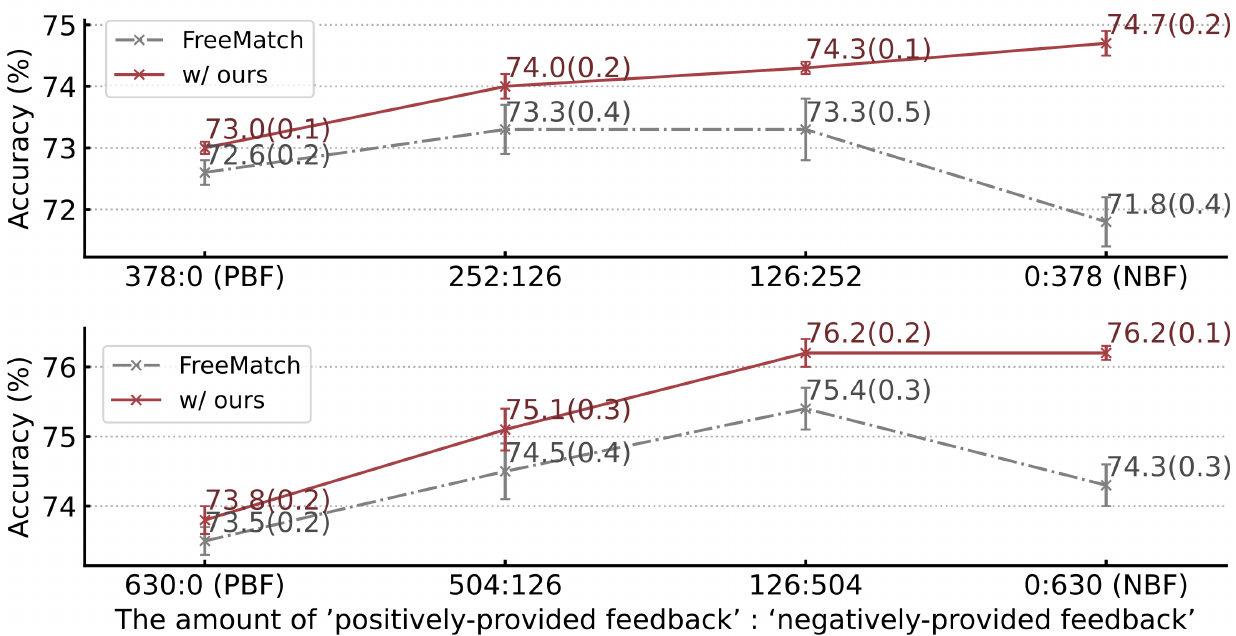}
\vspace{-1.em}
\caption{\textbf{NBF leads to higher performance than PBF.} We compare different user-feedback configurations when the total number of feedback is 378 (top) and 630 (bottom). Positive and negative feedback refers to feedback from correct and incorrect model predictions, respectively. We run three random seed experiments and describe the average performance and standard deviation in the parenthesis.}
\label{fig:feedbacktype}
\end{figure}

% Our approach enables a baseline to effectively leverage feedback (particularly with NBF).

%% file: table/ablation_curriculum.tex
% \begin{table}[t]
% \tablestyle{2pt}{1.15}
% \newcolumntype{g}[1]{>{\columncolor[gray]{0.9}} x{#1}}
% \begin{tabular}{y{32} x{15} ;{0.2pt/1.5pt} x{25} x{25} x{30}}
% method & feed. & \multicolumn{3}{c}{\cellcolor{white}negatively biased feedback}   \\
% CaB & &  & {\tiny \cmark} &  {\tiny \cmark}\\
% Curric. & & & & {\tiny \cmark} \\
% \shlinesmall
% FreeMatch &   & 72.0 & 74.2 & \baseline{\textbf{74.8}\,{\tiny(+0.6)}} \\
% AdaMatch  & \multirow{-2}{*}{\cellcolor{white}368}& 64.5 & 71.3 & \baseline{72.0\,{\tiny(+0.7)}}\\
% \hdashline
% FreeMatch &   & 74.4 & 75.5 & \baseline{\textbf{76.1}\,{\tiny(+0.6)}} \\
% AdaMatch  & \multirow{-2}{*}{\cellcolor{white}630}& 67.7 & 73.4 & \baseline{74.3\,{\tiny(+0.9)}}
% \end{tabular}
% \label{tab:curriculum} \vspace{-1.em}
% \caption{Ablation study of curriculum adaptation}
% \label{tab:curriculum} \vspace{-.5em}
% \end{table}

\begin{table}[t]
\tablestyle{0.7pt}{1.15}
\newcolumntype{g}[1]{>{\columncolor[gray]{0.9}} x{#1}}
\vspace{-1.em}
\resizebox{0.55\textwidth}{!}{
\begin{tabular}{y{70} x{30} ;{0.2pt/1.5pt} x{50} x{50} x{50}}
method & feed. & \multicolumn{3}{c}{\cellcolor{white}negatively biased feedback (NBF)}   \\
% w/\,ours & &  & {\tiny \cmark} &  \baseline{{\tiny \cmark}}\\
{\tiny \#\,$x_{ulb}$,\,\#\,$x_{LD}$,\,\#\,$x_{lb}$} & &  112, 0, 16 & 112, 48, 16 & \baseline{64, 48, 16}\\
total batch size & &  128 & 176 & \baseline{128}\\
\shlinesmall
FreeMatch~\cite{freematch} &   & 72.0 & 74.2 & \baseline{\textbf{74.8}\,{\tiny(+0.6)}} \\
AdaMatch~\cite{adamatch}  & \multirow{-2}{*}{\cellcolor{white}368}& 64.5 & 71.3 & \baseline{72.0\,{\tiny(+0.7)}}\\
\hdashline
FreeMatch~\cite{freematch}  &   & 74.4 & 75.5 & \baseline{\textbf{76.1}\,{\tiny(+0.6)}} \\
AdaMatch~\cite{adamatch}   & \multirow{-2}{*}{\cellcolor{white}630}& 67.7 & 73.4 & \baseline{74.3\,{\tiny(+0.9)}}
\end{tabular}}
\vspace{+.1em}
\caption{In the mini-batch, diminishing the number of unlabeled samples and adding our defending samples achieves better performance with our approach. We ablate them by changing the ratio $\mu$ in \secref{sec:prerequ}, while keeping the size of labeled samples.}
\label{tab:curriculum}
\vspace{-3.5em}
\end{table}

%% file: fig/num_label.tex
\begin{figure}[t]\centering
\includegraphics[width=0.6\linewidth]{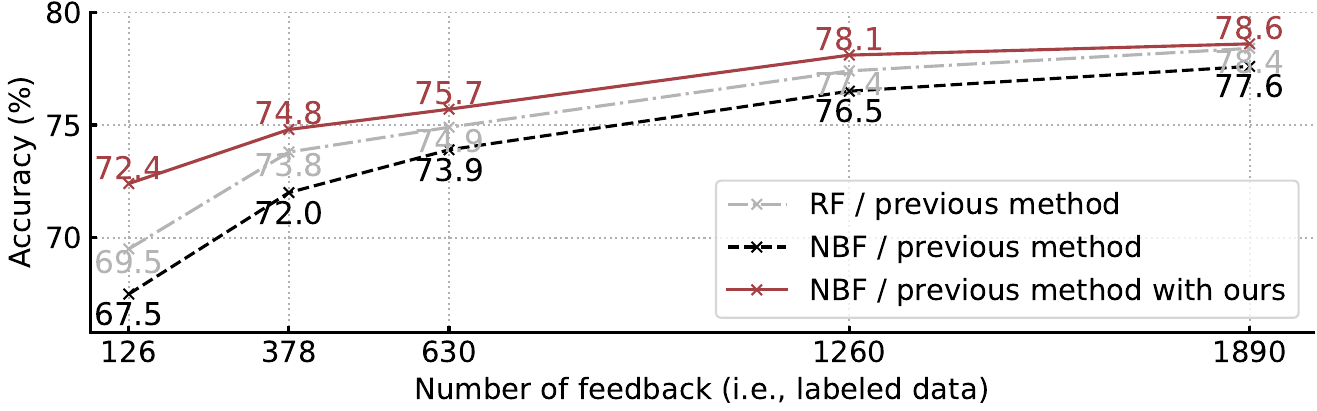}
\vspace{-1.em}
\caption{\textbf{More reliable adaptation with NBF.} In addition to \tabref{tab:domainnet2}, we conduct experiments with different amounts (1,3,5,10, and 15 labeled data per class) of feedback using FreeMatch~\cite{freematch}. The number of available feedback is likely to be small in practice. In this case, our method achieves large performance improvement, \eg, our method increases the baseline performance by $+4.9$ when one feedback per class is available.}
\label{fig:num_label}
\end{figure}
% \caption{\textbf{More efficient adaptation with NBF.} In addition to \tabref{tab:domainnet2}, we conduct experiments with different amounts (1,3,5,10, and 15 labeled data per class) of feedback using FreeMatch. 
% We show that the performance gap between our approach and the baselines are larger when the number of available feedback is smaller.}

%% file: table/ablation_selection.tex
% {\scriptsize $\times$} $\circ$
\begin{table}[t]
\vspace{-1.em}
\tablestyle{1.5pt}{1.15}
\newcolumntype{g}{>{\columncolor[gray]{0.9}} x{40}}
\resizebox{0.75\textwidth}{!}{
\begin{tabular}{y{70}  x{25} ;{0.2pt/1.5pt} x{40} g x{40}x{40} ;{0.2pt/1.5pt} x{50}}
selection strategy & & random & \baseline{random} & kmeans & cosine & baseline \\
class-aware &  & {\tiny \xmark} & \baseline{{\tiny \cmark}} & {\tiny \cmark} & {\tiny \cmark} & -\\
\shlinesmall
FreeMatch~\cite{freematch} & Res. & 74.1 & \baseline{\textbf{74.8}} &74.6 &  74.0&72.0 \\
% \multirow{-2}{*}{\rotatebox[origin=c]{90}{Res.}}&AdaMatch~\cite{adamatch} & 69.1 & \baseline{72.0} & 71.6 & 71.3 & 64.5\\
FreeMatch~\cite{freematch} & ViT. & 75.0 & \baseline{\textbf{75.7}} &75.6 &  75.1 &73.9 \\
% \multirow{-2}{*}{\rotatebox[origin=c]{90}{ViT.}}&AdaMatch~\cite{adamatch} & 74.3  & \baseline{75.9} & 75.7& 75.2& 73.7\\
\\[-.8em]
filtering rate  & & 0.2 & 0.4 & 0.6 & 0.8  & baseline only\\
\shlinesmall
FreeMatch~\cite{freematch}  & Res. & 74.5 & \textbf{74.8} & 74.3 & 73.7 & 72.0 \\
FreeMatch~\cite{freematch}  & ViT. & 75.5 & 75.7 & \textbf{75.9}  & 75.5 & 73.9 \\
\end{tabular}}
\vspace{+.1em}
\caption{We ablate a component of our approach with 378 feedback: $x_{LD}$ selection strategy and filtering rate $p$ for bank generation.}
\vspace{-3.5em}
\label{tab:selection} 
\end{table}
% We also report the performance of the baseline without our approach (rightmost).

%% file: sec/8conclusion.tex
User feedback can play an integral part in adapting the practical ML product to the target environment.
However, we have shown that naive adaptation using existing SemiSDA methods led to undesirable adaptation results.
We explained this through the lens of Negatively-Biased Feedback (NBF).
In this paper, we uncovered the unexpected results of NBF and presented a scalable solution, \textit{Retrieval Latent Defending}. This method prevents the mini-batch from becoming overly dependent on labeled samples that may have a biased distribution within the overall target distribution.
Under the diverse DA benchmarks, from the simulation study to the medical imaging task, we demonstrated the practical problem caused by NBF and the effectiveness of our approach by combining it with multiple SemiSDA baselines. 
We hope our efforts will inspire future DA works on leveraging user feedback to improve an ML model in the deployment environment.

\vspace{.1em}\noindent
\textbf{Broader impact.} The proposed setup assumes that an ML product obtains feedback as a form of annotations (\ie, labeled data). In some cases, users can provide feedback in different forms, like thumbs up \& down and rating of model prediction, or noise feedback whose information is different from the ground truth.
Further research considering these points will pave the way for developing safer and more reliable adapting strategies.

\vspace{.1em}\noindent
\textbf{Acknowledgment.} 
We sincerely appreciate the abundant support provided by Lunit Inc., and we would like to thank Donggeun Yoo, Seonwook Park, and Sérgio Pereira for their valuable feedback.

%% file: sec/appendix.tex
\title{Supplementary Material on\\Is user feedback always informative? \\Retrieval Latent Defending for Semi-Supervised Domain Adaptation without Source Data} 

\titlerunning{NBF \& RLD}

\author{Junha Song\inst{1,2} \and
Tae Soo Kim\inst{2} \and
Junha Kim\inst{2} \and
Gunhee Nam\inst{2} \and
\\
Thijs Kooi\inst{2} \and
Jaegul Choo\inst{1} 
}

\authorrunning{J. Song et al.}

\institute{
KAIST \email{\{sb020518, jchoo\}@kaist.ac.kr} \and
Lunit Inc. \email{\{junha.kim, taesoo.kim, ghnam, tkooi\}@lunit.io}
}

\maketitle
\setcounter{figure}{6}
\setcounter{table}{8}
\setcounter{equation}{2}

\def\thesection{\Alph{section}}
\setcounter{section}{0}

\newcommand{\figrefm}[1]{Figure~\ref{#1}}
\newcommand{\tabrefm}[1]{Table~\ref{#1}}
\newcommand{\equrefm}[1]{Eq.~(\ref{#1})}
\newcommand{\secrefm}[1]{Section~\ref{#1}}
\newcommand{\figrefs}[1]{Figure~\ref{#1}}
\newcommand{\tabrefs}[1]{Table~\ref{#1}}
\newcommand{\equrefs}[1]{Eq.~(\ref{#1})}
\newcommand{\secrefs}[1]{Section~\ref{#1}}

\noindent In this supplementary material, we provide:
\vspace{-.2em}
\begin{enumerate}[label=\Alph*.]
    \setlength\itemsep{+.5em}
    \item[\ref{sec:addrelated}.] Comparison with Related Work
        \begin{enumerate}[label=A.\arabic*]
        \setlength\itemsep{+.2em}
        \item[\ref{sec:activeda}.] Active Domain Adaptation
        \item[\ref{sec:classim}.] Class-Imbalanced Semi-Supervised Learning
        \item[\ref{sec:tta}.] Test-time Adaptation
        \item[\ref{sec:learnfeedback}.] Learning with User Feedback
        \end{enumerate}
    \item[\ref{sec:addsimulation}.] Further understanding with Simulation Study
    \item[\ref{sec:addabl}.] Additional Ablation Study
    \item[\ref{sec:adddetail}.] Additional Experimental Details
    \item[\ref{sec:adddisc}.] Additional Discussion
        \begin{enumerate}[label=E.\arabic*]
        \setlength\itemsep{+.2em}
        \item[\ref{sec:novel}.] Technique novelty
        \item[\ref{sec:efficiency}.] Computational overhead
        \item[\ref{sec:limitation}.] Limitations
        \end{enumerate}
    \item[\ref{sec:allshift}.] Results of All Domain Shifts.
\end{enumerate}

% \noindent
% The table or figure in the main paper and the supplementary are denoted by the superscript \bm{$_{m}$} and \bm{$_{s}$}, respectively.

\section{Comparison with Related Work}\label{sec:addrelated}
\vspace{-.5em}
\subsection{Active Domain Adaptation}\label{sec:activeda}
\vspace{-.3em}

\begin{figure}[t!]\centering
\vspace{-1.5em}
\includegraphics[width=0.9\linewidth]{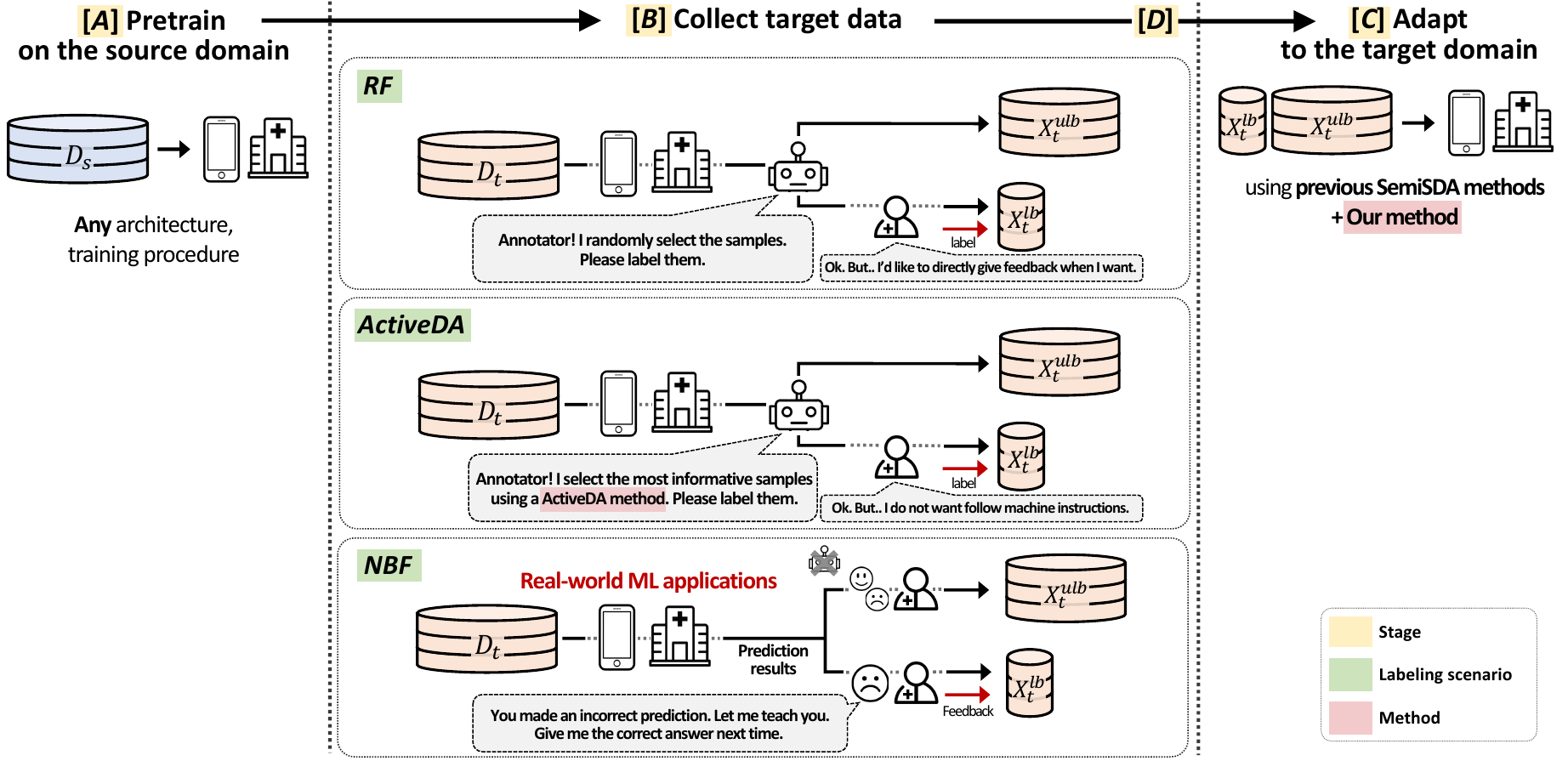}
\vspace{-1.em}
\caption{Comparison between labeling scenarios: random feedback\,(RF), active domain adaptation\,(ActiveDA), and negatively biased feedback\,(NBF).}
\label{fig:activeda}
\vspace{-1.em}
\end{figure}

Active domain adaptation (ActiveDA) aims to select the most informative samples being labeled by annotators, given a limited annotating budget.
As shown in \figrefs{fig:activeda}, the machine selects some samples using ActiveDA methods and instructs annotators to label the selected samples. 
Several ActiveDA methods have been proposed, such as CLUE~\cite{clue}, which employs an entropy-based clustering algorithm to preserve the uncertainty and diversity of labeled data. SDM-AG~\cite{sdm} and DiaNa~\cite{diana} utilize margin functions between the source and target domains to identify informative samples. 
In contrast to this ActiveDA scenario, we present an NBF scenario where there is no machine-instructed sample selection, and instead, users \textit{directly} provide feedback as a response to the prediction result. It may lead to more flexible applications since {\small (1)} users have the freedom to choose samples, and {\small (2)} individual users can impose different standards in selecting samples.

% , which may be more prevalent in real-world applications. Here, we assume that there is no machine-instructed sample selection. Instead, \textit{users} directly provide feedback as a response to the prediction result. 
% users have the freedom to choose the samples, which may lead to more flexible adaptation scenarios since individual users may impose different standards in selecting samples.

\footnotetext[1]{\label{note1}{\scriptsize 
For DiaNA~\cite{diana}, we utilize their proposed `informativeness scoring mechanism' to maintain a pretrained model-agnostic property.}}

\footnotetext[2]{\label{note2}{\scriptsize 
If not specified, we use ResNet-50 and report the average accuracy (\%) of seven domain shift scenarios in \tabrefm{tab:domainnet1} for additional studies.}}

We note that ActiveDA methods are for \colorbox{highlight!85}{\makebox(32,6){\strut\textcolor{black}{stage\,B}}} of \figrefs{fig:activeda}, while our method is for \colorbox{highlight!85}{\makebox(32,6){\strut\textcolor{black}{stage\,C}}} and proposed to alleviate the problem caused by NBF.
Although out of our scope, we evaluate our method under ActiveDA labeling scenarios, where CLUE and DiaNA\textsuperscript{1} are employed. 
The results in \tabrefs{tab:activeda} suggest two points. First, our method complements existing ActiveDA methods, consistently improving their performance. This highlights the importance of adapting the model with a balanced supervised signal throughout adaptation (\ie, stage C) using our method, even when ActiveDA methods like CLUE respect the diversity of labeled samples. Second, our method achieves significant performance gains regardless of the labeling scenario, showing that our method can be applied for reliable adaptation even when the distribution of labeled data is unknown.
% \colorbox[HTML]{F4CBCD}{\makebox(50,6){\strut\textcolor{black}{

\input{table/activeda}

\subsection{Class-Imbalanced Semi-Supervised Learning}\label{sec:classim}

SemiSDA and SemiSL methods often struggle with the different numbers of labeled data \textit{\textbf{between classes}}, known as class imbalance~\cite{oh2022daso}. To address this problem, class-imbalanced SemiSL works like CReST~\cite{wei2021crest} propose to balance the quantity of labeled data by using pseudo labels~\cite{kim2020distribution, lee2021abc} in stage\,D (\ie, generation in CReST) of \figrefs{fig:activeda}. Recent advancements like DASO~\cite{oh2022daso} further reduce the imbalance effect using both a similarity-based and linear classifier. 
Despite such advances in class-imbalanced SemiSL, the biased (\ie, imbalanced) label distribution \textit{\textbf{within the same class}} has been \textcolor{darkred}{overlooked in the SemiSDA, SemiSL, and class-imbalanced SemiSL works}. Therefore, we introduce the new concept of biased labeled data called NBF and demonstrate its unexpected
influence on adaptation performance. 

Even though our focus in this paper is on the bias within the same class, accounting for the imbalance between classes can still be crucial for reliable domain adaptation. For example, in the medical domain, while radiologists are likely to log the mistakes of the model, the amount of feedback from false negative samples may be small compared to those from false positive samples, given the natural prevalence of disease (e.g., lung cancer is less than 1 in 1000). We simulate this example scenario and evaluate our method in \tabrefs{tab:mimic2}.

\input{table/mimic_semis2}

\paragraph{Under different feedback configurations.} 
We take various feedback configurations into account, as depicted in \tabrefs{tab:mimic2}.
Assuming the model acquires 80 or 160 feedback instances for each finding, we alter the feedback quantities from false positive (FP) and false negative (FN) errors, which is similar to the setup of class-imbalanced SemiSL~\cite{wei2021crest, kim2020distribution}. We only consider two radiographic findings for simplification.
The results show that our method can also mitigate the intended impact of NBF even with the class-imbalanced scenario.  
Interestingly, we observe better performance when FP feedback is larger than FN feedback, which makes our method suitable for the medical domain, where radiographic findings are rarely detected due to the natural prevalence of the disease. 

\begin{figure}[t!]\centering
\includegraphics[width=0.99\linewidth]{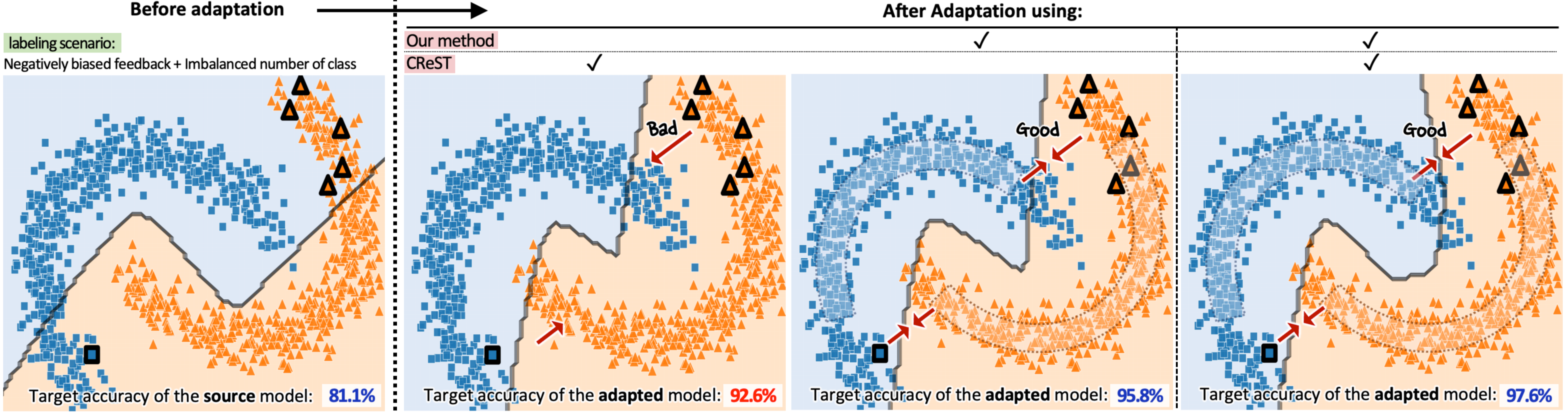}
\vspace{-1.em}
\caption{Our contribution focuses on introducing NBF and analyzing the effect of NBF on adaptation. However, it may be required to consider both an NBF and a class-imbalanced scenario in real-world applications. Hence, we first simulate this scenario and perform adaptation using i) a SemiSDA method (\ie, Pseudo labeling~\cite{arazo2020pseudo}) with our method and ii) a class-imbalanced SemiSL method (\ie, CReST~\cite{wei2021crest}).}
\label{fig:crest}
\vspace{-1.5em}
\end{figure}

\paragraph{Combining with class-imbalanced SemiSL methods.}
One naive way to more reliably adapt to the challenging scenario could involve combining our method with class-imbalanced SemiSL methods in stages C and D of Figure~\ref{fig:activeda}. To evaluate this approach, we conduct an additional simulation study in Figure~\ref{fig:crest}. 
The simulation replicates the NBF scenario by selecting only misclassified samples within the same class. We further introduce class imbalance by varying the number of feedback points between the blue and orange classes (leftmost sub-figure).

By adapting the model with different approaches, we can find two interesting takeaways: {\small (1)} the approach proposed in CReST~\cite{wei2021crest} was not designed to solve the unexpected effect of NBF, so it struggles with adaptation under the challenging scenario. {\small (2)} our method achieves better adaptation performance than using only CReST, and outperforms other results by combining with CReST. These results highlight the importance of considering an NBF case as well as a class-imbalance problem and the efficiency of our method. We hypothesize that defending the latent class space throughout adapting iterations helps the model to be robust to the effect of NBF, different from a previous generation-based approach in CReST~\cite{wei2021crest}.
In addition, a discussion about zero feedback for certain classes is provided in \secrefs{sec:addabl}.
% while previous class-imbalanced SemiSL methods periodically balancing the number of labeled data

\vspace{-.5em}
\subsection{Test-time Adaptation}\label{sec:tta}
\vspace{-.5em}

% Pre-training the model with all potential target domains is difficult and expensive. 
% This issue has led to the development of a new setting called as test-time adaptation (TTA), in which the adaptation is performed using the unlabeled target sample encountered by the model after deployment, regardless of the source domain on which the model was pre-trained.

To mitigate performance degradation caused by domain shift, models deployed on edge devices like smartphones and self-driving cars can be adapted to the target domain in an online manner, referred to as test-time adaptation (TTA). TTA assumes two practical settings: {\small i)} adapting without source data and {\small ii)} storing a limited amount of unlabeled target data. 
% Several TTA methods~\cite{eata, cotta, ecotta} address these constraints. 
For instance, TENT~\cite{wang2020tent, eata, ecotta, cdtta} leverages the current batch of unlabeled data to update the model's batch normalization parameters. Alternatively, methods like NOTE~\cite{note} and ContraTTA~\cite{contrastivetta} employ a target memory bank where a small amount of data (\eg, \textbf{\textit{16k}} image features in ContraTTA) can be only stored and used for adaptation. 

% 우리의 프레임워크는 TTA의 한 에피소드로 여길 수 있다. 하지만 좀 더 우리의 프레임워크를 TTA 시나리오로 확장하기 위해서, 우리는 전체 타겟데이터를 조금만 저장할 수 있는 상황을 가정하고, 이러한 경우에 우리 방법론이 효과가 있는지 확인해보았다. 

\paragraph{Extension to a TTA scenario.} 
Our setup illustrated in \figrefm{fig:custom_setup} also assumes a source-free setup, so it can be easily extended to a TTA scenario by employing the memory bank. 
In particular, on a periodic basis, when 10\% of the target training data is encountered, an adaptation is executed following a TTA setup of TTT++~\cite{ttt++} and ContraTTA~\cite{contrastivetta}, where unlabeled data in the memory bank and labeled data are utilized. 
The memory bank size is set to \textbf{\textit{5k}} pseudo labels, and FreeMatch~\cite{freematch} is used for a SemiSDA baseline algorithm. 
It should be noted that since previous TTA works do not consider the utilization of labeled data, we can not use them as a baseline or compare the adaptation performance directly (but we attempt to alleviate this problem and implement comparisons in \secrefs{sec:addabl}.). 
The results in \tabrefs{tab:smalltarget} show that our method works well even with a smaller amount of unlabeled data in the memory bank. 
We find this result very surprising and wish to continue in this direction for future research.

% An adaptation was executed each time 10\% of the overall target data was gathered; in this adaptation, only unlabeled data was utilized in the memory and feedback banks. 

 % We evaluate the baseline and our approach, assuming that ML products have a smaller quantity of unlabeled target data.
% In order to simulate this scenario, we select only a small percentage of the target data and use them to perform model adaptation.
% \tabrefs{tab:smalltarget} shows the results.
% We observe that our approach works well even with the smaller amount of unlabeled data, and using a larger unlabeled dataset (\ie, usage rate\,${\approx}$\,100\%) allows the model to achieve better performance.

\vspace{-.5em}
\input{table/ablation_smalltarget}

\subsection{Learning with User Feedback}\label{sec:learnfeedback} 
Learning with User Feedback has garnered significant attention for its effectiveness in capturing users' preferences or intentions~\cite{wirth2017survey, stiennon2020learning, macglashan2017interactive, ouyang2022training}. 
Reinforcement learning from human feedback is a powerful technique for model optimization based on human-provided rewards~\cite{knox2008tamer, warnell2018deep, schulman2022chatgpt, touvron2023llama}. Another application is interactive image segmentation~\cite{sofiiuk2022reviving, sofiiuk2020f, chen2022focalclick}, where users provide pixel-level annotations, enabling the model to enhance its understanding of user preferences over time.

\section{Further understanding with Simulation Study}\label{sec:addsimulation}
In this section, we provide additional details and understanding about the simulation study in \figrefm{fig:toy_exper}.

\paragraph{Network architecture.} 
We build the model consisting of three fully connected layers and Relu activation functions. This model takes the point coordinate as input and returns the class label as output. Please refer to example codes found in the `sklearn.datasets.make{\tiny \_}blobs' documents~\cite{scikit-learn}.

\begin{figure}[t!]\centering
\includegraphics[width=1.\linewidth]{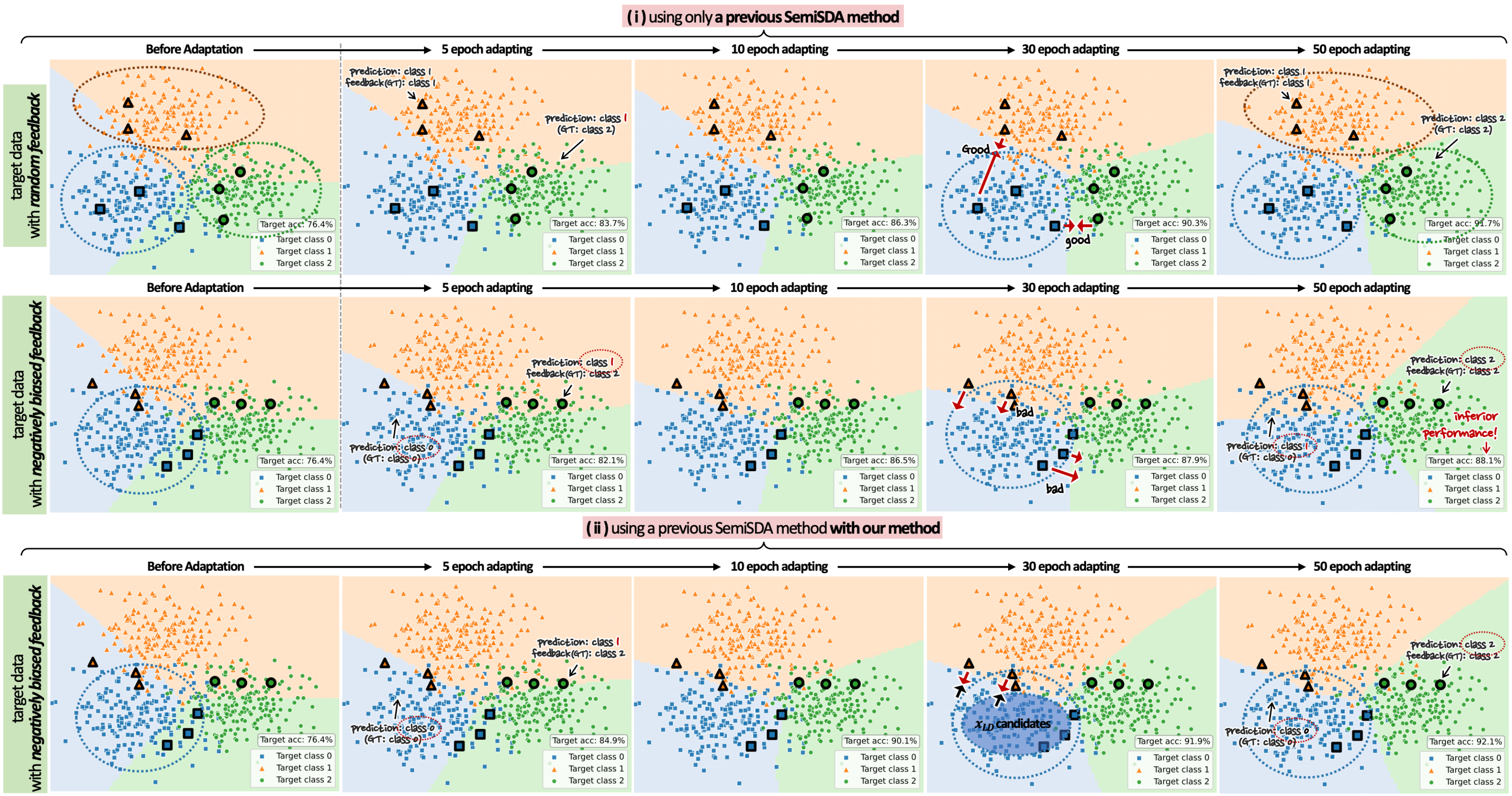}
\vspace{-2.em}
\caption{During the adapting process, an updated decision boundary of the model is depicted. The details can be found in \secrefm{sec:nbf_effect}.}
\label{fig:naviesemisda}
\end{figure}

\paragraph{Baseline.} One simple SemiSL method, Pseudo labeling~\cite{arazo2020pseudo}, can be easily applied to the toy experiment. Given a mini-batch with labeled data {\small $\{(x^{b}_{lb}, y^{b}_{lb}):b\in[1..\,B]\}$} and unlabeled data {\small $\{(x^{b}_{ulb}):b\in[1..\,\mu 
{\cdot}B]\}$}, we simply adapt the model with cross-entropy losses as the following:
\vspace{-.7em}
\begin{equation}\label{equ:pseudolabel}
    \resizebox{.9\hsize}{!}{
    $\mathcal{L}_{sup} = \frac{1}{B}\sum^{B}_{b=1} \mathcal{H}(y_{lb}^{b},f_{\theta}(x_{lb}^{b})), ~~ \mathcal{L}_{unsup} = \frac{1}{\mu \cdot B}\sum_{b=1}^{\mu \cdot B} \mathcal{H}(\argmax_{c} \left[f_{\theta}(x^{b}_{ulb})\right]_c, f_{\theta}(x_{ulb}^{b})).$}
    \vspace{-.1em}
\end{equation}
{\footnotesize $f_{\theta}(\cdot)$} is the output probability from the model and {\footnotesize $\argmax_{c} \left[f_{\theta}(x^{b}_{ulb})\right]_c$} refers to the pseudo label. 
As shown in the equation, the updating model {\footnotesize $f_{\theta}$} continuously predicts pseudo labels for the unlabeled data. So, the pseudo labels can be changed based on an updated decision boundary. \figrefs{fig:naviesemisda} presents this phenomenon as the adapting epoch progresses.

\begin{figure}[t!]\centering
% \vspace{-1.em}
\includegraphics[width=1.\linewidth]{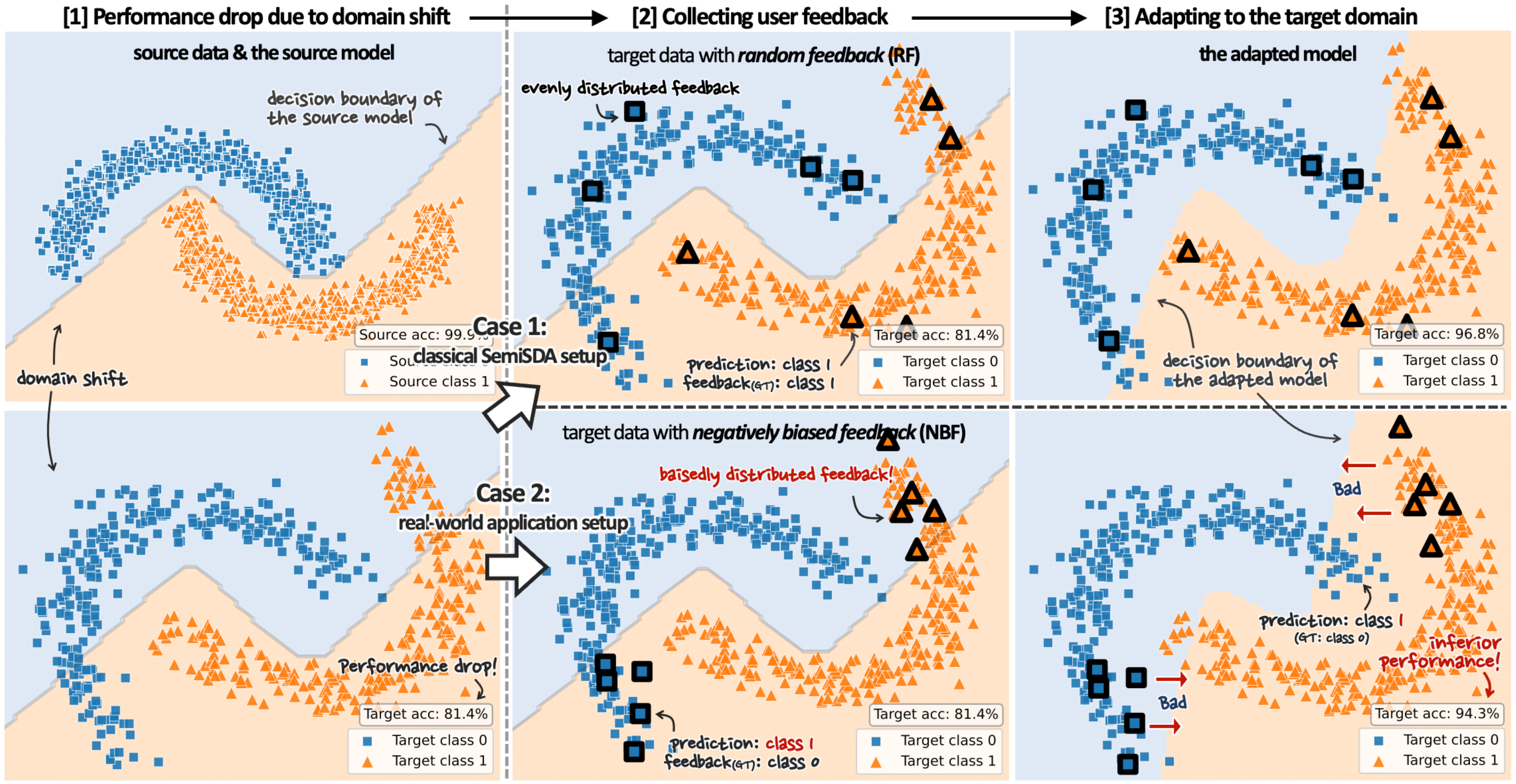}
\vspace{-2.em}
\caption{\textbf{Effect of negatively biased feedback.} We conduct an additional simulation study with two moon dataset.
We make the same observations of \figrefm{fig:toy_exper}, \ie, NBF is biasedly distributed, leading to inferior adaptation performance compared to RF. The experimental details are provided in \secrefm{sec:nbf_effect} and \secrefs{sec:addsimulation}.}
\vspace{-1.em}
\label{fig:toyexper2}
\end{figure}

\paragraph{Additional study on two moon dataset.} To better understand the unexpected influence of NBF on domain adaptation, we conducted additional simulations using the two moon datasets from scikit-learn~\cite{scikit-learn}. 
As shown in \figrefs{fig:toyexper2}, we generate source and target data so that they have domain shifts. After pre-training a model on the source data, we evaluate its performance on the target domain, observing a performance drop due to the shift (99.9\%→81.4\%). After we simulate user-provided feedback under two scenarios (\ie, RF and NBF), we adapt the model to the target data in a semi-supervised manner~\cite{arazo2020pseudo}. 
The results highlight crucial observations shown in \secrefm{sec:nbf_effect}: the distribution of label data significantly impacts adaptation performance. Notably, biased feedback distribution (NBF) leads to poorer performance compared to evenly distributed feedback (RF). 
In our main paper, we showed that this problem remained the same even with state-of-the-art SemiSDA methods and under different DA benchmarks.

\section{Additional Ablation Study}\label{sec:addabl}

\paragraph{Reliable sample filtering.}
An important design of our approach is to retain only samples having reliable pseudo labels
among {\small $\{(x^{n}_{ulb},\hat{y}^{n}_{ulb}){}:n\in[1..\,N_{ulb}]\}$}. 
We evaluate the adaptation performance with variations in the filtering ratio $p\%$ in \tabrefm{tab:selection}.
A higher $p$ increases the likelihood of the bank being contaminated with samples with incorrect pseudo labels (\ie, $y_{ulb}{\neq}\hat{y}_{ulb}$) while a lower $p$ decreases the diversity of the defending samples.
We observe that our approach is robust to the hyper-parameter $p$, yet achieves reasonable performance with $p=0.4$. 

\paragraph{Combining with SFDA methods.}\label{sec:addsfda} 
Recent SFDA methods~\cite{shot, guidingps} have shown promise in computing the unsupervised loss {\small $\mathcal{L}_{unsup}$}. So, we explore their potential as baselines within our framework. To construct the overall loss function $\mathcal{L}_{total}$ in \equrefm{equ:overallloss}, we simply combine their {\small $\mathcal{L}_{unsup}$} with the supervised loss {\small $\mathcal{L}_{sup}$} of FreeMatch~\cite{freematch} since SFDA methods do not take the utilization of supervised loss into account.
The results are presented in \tabrefs{tab:domainnet3}. 
Interestingly, some SFDA works~\cite{shot, contrastivetta,guidingps} using sophisticated methods, such as k-means clustering~\cite{kmeans} and contrastive learning~\cite{moco}, are likely to be less susceptible to NBF. However, the trend is not consistent for all methods. NRC~\cite{nrc}, using a strategy of nearest neighbors, shows sub-optimal performance under an NBF assumption. Notably, all SFDA methods achieve their best adaptation performance when combined with our method. This suggests that even methods that \textit{partially} mitigate NBF's unexpected effects can further benefit from our method.

\input{table/domainnet_semis_3}

\paragraph{Number of appended defending samples.} \label{sec:ablk} 
As mentioned in \secrefm{sec:rld}, we incorporate $k$ defending samples for each labeled data point $(x^{b}_{lb}, y^{b}_{lb})$ to decrease the unexpected impact of NBF on the supervised signal. 
To understand how the value of $k$ affects performance,  we conducted an ablation study in \tabrefs{tab:numappend}. We fix the number of labeled data points to 16 and maintain the total batch size at 128 by adjusting the ratio $\mu$ in \equrefm{equ:unsuploss}. 
For instance, with $k{=}4$, the ratio $\mu$ is set to 3 (\ie, $16+16{\times}k+16{\times}\mu=128$).
Our experiments across two different architectures reveal that a $k{=}3$ value generally yields good adaptation performance.
Consequently, we adopt $k{=}3$ for all experiments.

\input{table/ablation_numappend}

\input{table/domainnet_semis_4}

\paragraph{Under a zero feedback scenario.}\label{sec:zerofeedback} 
We note that, as previous SemiSDA~\cite{mme,adamatch} and SemiSL~\cite{fixmatch, freematch} works, we assume that a user provides a small amount of feedback (\ie, labeled data) during their interaction with an ML application. Nevertheless, we wondered about a broader question: how can our method be used when no feedback is received? This scenario, while beyond the scope of our work, presents an intriguing area for further exploration, so we attempt to investigate the potential impact of our method under such a scenario.
We initially opted to use SFDA baselines of \tabrefs{tab:domainnet3}, which have demonstrated potential in the absence of labeled target data, and assess their performance within an SFDA setup (\ie, only  {\small $\mathcal{L}_{unsup}$} in \tabrefs{tab:domainnet4}).
Then, pseudo-feedback is generated by randomly selecting small unlabeled data sets and their pseudo-labels from samples with high predicted probabilities.
With the pseudo-feedback and unlabeled target data, we conduct SemiSDA and report the results (\ie, the overall loss  {\small $\mathcal{L}_{total}$} in \tabrefs{tab:domainnet4}). We find that {\small i)} simulating pseudo-feedback has a minor influence on SFDA baselines, yet {\small ii)} the adaptation performance is enhanced by combining with our method. Based on these results,  we believe that even in the absence of feedback for certain classes, SemiSDA with our method can achieve good adaptation performance by leveraging the pseudo-feedback.

\section{Additional Experimental Details}\label{sec:adddetail}

\paragraph{Details for medical experiments.}\label{sec:detailsmedical} We use DenseNet-121~\cite{densenet} provided by the TorchXRayVision repository~\cite{torchxray}. This architecture consists of a shared backbone and multiple classification heads for radiographic findings. 
When given a 256x256 image as input, it generates sigmoid values for thirteen different findings. 

The majority of SemiSDA methods, such as AdaMatch~\cite{adamatch} and FreeMatch~\cite{freematch}, depend on consistency regularization, which requires image augmentation strategies, such as ColorJitter and GaussianBlur~\cite{pytorch}. Unfortunately, applying them to medical images remains challenging, as most strategies have been proposed specifically for natural images.
As a result, we employ Pseudo-labeling~\cite{arazo2020pseudo}, a fundamental SemiSL algorithm that (i) obviates the necessity for image augmentations and (ii) can be easily implemented for a multi-finding binary classification setup.
To be more specific, we substitute the cross-entropy $\mathcal{H}(\cdot, \cdot)$ in \equrefs{equ:pseudolabel} with the binary cross-entropy loss. To generate pseudo labels (\ie, presence or absence in \tabrefm{tab:mimic}\,(top)), thresholds that are pre-calculated in the source domain are used. The hyper-parameters for model updates are the following. 
\vspace{-1.em}
\begin{center}
\tablestyle{1pt}{1.05}
\resizebox{0.65\textwidth}{!}{
\begin{tabular}{y{70}x{60}x{60}x{60}x{60}}
 & batch size & learning rate & optimizer & weight decay \\
\shlinesmall
pre-training & 128 & 1e-3 & Adam & 1e-5 \\
adaptation & 128 & 1e-4 & Adam & 1e-5
\end{tabular}}\vspace{-1.em}
\end{center}

\paragraph{Details for semantic segmentation experiments.}
Our experiment leverages the GTA5~\cite{gta5} and Cityscapes~\cite{cityscapes} datasets as the source and target domains. To compute the supervised {\small $\mathcal{L}_{sup}$} and unsupervised losses {\small $\mathcal{L}_{unsup}$} in \equrefm{equ:unsuploss}, we employ baseline algorithms: IAST~\cite{iast} in LabOR~\cite{labor} and RIPU~\cite{ripu}. Following previous works~\cite{labor, ripu}, we utilize ResNet-101 as the backbone architecture and DeepLab-v2 as the segmentation model. 
Further details regarding implementation and hyper-parameter for adaptation can be found in the publicly available codebase of RIPU~\cite{ripu}. 
One of our method's key strengths is its simplicity, which makes it readily applicable to various tasks like semantic segmentation. 
To be more specific, we first identify pixel points in an image that have the top 40\% probabilities for each class.
Among them, we select three pixels (\ie, defending pixels) for each labeled pixel in order to balance the supervised signal (\ie, {\small $\mathcal{L}_{total}$} in \equrefm{equ:overallloss}) and obtain robust adaptation performance to the unexpected effect of NBF.

\section{Additional Discussion}\label{sec:adddisc}

\footnotetext[3]{\label{note3}{\scriptsize 
We evaluate SSNL using the same experimental setup in \tabrefs{tab:domainnet3}.}}

\subsection{Technique novelty}\label{sec:novel}
Compared to previous works, our approach, retrieval latent defending, distinguishes itself in how balancing is applied to solve the novel NBF problem.:
{\small (\textit{i})} 
We initially anticipated that conventional tricks using confident pseudo labels or balancing strategy, such as CReST~\cite{wei2021crest} for class-imbalance, CLUE~\cite{clue}, DiaNA~\cite{diana} for ActiveDA, GuidSP~\cite{guidingps} and SSNLL~\cite{ssnll} for noisy pseudo labels, would ameliorate the NBF issue. However, as shown in the table below\textsuperscript{\,\ref{note3}},
we found these methods to fall short due to their lack of specific targeting of the novel problem by NBF, thereby underscoring the need for our tailored approach.
{\small (\textit{ii})}
Our strategy diverges from the \textit{dataset}-level balancing approaches in \cite{wei2021crest, clue, ssnll}. Instead, we focus on enhancing the supervised signal within a \textit{minibatch} through iterative retrieval of defending samples, which helps in fortifying latent spaces against the unexpected issue by NBF as illustrated in \figrefm{fig:approach} and \tabrefm{tab:mimic}. Surprisingly, this distinct method not only effectively addresses the NBF problem but also leads to substantial improvements in adaptation performance.

\begin{table}[h]
\tablestyle{2pt}{1.15}
\vspace{-1.3em}
\resizebox{0.99\textwidth}{!}{
\begin{tabular}{x{50} ;{0.5pt/1.pt} x{70} ;{0.5pt/1.pt} x{70} x{70} ;{0.5pt/1.pt} x{70} x{70}}
% \hline
method                 & CReST{\tiny \,(CVPR21)} & CLUE{\tiny \,(ICCV21)} & DiaNA{\tiny \,(CVPR23)} & SSNLL{\tiny \,(IROS22)} & GuidSP{\tiny \,(CVPR23)}\\
\shline
reference              & \figrefs{fig:crest}      & \tabrefs{tab:activeda}       & \tabrefs{tab:activeda}        & -              & \tabrefs{tab:domainnet3}       \\
accuracy               & 92.6              & 68.6               & 68.1                & 68.9           & 69.2         \\
\gr
w/ ours                & 95.8 (\textbf{+3.2})        & 71.5 (\textbf{+2.9})        & 71.3 (\textbf{+3.2})         & 71.4 (\textbf{+2.5})    & 71.6 (\textbf{+2.4}) \\
% \hline
\end{tabular}}
\label{tab:novel} 
\vspace{-1.8em}
\end{table}

\footnotetext[4]{\label{note4}{\scriptsize We specify the database size when the real domain of the DomainNet dataset serves as the target domain.}}

\subsection{Computational overhead}\label{sec:efficiency}
Our method incurs only negligible overhead, as the only additional data$^{\dag}$ that needs to be stored are pseudo labels.
As shown in the following table\textsuperscript{\,\ref{note4}}, our method results in  an additional 0.1 MB of memory and a 3\% increase in running time compared to existing SemiSDA~\cite{adamatch, freematch} and SFDA~\cite{guidingps} methods, but these modest increases facilitate significant performance enhancements. We adhere to the standard practices of SemiSDA and SFDA, which involve storing target images in a database (DB)$^\ddag$. 

\begin{table}[h]
\tablestyle{2pt}{1.15}
\vspace{-1.3em}
\resizebox{0.99\textwidth}{!}{
\begin{tabular}{x{50} ;{0.5pt/1.pt} x{70} x{70} ;{0.5pt/1.pt} x{70} x{70} ;{0.5pt/1.pt} x{70} x{70}}
% \hline
method                 & AdaMatch{\tiny \,(ICLR22)}  & w/ ours & GuidSP{\tiny \,(CVPR23)} & w/ ours & FreeMatch{\tiny \,(ICLR23)} & w/ ours  \\
\shline
reference              & \tabrefm{tab:domainnet2}      & \baseline{\tabrefm{tab:domainnet2}}     & \tabrefs{tab:domainnet3}    & \baseline{\tabrefs{tab:domainnet3}}   & \tabrefs{tab:smalltarget}      & \baseline{\tabrefs{tab:smalltarget}}      \\
DB size$^\ddag$                & 55k\,images      & \baseline{55k\,images}     & 55k\,images     & \baseline{55k\,images}   & 5k\,images        & \baseline{5k\,images}   \\
add.\,data\bm{$^{\dag}$}              & 0\,MB            & \baseline{\textcolor{black}{0.1\,MB}}        & \textcolor{darkred}{53.8\,MB}       & \baseline{53.9\,MB}     & 0\,MB             & \baseline{\textcolor{black}{0.01\,MB}}             \\
run.\,time              & 132\,min         & \baseline{136\,min}        & 150\,min        & \baseline{155\,min}      & 14\,min          & \baseline{15\,min}               \\
accuracy               & 64.5             & \baseline{72.0 (\textbf{+7.5})}                & 70.2            & \baseline{71.8 (\textbf{+1.6})}   & 66.9              & \baseline{68.9 (\textbf{+2.0})}               \\
% \hline
\end{tabular}}
\label{tab:efficiency} 
\vspace{-1.8em}
\end{table}

\subsection{Limitations.} \label{sec:limitation}
Machine learning (ML) powered products can collect target data in various ways. Beyond unlabeled data encountered in the target environment (e.g., driving scenes from a self-driving car), feedback containing valuable target information can be collected by users.
For example, a radiologist can log misdiagnosed chest X-ray images in the medical application.
However, leveraging effectively such feedback to enhance the deployed model has yet to be well studied. So, this paper addressed this issue by proposing a framework, domain adaptation with user feedback, as illustrated in \figrefm{fig:custom_setup}. Moreover, we identified potential issues (\ie, the unexpected impact of NBF) and introduced a simple and scalable solution (\ie, retrieval latent defending).

However, a few more considerations need to be made before this framework is applied in the real world. 
(1) Current SemiSDA and SemiSL works typically conduct a single adaptation round using all target training data. 
In practice, however, periodic adaptation may be required since the model can continuously collect new data.
According to CoTTA~\cite{cotta}, EATA~\cite{eata}, and EcoTTA~\cite{ecotta}, studies to make initial TTA research~\cite{wang2020tent, ttt++, contrastivetta} more realistic, long-term adaptation can lead to catastrophic forgetting and error accumulation. They attempt to address this problem by utilizing continual learning strategies, \eg, random parameter restoration and knowledge distillation.
Repeated adaptation processes in our setup might result in similar issues, suggesting a potential connection to continual learning techniques within the SemiSDA methods. 
(2) More SemiSDA methods specializing in medical imaging still need to be developed. We employed the native SemiSDA method, Pseudo-Labeling~\cite{arazo2020pseudo}, in \tabrefm{tab:mimic}.
Developing SemiSDA methods specific to medical imaging has the potential to significantly improve adaptation performance beyond the results of \tabrefm{tab:mimic}. It is also a promising direction for future research.

\vspace{-.5em}
\section{Results of all domain shifts}\label{sec:allshift}

In addition to \tabrefm{tab:domainnet2}, \tabrefm{tab:officehome2}, and \tabrefs{tab:domainnet3}, we report the adaptation results for all domain shift scenarios in 
\tabrefs{tab:ddd1}, 
\tabrefs{tab:ddd2},  
\tabrefs{tab:ooo1}, 
\tabrefs{tab:ooo2}, 
\tabrefs{tab:ddd3}, and 
\tabrefs{tab:ddd4}.

\newpage
\pagebreak

\input{table/domainnet_full_semis_3}
\input{table/domainnet_full_semis_5}
\input{table/officehome_full_semis_3}
\input{table/officehome_full_semis_5}
\input{table/domainnet_full_sfda_3}
\input{table/domainnet_full_sfda_5}

%% file: table/activeda.tex
\begin{table}[t]
\tablestyle{2.pt}{1.15}
\newcolumntype{g}[1]{>{\columncolor[gray]{0.9}} x{#1}}
\resizebox{0.99\textwidth}{!}{
\begin{tabular}{y{70} ;{0.2pt/1.5pt} x{55} x{55} x{55} x{55} x{55}  ;{0.2pt/1.5pt} x{55} x{55}}
                       & \multicolumn{7}{c}{\cellcolor{white}\textbf{state B}}                                     \\
feed. amount           & \multicolumn{5}{c ;{0.2pt/1.5pt}}{378 (3 labeled data per class)}         & 1890        & 5040        \\
\textbf{stage C}                & RF                             & NBF         & Entropy~\cite{shen2017entropy}     & CLUE~\cite{clue} & DiaNA~\cite{diana} & CLUE~\cite{clue}        & CLUE~\cite{clue}        \\
\shlinesmall
AdaMatch [7]           & 67.6                           & 64.5        & 65.9        & 68.6         & 68.1          & 76.1        & 80.3        \\
\gr
w/ ours & 71.1\,{\tiny (+3.5)}                    & 72.0\,{\tiny (+7.5)} & 71.1\,{\tiny (+5.2)} & 71.5\,{\tiny (+2.9)}  & 71.3\,{\tiny (+3.2)}   & 78.0\,{\tiny (+1.9)} & 81.4\,{\tiny (+1.1)}
\end{tabular}}
\vspace{+.1em}
\caption{We evaluate a SemiSDA method~\cite{adamatch} and our method under diverse labeling scenarios. The scenarios include our proposed NBF and ActiveDA scenarios~\cite{clue, diana}. The difference between ActiveDA and our method is illustrated in \figrefs{fig:activeda}.}
\label{tab:activeda}
\vspace{-2.5em}
\end{table}

%% file: table/mimic_semis2.tex
\begin{table*}[t!]
\tablestyle{4.pt}{1.2}
\resizebox{\textwidth}{!}{
\begin{tabular}{x{5}  y{70} x{45} x{45} ;{0.5pt/1.5pt} x{80} ;{0.5pt/1.5pt} x{75} x{75}}
& method           & feedback  & FP : FN & average & fracture & pneumothorax \\
\shline
& Source model     &    -      &    -              & .6768  & .6642                     & .6894       \\
\shlinesmall
& Pseudo-Label.~\cite{arazo2020pseudo}      & {\scriptsize RF}     & -                 & .7325  & .7541                     & .7109       \\
\hdashline
&                & {\scriptsize NBF}     & 40 : 40             & .7173 {\tiny (-.0152)}  & .7414 {\tiny (-.0127)}       & .6931 {\tiny (-.0178)}       \\
\gr
\cellcolor{white} & with ours          & {\scriptsize NBF}    & 40 : 40             & .7334{\tiny (+.0162)}  & .7625 {\tiny (+.0211)}        & .7044 {\tiny (+.0113)}       \\
\cdashline{2-7}
&                 & {\scriptsize NBF}     & 75 : 5              & .7248 {\tiny (-.0077)}  & .7494 {\tiny (-.0047)}       & .7002 {\tiny (-.0107)}       \\
\gr
\cellcolor{white} &  with ours          & {\scriptsize NBF}     & 75 : 5              & \textbf{.7361} {\tiny (+.0113)}  & .7653 {\tiny (+.0159)}      & .7070 {\tiny (+.0068)}       \\
\cdashline{2-7}
&                 & {\scriptsize NBF}     & 5 : 75              & .7170 {\tiny (-.0155)}  & .7420 {\tiny (-.0121)}       & .6921 {\tiny (-.0188)}       \\
\gr
\multirow{-7}{*}{\cellcolor{white} \rotatebox[origin=c]{90}{{\tiny 80 feedback}}} &  with ours           & {\scriptsize NBF}     & 5 : 75              & .7315 {\tiny (+.0145)}  & .7679 {\tiny (+.0260)}     & .6951 {\tiny (+.0030)}      \\
\shlinesmall
& Pseudo-Label.~\cite{arazo2020pseudo}      & {\scriptsize RF}      & -                 & .7353  & .7565     & .7141       \\
\hdashline
 &                 & {\small {\scriptsize NBF}}      & 80 : 80             & .7162 {\tiny (-.0192)} & .7429 {\tiny (-.0136)} & ..6894 {\tiny (-.0247)} \\
\gr
\cellcolor{white} & with ours          & {\scriptsize NBF}      & 80 : 80             & .7331 {\tiny (+.0169)} & .7680 {\tiny (+.0251)} & .6983 {\tiny (+.0088)} \\
\cdashline{2-7}
&                 & {\scriptsize NBF}      & 155 : 5              & .7237 {\tiny (-.0117)} & .7559 {\tiny (-.0007)} & .6915 {\tiny (-.0227)} \\
\gr
\cellcolor{white} & with ours          & {\scriptsize NBF}      & 155 : 5              & \textbf{.7358} {\tiny (+.0121)} & .7665 {\tiny (+.0106)} & .7051 {\tiny (+.0136)} \\
\cdashline{2-7}
&                 & {\scriptsize NBF}   & 5 : 155              & .7166 {\tiny (-.0188)} & .7438 {\tiny (-.0128)} & .6894 {\tiny (-.0248)} \\
\gr
\multirow{-7}{*}{\cellcolor{white} \rotatebox[origin=c]{90}{{\tiny 160 feedback}}} & with ours           & {\scriptsize NBF}      & 5 : 155              & .7300 {\tiny (+.0134)} & .7696 {\tiny (+.0258)} & .6904 {\tiny (+.0010)} \\
\shlinesmall
& Fully supervised &     -    &     -              & .7744 & .8003 & .7486
\end{tabular}}
\vspace{+.1em}
\caption{\textbf{Adaptation with different feedback configurations on MIMIC-CXR-V2.} 
We conduct additional experiments to \tabrefm{tab:mimic}, where 
the same pre-trained model is utilized, and the two radiographic findings are considered for simplification.
We compare different {\scriptsize NBF} configurations as we vary the amount of feedback from false positives (FP) and false negatives (FN) errors.}
\vspace{-3.em}
\label{tab:mimic2}
\end{table*}

% Adaptation with different feedback configurations

%% file: table/ablation_smalltarget.tex
\begin{table}[t]
\tablestyle{1.8pt}{1.15}
\newcolumntype{g}[1]{>{\columncolor[gray]{0.9}} x{#1}}
\resizebox{0.8\textwidth}{!}{
\begin{tabular}{y{70} x{30} x{30} ;{0.2pt/1.5pt} x{50} x{50} x{50} x{50}}
\multicolumn{2}{l}{memory bank size = 5k}  &  & \multicolumn{4}{c}{percentage of target data encountered in target domain} \\
method & feed. & amo. &  \multicolumn{4}{c}{ 10\%\, $\xrightarrow{\hspace*{3em}}$ \, 40\% $\xrightarrow{\hspace*{3em}}$  \, 70\% $\xrightarrow{\hspace*{3em}}$  100\%\,} \\
% 2.8k & 11k & 20k & 28k \\
\shline
FreeMatch~\cite{freematch} &    {\scriptsize RF} & & 68.4        & 71.4        & 73.0        & 73.4  \\
          &     {\scriptsize NBF}                & & 66.9        & 69.5        & 71.0        & 71.5        \\
\gr
w/\,ours    & {\scriptsize NBF} & \multirow{-3}{*}{\cellcolor{white}368} & 68.9\,{\tiny(+2.0)} & 72.4\,{\tiny(+2.9)} & 73.6\,{\tiny(+2.6)} & 74.3\,{\tiny(+2.8)} \\
FreeMatch~\cite{freematch} &     {\scriptsize RF} & & 71.2        & 73.7        & 74.7        & 75.4        \\
          &    {\scriptsize NBF}                  & & 69.8        & 72.2        & 73.2        & 73.9        \\
\gr
w/\,ours    & {\scriptsize NBF} & \multirow{-3}{*}{\cellcolor{white}630} & 71.5\,{\tiny(+1.7)} & 74.1\,{\tiny(+1.9)} & 75.0\,{\tiny(+1.8)} & 75.5\,{\tiny(+1.6)}
\end{tabular}}\vspace{+.1em}
\caption{We evaluate our approach on a TTA scenario, where labeled data and unlabeled target data in a memory bank are only available for adaptation like ContraTTA~\cite{contrastivetta}. In the real, painting, scratch, and clipart domains of DomainNet-126, 10\% of the data consists of 5.5k, 2.4K, 1.9k, and 1.5k images. In the table, 40\% means that the model has encountered 40\% of the unlabeled target training data. } 
\vspace{-3.em}
\label{tab:smalltarget} 
\end{table}
% \multicolumn{1}{c}{}

% \cellcolor{white}

%% file: table/domainnet_semis_3.tex
\begin{table}[t]
\tablestyle{2pt}{1.2}
\newcolumntype{g}{>{\columncolor[gray]{0.9}} x{45}}
\resizebox{0.75\textwidth}{!}{
\begin{tabular}{x{6} y{50} ;{0.2pt/1.5pt} x{30} x{45} g ;{0.2pt/1.5pt} x{30} x{45} g}
&\multicolumn{1}{l}{feed.\,amount} & \multicolumn{3}{c}{378~{\tiny (3\,labeled data\,per\,class)}}    & \multicolumn{3}{c}{630~{\tiny (5\,labeled data\,per\,class)}}                                      \\
&method & {\scriptsize RF} & {\scriptsize NBF} & w/\,ours & {\scriptsize RF} & {\scriptsize NBF} & w/\,ours  \\
\shline
&SHOT~\cite{shot}&69.6&70.7\,{\tiny(+1.1)}&{71.5\,{\tiny(+0.8)}}&71.1&72.3\,{\tiny(+1.2)}&{73.0\,{\tiny(+0.7)}}\\
&NRC~\cite{nrc}&66.3&64.9\,{\tiny(-1.4)}&{69.3\,{\tiny(+4.4)}}&68.5&66.4\,{\tiny(-2.1)}&{69.6\,{\tiny(+3.2)}}\\
&ContraTTA~\cite{contrastivetta}&68.6&69.2\,{\tiny(+0.6)}&{71.6\,{\tiny(+2.4)}}&70.1&70.5\,{\tiny(+0.4)}&{72.4\,{\tiny(+1.9)}}\\
\multirow{-4}{*}{\rotatebox[origin=c]{90}{ResNet-50}}&GuidingSP~\cite{guidingps}&69.7&70.2\,{\tiny(+0.5)}&{\textbf{71.8}\,{\tiny(+1.6)}}&70.5&71.0\,{\tiny(+0.5)}&{\textbf{72.8}\,{\tiny(+1.8)}}\\
\shlinesmall
&SHOT~\cite{shot}&73.4&73.7\,{\tiny(+0.3)}&{74.1\,{\tiny(+0.4)}}&74.4&74.8\,{\tiny(+0.4)}&{75.4\,{\tiny(+0.6)}}\\
&NRC~\cite{nrc}&72.2&71.9\,{\tiny(-0.3)}&{72.9\,{\tiny(+1.0)}}&73.9&73.7\,{\tiny(-0.2)}&{74.6\,{\tiny(+0.9)}}\\
&ContraTTA~\cite{contrastivetta}&72.8&73.4\,{\tiny(+0.6)}&{74.9\,{\tiny(+1.5)}}&73.9&74.8\,{\tiny(+0.9)}&{76.4\,{\tiny(+1.6)}}\\
\multirow{-4}{*}{\rotatebox[origin=c]{90}{ViT-S}}&GuidingSP~\cite{guidingps}&73.3&73.7\,{\tiny(+0.4)}&{\textbf{75.0}\,{\tiny(+1.3)}}&74.1&74.9\,{\tiny(+0.8)}&{\textbf{76.4}\,{\tiny(+1.5)}}\\
% \hline
\end{tabular}}
\vspace{+.1em}
\caption{\textbf{Comparisons on DomainNet-126.} We combine our method and \textit{SFDA} methods. The average accuracy~(\%) of seven domain-shift scenarios is reported. We use the same pre-trained model as in \tabrefm{tab:domainnet2}.}
\label{tab:domainnet3}
\end{table}

%% file: table/ablation_numappend.tex
% {\scriptsize $\times$} $\circ$
\begin{table}[t]
% \vspace{-2.em}
\tablestyle{1.5pt}{1.15}
\newcolumntype{g}[1]{>{\columncolor[gray]{0.9}} x{#1}}
\resizebox{.65\textwidth}{!}{
\begin{tabular}{y{60}x{20}x{35}x{35}x{35}x{35} ;{0.2pt/1.5pt} x{35}}
  &   & $k=1$ & $k=2$ & \baseline{$k=3$} & $k=4$  & baseline\\
\shline
FreeMatch~\cite{freematch}  & Res. & 74.0 & 74.6  & \baseline{\textbf{74.8}}  &  74.4 & 72.0 \\
FreeMatch~\cite{freematch}  & ViT. & 75.5 & \textbf{75.9}  & \baseline{75.7}  &  75.4 & 73.9 \\
\end{tabular}}
\vspace{+.1em}
\caption{We ablate \textbf{the number of defending samples $k$} in \equrefm{equ:overallloss}. We also report the performance of the baseline without our approach (rightmost). }
\vspace{-2.5em}
\label{tab:numappend} 
\end{table}

%% file: table/domainnet_semis_4.tex
% {\scriptsize $\times$} $\circ$
\begin{table}[t]
\tablestyle{1.5pt}{1.15}
\newcolumntype{g}{>{\columncolor[gray]{0.9}} x{50}}
\resizebox{0.9\textwidth}{!}{
\begin{tabular}{y{120} ;{1pt/1.pt} x{78} ;{1pt/1.pt} x{40} g ;{1pt/1.pt} x{40} g }
% selection strategy & & random & \baseline{random} & kmeans & cosine & baseline \\
% class-aware &  & {\tiny \xmark} & \baseline{{\tiny \cmark}} & {\tiny \cmark} & {\tiny \cmark} & -\\
% \shlinesmall
% FreeMatch~\cite{freematch} & Res. & 73.9 & \baseline{\textbf{74.8}} &74.6 &  74.2&72.0 \\
% % \multirow{-2}{*}{\rotatebox[origin=c]{90}{Res.}}&AdaMatch~\cite{adamatch} & 69.1 & \baseline{72.0} & 71.6 & 71.3 & 64.5\\
% FreeMatch~\cite{freematch} & ViT. & 74.9 & \baseline{\textbf{75.7}} &75.6 &  75.3&73.9 \\
% % \multirow{-2}{*}{\rotatebox[origin=c]{90}{ViT.}}&AdaMatch~\cite{adamatch} & 74.3  & \baseline{75.9} & 75.7& 75.2& 73.7\\
% \\[-.5em]
 & only $\mathcal{L}_{unsup}$ & \multicolumn{4}{c}{\cellcolor{white} the overall loss $\mathcal{L}_{total}$ in \equrefm{2}}  \\
\textbf{pseudo-feedback} per class & 0  & 3 & w/\,ours & 5 & w/\,ours  \\
\shlinesmall
NRC~\cite{nrc}                             & 63.5 & 63.4  &       64.6\,{\tiny(+1.2)}  & 63.4  & 64.4\,{\tiny(+1.0)}               \\
ContrastiveTTA~\cite{contrastivetta}       & 66.6 & 66.6  &      67.4\,{\tiny(+0.8)}            & 66.5  & 67.2\,{\tiny(+0.7)}              \\
\end{tabular}}
\caption{
Although out of our scope, we consider \textbf{a zero-feedback scenario} in which a user does not provide any feedback. To evaluate our method in this scenario, we leverage unlabeled target data and their pseudo label for semi-supervised adaptation.}
\label{tab:domainnet4} 
\vspace{-2.5em}
\end{table}

% Our setup assumes that a user provides small amount of feedback while interacting with an ML application like previous SemiSDA~\cite{mme,adamatch} and SemiSL~\cite{fixmatch, freematch} works. Although out of our scope, 

%% file: table/domainnet_full_semis_3.tex
\begin{table*}[t]
\tablestyle{3.5pt}{1.2}
\resizebox{.99\textwidth}{!}{
\begin{tabular}{x{5} y{50} x{35} x{50} ;{0.5pt/1.pt} x{35} x{35} x{35} x{35} x{35} x{35} x{35}}  
& method    & feedback & average & real{\scriptsize →}clip. & real{\scriptsize →}pain. & pain.{\scriptsize →}clip. & clip.{\scriptsize →}scat. & scat.{\scriptsize →}pain. & real{\scriptsize →}scat. & pain.{\scriptsize →}real \\
\shline

&Source model &                    -               & 56.5 & 56.1 & 63.7 & 55.2 & 48.0 & 51.7 & 45.8 & 74.7 \\
\cdashline{2-11}
&FixMatch~\cite{fixmatch}  & {\scriptsize RF}      & 67.6 & 66.2 & 68.3 & 68.2 & 61.0 & 69.8 & 58.7 & 80.8 \\
&          &                  {\scriptsize NBF}    & 63.4 & 62.4 & 65.1 & 64.8 & 55.8 & 64.6 & 52.7 & 78.4 \\
\grgr
\cellcolor{white}&w/\,ours &   {\scriptsize NBF}   & 73.2 & 75.0 & 74.3 & 74.7 & 66.9 & 71.8 & 65.4 & 84.1 \\
&UDA~\cite{uda}  &             {\scriptsize RF}    & 69.2 & 68.7 & 70.0 & 69.8 & 62.8 & 70.9 & 60.0 & 82.0 \\
&          &                  {\scriptsize NBF}    & 64.9 & 64.5 & 66.0 & 67.3 & 57.2 & 66.3 & 53.8 & 79.5 \\
\grgr
\cellcolor{white}&w/\,ours &  {\scriptsize NBF}    & 73.4 & 76.2 & 74.0 & 74.7 & 67.4 & 71.9 & 65.7 & 84.1 \\
&FlexMatch~\cite{flexmatch}  & {\scriptsize RF}    & 73.3 & 76.7 & 74.0 & 75.6 & 66.9 & 73.2 & 64.4 & 82.5 \\
&          &                  {\scriptsize NBF}    & 71.4 & 74.8 & 72.2 & 74.5 & 63.8 & 71.1 & 61.7 & 81.4 \\
\grgr
\cellcolor{white}&w/\,ours &    {\scriptsize NBF}  & 74.7 & 77.9 & 74.8 & 77.8 & 68.9 & 72.2 & 66.9 & 84.4 \\
&FreeMatch~\cite{freematch}  & {\scriptsize RF}    & 73.8 & 76.6 & 74.2 & 75.5 & 67.7 & 73.5 & 65.1 & 84.0  \\
&          &                  {\scriptsize NBF}    & 72.0 & 75.5 & 72.9 & 74.6 & 65.0 & 72.3 & 62.0 & 81.7  \\
\grgr
\cellcolor{white}&w/\,ours &   {\scriptsize NBF}   & 74.8 & 78.1 & 74.5 & 77.1 & 68.8 & 72.4 & 67.3 & 85.0  \\
\cdashline{2-11}
&MME~\cite{mme}  &             {\scriptsize RF}    & 69.5 & 70.0 & 71.2 & 69.3 & 63.5 & 69.6 & 61.7 & 81.5 \\
&          &                  {\scriptsize NBF}    & 68.4 & 69.5 & 70.7 & 69.1 & 61.5 & 69.0 & 58.8 & 80.2 \\
\grgr
\cellcolor{white}&w/\,ours &   {\scriptsize NBF}   & 70.8 & 72.9 & 71.6 & 72.9 & 64.0 & 68.4 & 62.1 & 83.5 \\
&CDAC~\cite{cdac}  &           {\scriptsize RF}    & 68.3 & 67.1 & 69.0 & 68.9 & 62.6 & 69.9 & 59.5 & 81.1 \\
&          &                  {\scriptsize NBF}    & 64.6 & 64.5 & 66.2 & 66.3 & 56.9 & 65.8 & 53.6 & 78.6 \\
\grgr
\cellcolor{white}&w/\,ours &   {\scriptsize NBF}   & 73.2 & 76.1 & 73.9 & 74.4 & 67.0 & 71.2 & 65.8 & 84.1 \\
&AdaMatch~\cite{adamatch}  &   {\scriptsize RF}    & 67.6 & 66.6 & 68.5 & 68.5 & 60.3 & 69.2 & 58.7 & 81.5 \\
&          &                  {\scriptsize NBF}    & 64.5 & 64.3 & 66.1 & 65.6 & 56.9 & 65.6 & 54.2 & 78.9 \\
\grgr
\multirow{-21}{*}{\cellcolor{white}\rotatebox[origin=c]{90}{ResNet-50~\cite{resnet}}} 
&w/\,ours &                    {\scriptsize NBF}   & 72.0 & 74.5 & 72.7 & 73.9 & 65.5 & 70.0 & 64.3 & 83.2 \\
\cdashline{2-11}
&Fully sup. &                    -                 & 83.6 & 85.6 & 81.4 & 85.6 & 80.4 & 81.4 & 80.4 & 90.1 \\

\shlinesmall

&Source model &                    -               & 64.5 & 63.6 & 70.2 & 61.6 & 56.7 & 65.5 & 53.5 & 80.5 \\
\cdashline{2-11}
&FixMatch~\cite{fixmatch}  & {\scriptsize RF}      & 74.6 & 75.5 & 77.1 & 73.8 & 67.7 & 75.9 & 67.1 & 85.1 \\
&          &                  {\scriptsize NBF}    & 73.0 & 73.8 & 75.4 & 74.0 & 65.1 & 72.8 & 66.1 & 83.8 \\
\grgr
\cellcolor{white}&w/\,ours &   {\scriptsize NBF}   & 75.6 & 77.1 & 77.7 & 77.3 & 67.8 & 76.8 & 68.0 & 84.7 \\
&UDA~\cite{uda}  &             {\scriptsize RF}    & 74.8 & 75.5 & 77.1 & 74.0 & 67.9 & 76.1 & 67.4 & 85.4 \\
&          &                  {\scriptsize NBF}    & 73.3 & 74.1 & 75.6 & 74.3 & 65.4 & 73.2 & 66.3 & 83.9 \\
\grgr
\cellcolor{white}&w/\,ours &  {\scriptsize NBF}    & 75.8 & 77.1 & 77.8 & 77.6 & 68.2 & 77.1 & 68.2 & 84.9 \\
&FlexMatch~\cite{flexmatch}  & {\scriptsize RF}    & 74.9 & 75.5 & 77.0 & 74.7 & 68.4 & 76.2 & 66.7 & 85.7 \\
&          &                  {\scriptsize NBF}    & 73.9 & 74.5 & 76.6 & 75.1 & 66.1 & 74.5 & 66.4 & 84.1 \\
\grgr
\cellcolor{white}&w/\,ours &    {\scriptsize NBF}  & 75.8 & 77.2 & 77.5 & 77.9 & 68.3 & 77.0 & 67.9 & 85.0 \\
&FreeMatch~\cite{freematch}  & {\scriptsize RF}    & 74.9 & 75.3 & 76.8 & 74.5 & 68.1 & 76.5 & 67.0 & 86.0 \\
&          &                  {\scriptsize NBF}    & 73.9 & 74.6 & 76.4 & 75.0 & 66.0 & 74.5 & 66.5 & 84.1 \\
\grgr
\cellcolor{white}&w/\,ours &   {\scriptsize NBF}   & 75.7 & 76.9 & 77.5 & 77.9 & 68.1 & 76.7 & 67.8 & 85.2 \\
\cdashline{2-11}
&MME~\cite{mme}  &             {\scriptsize RF}    & 73.2 & 74.0 & 74.8 & 73.0 & 66.5 & 74.6 & 65.2 & 84.3 \\
&          &                  {\scriptsize NBF}    & 72.7 & 73.2 & 74.8 & 73.8 & 65.3 & 73.0 & 64.8 & 83.8 \\
\grgr
\cellcolor{white}&w/\,ours &   {\scriptsize NBF}   & 74.1 & 75.4 & 75.9 & 76.2 & 66.2 & 74.7 & 66.4 & 84.2 \\
&CDAC~\cite{cdac}  &           {\scriptsize RF}    & 74.2 & 74.8 & 76.3 & 73.8 & 67.5 & 75.5 & 66.6 & 84.9 \\
&          &                  {\scriptsize NBF}    & 72.8 & 73.6 & 74.9 & 73.9 & 65.0 & 72.8 & 65.4 & 83.8 \\
\grgr
\cellcolor{white}&w/\,ours &   {\scriptsize NBF}   & 75.4 & 76.7 & 77.6 & 77.2 & 67.6 & 76.2 & 67.9 & 84.6 \\
&AdaMatch~\cite{adamatch}  &   {\scriptsize RF}    & 74.7 & 75.3 & 76.9 & 73.8 & 68.0 & 76.3 & 67.1 & 85.5 \\
&          &                  {\scriptsize NBF}    & 73.7 & 74.7 & 76.2 & 74.7 & 65.7 & 74.0 & 66.8 & 84.0 \\
\grgr
\multirow{-21}{*}{\cellcolor{white}\rotatebox[origin=c]{90}{ViT-S~\cite{vit}}} 
&w/\,ours &                    {\scriptsize NBF}   & 75.9 & 76.9 & 77.8 & 77.8 & 68.5 & 76.6 & 68.3 & 85.1 \\
\cdashline{2-11}
&Fully sup. &                    -                 & 85.4 & 87.8 & 83.4 & 87.8 & 81.3 & 83.4 & 81.3 & 92.7 \\

\end{tabular}}
\vspace{-.1em}
\caption{\textbf{Adaptation results with SemiSL and SemiSDA methods on DomainNet-126.} The adaptation performance on various domain shifts is reported, where the number of labeled data per class is 3. The details can be found in \tabrefm{1}.}
\label{tab:ddd1} 
\end{table*}
% in seven domain-shift scenarios 

% ViT-S~\cite{vit}

%% file: table/domainnet_full_semis_5.tex
\begin{table*}[t]
\tablestyle{3.5pt}{1.2}
\resizebox{0.99\textwidth}{!}{
\begin{tabular}{x{5} y{50} x{35} x{50} ;{0.5pt/1.pt} x{35} x{35} x{35} x{35} x{35} x{35} x{35}} 
& method    & feedback & average & real{\scriptsize →}clip. & real{\scriptsize →}pain. & pain.{\scriptsize →}clip. & clip.{\scriptsize →}scat. & scat.{\scriptsize →}pain. & real{\scriptsize →}scat. & pain.{\scriptsize →}real \\
\shline

&Source model &                    -               & 56.5 & 56.1 & 63.7 & 55.2 & 48.0 & 51.7 & 45.8 & 74.7 \\
\cdashline{2-11}
&FixMatch~\cite{fixmatch}  & {\scriptsize RF}      & 71.5 & 71.3 & 70.9 & 73.1 & 65.5 & 71.9 & 65.1 & 83.0 \\
&          &                  {\scriptsize NBF}    & 66.1 & 66.2 & 67.6 & 67.6 & 57.4 & 67.4 & 56.5 & 79.8 \\
\grgr
\cellcolor{white}&w/\,ours &   {\scriptsize NBF}   & 75.1 & 77.2 & 75.7 & 77.2 & 69.8 & 73.9 & 68.0 & 84.1 \\
&UDA~\cite{uda}  &             {\scriptsize RF}    & 72.9 & 73.4 & 72.6 & 74.6 & 67.1 & 73.1 & 65.9 & 83.4 \\
&          &                  {\scriptsize NBF}    & 68.8 & 70.3 & 68.7 & 71.1 & 60.3 & 70.3 & 60.5 & 80.5 \\
\grgr
\cellcolor{white}&w/\,ours &  {\scriptsize NBF}    & 75.3 & 78.3 & 75.2 & 77.9 & 69.6 & 73.9 & 68.1 & 84.3 \\
&FlexMatch~\cite{flexmatch}  & {\scriptsize RF}    & 75.3 & 78.5 & 74.6 & 77.5 & 70.3 & 73.8 & 68.7 & 83.8 \\
&          &                  {\scriptsize NBF}    & 73.9 & 77.3 & 74.0 & 76.3 & 66.2 & 73.8 & 67.2 & 82.6 \\
\grgr
\cellcolor{white}&w/\,ours &    {\scriptsize NBF}  & 76.0 & 79.5 & 75.6 & 78.7 & 70.2 & 74.3 & 69.0 & 84.7 \\
&FreeMatch~\cite{freematch}  & {\scriptsize RF}    & 75.6 & 78.6 & 74.9 & 77.6 & 70.2 & 74.3 & 69.0 & 84.7 \\
&          &                  {\scriptsize NBF}    & 74.4 & 77.6 & 74.5 & 76.3 & 66.8 & 73.8 & 68.2 & 83.5 \\
\grgr
\cellcolor{white}&w/\,ours &   {\scriptsize NBF}   & 76.1 & 79.6 & 75.5 & 78.6 & 70.4 & 74.5 & 69.3 & 84.9 \\
\cdashline{2-11}
&MME~\cite{mme}  &             {\scriptsize RF}    & 71.2 & 71.3 & 72.1 & 71.8 & 65.6 & 70.7 & 64.6 & 82.6 \\
&          &                  {\scriptsize NBF}    & 70.1 & 71.4 & 71.4 & 70.4 & 62.1 & 70.7 & 62.7 & 81.8 \\
\grgr
\cellcolor{white}&w/\,ours &   {\scriptsize NBF}   & 72.5 & 74.5 & 72.7 & 74.9 & 66.4 & 70.7 & 64.6 & 83.8 \\
&CDAC~\cite{cdac}  &           {\scriptsize RF}    & 71.7 & 71.5 & 71.7 & 73.0 & 66.1 & 72.0 & 64.8 & 82.9 \\
&          &                  {\scriptsize NBF}    & 68.1 & 69.5 & 68.9 & 69.3 & 59.8 & 69.4 & 59.7 & 80.0 \\
\grgr
\cellcolor{white}&w/\,ours &   {\scriptsize NBF}   & 74.9 & 77.0 & 74.9 & 77.0 & 69.6 & 73.4 & 67.9 & 84.2 \\
&AdaMatch~\cite{adamatch}  &   {\scriptsize RF}    & 70.9 & 70.6 & 70.4 & 72.7 & 65.3 & 70.8 & 63.7 & 83.0 \\
&          &                  {\scriptsize NBF}    & 67.7 & 69.0 & 68.7 & 69.7 & 59.5 & 67.6 & 58.8 & 80.4 \\
\grgr
\multirow{-21}{*}{\cellcolor{white}\rotatebox[origin=c]{90}{ResNet-50~\cite{resnet}}} 
&w/\,ours &                    {\scriptsize NBF}   & 74.3 & 76.7 & 74.4 & 76.8 & 68.8 & 72.8 & 66.2 & 84.1 \\
\cdashline{2-11}
&Fully sup. &                    -                 & 83.6 & 85.6 & 81.4 & 85.6 & 80.4 & 81.4 & 80.4 & 90.1 \\

\shlinesmall

&Source model &                    -               & 64.5 & 63.6 & 70.2 & 61.6 & 56.7 & 65.5 & 53.5 & 80.5 \\
\cdashline{2-11}
&FixMatch~\cite{fixmatch}  & {\scriptsize RF}      & 75.7 & 76.5 & 77.4 & 76.2 & 69.6 & 76.9 & 67.9 & 85.8 \\
&          &                  {\scriptsize NBF}    & 74.3 & 75.7 & 75.6 & 75.6 & 67.4 & 74.2 & 66.7 & 84.7 \\
\grgr
\cellcolor{white}&w/\,ours &   {\scriptsize NBF}   & 76.5 & 78.0 & 77.9 & 78.3 & 70.2 & 76.9 & 68.7 & 85.4 \\
&UDA~\cite{uda}  &             {\scriptsize RF}    & 75.9 & 76.7 & 77.4 & 76.4 & 69.8 & 76.9 & 68.1 & 85.9 \\
&          &                  {\scriptsize NBF}    & 74.5 & 75.9 & 76.0 & 76.0 & 67.6 & 74.4 & 67.0 & 84.9 \\
\grgr
\cellcolor{white}&w/\,ours &  {\scriptsize NBF}    & 76.7 & 78.2 & 78.2 & 78.8 & 70.6 & 76.9 & 68.8 & 85.5 \\
&FlexMatch~\cite{flexmatch}  & {\scriptsize RF}    & 76.0 & 76.5 & 77.2 & 76.8 & 70.1 & 77.3 & 68.1 & 86.2 \\
&          &                  {\scriptsize NBF}    & 75.1 & 76.2 & 76.6 & 76.2 & 68.9 & 75.5 & 67.4 & 85.1 \\
\grgr
\cellcolor{white}&w/\,ours &    {\scriptsize NBF}  & 76.9 & 78.9 & 77.9 & 79.1 & 70.4 & 77.6 & 68.6 & 86.0 \\
&FreeMatch~\cite{freematch}  & {\scriptsize RF}    & 76.0 & 76.7 & 77.1 & 76.6 & 69.9 & 77.1 & 68.0 & 86.3 \\
&          &                  {\scriptsize NBF}    & 75.1 & 76.2 & 76.4 & 76.3 & 69.0 & 75.6 & 67.4 & 85.1 \\
\grgr
\cellcolor{white}&w/\,ours &   {\scriptsize NBF}   & 76.8 & 78.5 & 77.8 & 78.5 & 70.5 & 77.4 & 68.8 & 85.9 \\
\cdashline{2-11}
&MME~\cite{mme}  &             {\scriptsize RF}    & 74.5 & 75.3 & 75.4 & 75.2 & 68.2 & 75.4 & 66.5 & 85.1 \\
&          &                  {\scriptsize NBF}    & 74.0 & 74.9 & 75.2 & 75.2 & 67.3 & 74.1 & 66.3 & 84.7 \\
\grgr
\cellcolor{white}&w/\,ours &   {\scriptsize NBF}   & 75.2 & 76.4 & 76.4 & 77.3 & 68.9 & 75.4 & 66.9 & 85.1 \\
&CDAC~\cite{cdac}  &           {\scriptsize RF}    & 75.4 & 76.3 & 76.9 & 75.5 & 69.2 & 76.4 & 67.8 & 85.6 \\
&          &                  {\scriptsize NBF}    & 74.1 & 75.1 & 75.3 & 75.4 & 67.4 & 73.9 & 66.5 & 84.6 \\
\grgr
\cellcolor{white}&w/\,ours &   {\scriptsize NBF}   & 76.2 & 77.8 & 77.4 & 78.3 & 70.0 & 76.4 & 68.5 & 85.3 \\
&AdaMatch~\cite{adamatch}  &   {\scriptsize RF}    & 75.9 & 76.6 & 77.1 & 76.6 & 70.0 & 76.9 & 68.2 & 86.1 \\
&          &                  {\scriptsize NBF}    & 75.1 & 76.2 & 76.7 & 76.3 & 68.1 & 75.5 & 67.5 & 85.2 \\
\grgr
\multirow{-21}{*}{\cellcolor{white}\rotatebox[origin=c]{90}{ViT-S~\cite{vit}}} 
&w/\,ours &                    {\scriptsize NBF}   & 76.7 & 78.6 & 78.0 & 78.8 & 69.6 & 77.2 & 68.8 & 86.0 \\
\cdashline{2-11}
&Fully sup. &                    -                 & 85.4 & 87.8 & 83.4 & 87.8 & 81.3 & 83.4 & 81.3 & 92.7 \\

\end{tabular}}
\caption{\textbf{Adaptation results with SemiSL and SemiSDA methods on DomainNet-126.} The adaptation performance on various domain shifts is reported, where the number of labeled data per class is 5. The details can be found in \tabrefm{1}.}
\label{tab:ddd2} 
\end{table*}
% in seven domain-shift scenarios 

% ViT-S~\cite{vit}

%% file: table/officehome_full_semis_3.tex
\begin{table*}[t]
\tablestyle{4pt}{1.2}
\resizebox{\textwidth}{!}{
\begin{tabular}{y{44} x{35} x{38} ;{0.5pt/1.pt} x{18} x{18} x{18} x{18} x{18} x{18} x{18} x{18} x{18} x{18} x{18} x{18} x{18}} 
method    & feedback & average  & a\,→\,c & a\,→\,p & a\,→\,r & c\,→\,a & c\,→\,p & c\,→\,r & p\,→\,a & p\,→\,c & p\,→\,r & r\,→\,a & r\,→\,c & r\,→\,p \\
\shline
Source & -                               & 57.6 & 44.2 & 65.6 & 71.6 & 47.3 & 60.2 & 58.2 & 47.9 & 40.8 & 69.8 & 60.6 & 46.5 & 78.1 \\
\hdashline
FreeMatch~\cite{freematch} & {\scriptsize RF}  & 71.4 & 56.6 & 79.7 & 76.3 & 67.9 & 83.2 & 74.5 & 65.5 & 58.6 & 78.3 & 69.4 & 62.4 & 84.8 \\
                           & {\scriptsize NBF} & 68.6 & 53.0 & 76.1 & 75.3 & 65.3 & 78.5 & 74.8 & 62.5 & 56.7 & 74.7 & 66.7 & 56.2 & 83.7 \\
\grgr
w/\,ours                   & {\scriptsize NBF} & 73.7 & 60.8 & 80.3 & 80.5 & 69.2 & 84.0 & 78.6 & 67.7 & 62.3 & 80.1 & 70.0 & 64.1 & 87.2 \\
UDA~\cite{uda}             & {\scriptsize RF}  & 72.2 & 56.1 & 81.0 & 76.8 & 68.0 & 83.4 & 75.6 & 67.1 & 59.7 & 79.7 & 69.8 & 62.7 & 86.4 \\
                           & {\scriptsize NBF} & 69.5 & 53.3 & 78.6 & 75.7 & 66.3 & 79.7 & 75.8 & 63.7 & 57.2 & 75.7 & 66.7 & 57.2 & 83.9 \\
\grgr
w/\,ours                   & {\scriptsize NBF} & 74.1 & 61.1 & 80.7 & 80.3 & 69.0 & 85.9 & 79.2 & 68.0 & 62.3 & 80.7 & 70.4 & 63.9 & 87.4 \\
FlexMatch~\cite{flexmatch} & {\scriptsize RF}  & 73.7 & 58.0 & 84.6 & 79.3 & 68.4 & 84.7 & 78.8 & 68.4 & 62.8 & 79.8 & 70.6 & 62.9 & 86.3 \\
                           & {\scriptsize NBF} & 72.1 & 56.1 & 79.0 & 77.8 & 68.4 & 83.4 & 77.6 & 67.5 & 60.1 & 79.2 & 68.8 & 60.5 & 86.2 \\
\grgr
w/\,ours                   & {\scriptsize NBF} & 74.7 & 60.8 & 81.7 & 81.1 & 70.0 & 85.8 & 79.8 & 68.8 & 61.4 & 81.4 & 70.2 & 65.7 & 89.4 \\
FreeMatch~\cite{freematch} & {\scriptsize RF}  & 74.0 & 58.5 & 85.0 & 79.4 & 68.2 & 84.7 & 79.2 & 68.4 & 62.5 & 80.4 & 71.0 & 63.7 & 87.0 \\
                           & {\scriptsize NBF} & 72.2 & 56.4 & 79.3 & 77.7 & 67.7 & 83.4 & 78.5 & 67.3 & 60.5 & 79.1 & 69.2 & 61.0 & 86.9 \\
\grgr
w/\,ours                   & {\scriptsize NBF} & 74.8 & 60.6 & 81.4 & 81.5 & 70.8 & 86.7 & 80.0 & 68.6 & 61.6 & 81.7 & 69.8 & 66.2 & 89.2 \\
\hdashline
MME~\cite{mme}             & {\scriptsize RF}  & 71.2 & 56.2 & 80.4 & 75.7 & 65.1 & 81.0 & 76.7 & 64.5 & 59.0 & 79.8 & 69.0 & 62.0 & 85.1 \\
                           & {\scriptsize NBF} & 70.2 & 55.0 & 77.6 & 76.8 & 65.1 & 82.2 & 77.7 & 61.1 & 57.1 & 77.1 & 68.8 & 58.1 & 85.4 \\
\grgr
w/\,ours                   & {\scriptsize NBF} & 73.4 & 60.5 & 81.4 & 80.0 & 68.6 & 84.8 & 78.4 & 65.3 & 61.3 & 79.8 & 69.8 & 62.8 & 87.5 \\
CDAC~\cite{cdac}           & {\scriptsize RF}  & 71.2 & 55.5 & 80.0 & 76.4 & 67.1 & 82.4 & 75.8 & 64.5 & 58.7 & 79.0 & 69.2 & 61.5 & 84.4 \\
                           & {\scriptsize NBF} & 69.0 & 54.1 & 76.2 & 75.4 & 64.1 & 79.5 & 75.4 & 63.9 & 57.9 & 75.2 & 66.5 & 55.8 & 83.6 \\
\grgr
w/\,ours                   & {\scriptsize NBF} & 74.3 & 63.7 & 81.3 & 80.4 & 70.0 & 85.4 & 79.0 & 67.9 & 62.2 & 80.3 & 69.6 & 65.1 & 86.9 \\
AdaMatch~\cite{adamatch}   & {\scriptsize RF}  & 70.9 & 55.4 & 80.4 & 75.9 & 65.7 & 81.5 & 74.6 & 65.9 & 58.7 & 78.4 & 68.8 & 61.5 & 84.3 \\
                           & {\scriptsize NBF} & 69.3 & 54.2 & 76.6 & 75.3 & 65.9 & 79.3 & 75.5 & 63.7 & 57.4 & 75.9 & 66.7 & 56.8 & 84.2 \\
\grgr
w/\,ours                   & {\scriptsize NBF} & 73.8 & 62.2 & 81.0 & 79.7 & 68.8 & 85.4 & 78.6 & 67.7 & 61.7 & 79.5 & 69.0 & 64.1 & 88.2 \\
\hdashline
Fully sup. & -                                 & 87.4 & 84.5 & 95.1 & 89.0 & 80.9 & 95.1 & 89.0 & 80.9 & 84.5 & 89.0 & 80.9 & 84.5 & 95.1 \\
\end{tabular}}
\caption{\textbf{Adaptation results with SemiSL and SemiSDA methods on OfficeHome.} The adaptation performance on various domain shifts is reported, where the number of labeled data per class is 3. The details can be found in \tabrefm{2}.}
\label{tab:ooo1} 
\end{table*}

%% file: table/officehome_full_semis_5.tex
\begin{table*}[t]
\tablestyle{4pt}{1.2}
\resizebox{\textwidth}{!}{
\begin{tabular}{y{44} x{35} x{38} ;{0.5pt/1.pt} x{18} x{18} x{18} x{18} x{18} x{18} x{18} x{18} x{18} x{18} x{18} x{18} x{18}} 
method    & feedback & average  & a\,→\,c & a\,→\,p & a\,→\,r & c\,→\,a & c\,→\,p & c\,→\,r & p\,→\,a & p\,→\,c & p\,→\,r & r\,→\,a & r\,→\,c & r\,→\,p \\
\shline
Source & -                               & 57.6 & 44.2 & 65.6 & 71.6 & 47.3 & 60.2 & 58.2 & 47.9 & 40.8 & 69.8 & 60.6 & 46.5 & 78.1 \\
\hdashline
FreeMatch~\cite{freematch} & {\scriptsize RF}  & 73.9 & 59.8 & 83.9 & 80.0 & 69.2 & 84.6 & 77.8 & 66.7 & 63.7 & 80.5 & 71.2 & 63.3 & 85.8 \\
                           & {\scriptsize NBF} & 72.2 & 57.6 & 82.4 & 76.3 & 68.2 & 82.0 & 76.6 & 65.3 & 61.4 & 78.0 & 71.4 & 60.9 & 86.6 \\
\grgr
w/\,ours                   & {\scriptsize NBF} & 75.3 & 63.9 & 84.6 & 79.0 & 70.0 & 85.7 & 79.1 & 68.6 & 64.8 & 81.1 & 73.0 & 65.4 & 88.6 \\
UDA~\cite{uda}             & {\scriptsize RF}  & 74.4 & 60.2 & 84.5 & 79.8 & 68.4 & 85.1 & 79.8 & 66.5 & 64.4 & 80.7 & 72.2 & 64.4 & 86.1 \\
                           & {\scriptsize NBF} & 73.0 & 58.7 & 82.6 & 77.4 & 68.6 & 82.5 & 77.3 & 66.9 & 61.9 & 78.8 & 71.4 & 62.0 & 87.4 \\
\grgr
w/\,ours                   & {\scriptsize NBF} & 76.0 & 64.9 & 84.5 & 79.4 & 71.2 & 85.8 & 79.8 & 71.4 & 65.4 & 80.5 & 74.2 & 66.3 & 88.9 \\
FlexMatch~\cite{flexmatch} & {\scriptsize RF}  & 75.9 & 64.3 & 84.9 & 82.1 & 69.6 & 85.7 & 80.4 & 69.2 & 65.7 & 82.3 & 74.2 & 65.4 & 87.3 \\
                           & {\scriptsize NBF} & 74.9 & 62.9 & 83.2 & 77.6 & 70.2 & 84.7 & 80.5 & 69.8 & 62.9 & 79.5 & 74.4 & 64.4 & 87.7 \\
\grgr
w/\,ours                   & {\scriptsize NBF} & 76.6 & 63.3 & 86.7 & 79.5 & 71.6 & 86.9 & 81.0 & 72.0 & 65.7 & 81.3 & 75.0 & 67.5 & 88.9 \\
FreeMatch~\cite{freematch} & {\scriptsize RF}  & 75.8 & 63.2 & 85.2 & 81.8 & 70.0 & 86.3 & 80.6 & 69.0 & 65.8 & 82.1 & 73.2 & 65.6 & 87.0 \\
                           & {\scriptsize NBF} & 75.0 & 63.2 & 83.6 & 77.4 & 70.0 & 84.9 & 80.5 & 70.4 & 62.6 & 79.8 & 74.6 & 63.9 & 88.9 \\
\grgr
w/\,ours                   & {\scriptsize NBF} & 76.6 & 63.4 & 85.6 & 79.8 & 71.8 & 86.3 & 81.2 & 71.8 & 65.3 & 81.8 & 74.8 & 67.5 & 89.6 \\
\hdashline
MME~\cite{mme}             & {\scriptsize RF}  & 73.5 & 59.6 & 82.4 & 78.7 & 67.3 & 83.6 & 79.2 & 67.3 & 62.4 & 80.5 & 71.4 & 63.2 & 86.6 \\
                           & {\scriptsize NBF} & 73.1 & 59.5 & 83.2 & 77.2 & 66.5 & 82.5 & 78.3 & 65.1 & 61.5 & 79.1 & 72.8 & 62.8 & 88.3 \\
\grgr
w/\,ours                   & {\scriptsize NBF} & 75.6 & 63.6 & 84.2 & 77.3 & 69.8 & 85.5 & 80.3 & 70.8 & 65.2 & 80.5 & 74.6 & 66.6 & 88.9 \\
CDAC~\cite{cdac}           & {\scriptsize RF}  & 73.5 & 59.7 & 83.4 & 79.3 & 68.6 & 84.5 & 78.1 & 66.3 & 63.4 & 80.5 & 69.8 & 63.4 & 85.1 \\
                           & {\scriptsize NBF} & 72.3 & 59.5 & 81.7 & 76.6 & 67.7 & 81.9 & 76.7 & 65.9 & 62.4 & 77.4 & 70.8 & 60.2 & 86.4 \\
\grgr
w/\,ours                   & {\scriptsize NBF} & 75.7 & 64.2 & 84.7 & 79.0 & 72.2 & 85.5 & 79.6 & 70.4 & 65.3 & 80.4 & 73.4 & 65.4 & 88.6 \\
AdaMatch~\cite{adamatch}   & {\scriptsize RF}  & 73.4 & 60.0 & 83.8 & 78.7 & 68.0 & 84.4 & 77.6 & 66.5 & 62.5 & 80.0 & 71.0 & 63.2 & 85.1 \\
                           & {\scriptsize NBF} & 72.7 & 60.2 & 81.7 & 76.9 & 67.1 & 81.5 & 77.2 & 66.3 & 61.8 & 78.7 & 71.2 & 62.0 & 87.1 \\
\grgr
w/\,ours                   & {\scriptsize NBF} & 75.5 & 63.4 & 84.4 & 78.8 & 70.0 & 86.0 & 79.4 & 70.2 & 65.3 & 80.6 & 72.8 & 66.6 & 88.4 \\
\hdashline
Fully sup. & -                                 & 87.4 & 84.5 & 95.1 & 89.0 & 80.9 & 95.1 & 89.0 & 80.9 & 84.5 & 89.0 & 80.9 & 84.5 & 95.1 \\
\end{tabular}}
\caption{\textbf{Adaptation results with SemiSL and SemiSDA methods on OfficeHome.} The adaptation performance on various domain shifts is reported, where the number of labeled data per class is 5. The details can be found in \tabrefm{2}.}
\label{tab:ooo2} 
\end{table*}

%% file: table/domainnet_full_sfda_3.tex
\begin{table*}[t]
\tablestyle{3.5pt}{1.2}
\resizebox{\textwidth}{!}{
\begin{tabular}{x{5} y{50} x{35} x{50} ;{0.5pt/1.pt} x{35} x{35} x{35} x{35} x{35} x{35} x{35}} 
& method    & feedback & average & real{\scriptsize →}clip. & real{\scriptsize →}pain. & pain.{\scriptsize →}clip. & clip.{\scriptsize →}scat. & scat.{\scriptsize →}pain. & real{\scriptsize →}scat. & pain.{\scriptsize →}real \\
\shline

&Source model &                    -                 & 56.5 & 56.1 & 63.7 & 55.2 & 48.0 & 51.7 & 45.8 & 74.7 \\
\cdashline{2-11}
&SHOT~\cite{shot}  & {\scriptsize RF}                & 69.6 & 70.2 & 70.9 & 69.6 & 63.4 & 69.1 & 61.4 & 82.8 \\
&          &                  {\scriptsize NBF}      & 70.7 & 71.7 & 72.7 & 71.0 & 64.1 & 69.7 & 62.0 & 83.6 \\
\grgr
\cellcolor{white}&w/\,ours &   {\scriptsize NBF}     & 71.5 & 73.8 & 72.8 & 73.5 & 64.6 & 69.8 & 62.6 & 83.6 \\
&NRC~\cite{nrc}  &             {\scriptsize RF}      & 66.3 & 66.1 & 69.3 & 64.8 & 58.0 & 67.9 & 57.6 & 80.6 \\
&          &                  {\scriptsize NBF}      & 64.9 & 63.1 & 68.4 & 63.6 & 56.9 & 67.1 & 55.1 & 80.4 \\
\grgr
\cellcolor{white}&w/\,ours &  {\scriptsize NBF}      & 69.3 & 70.2 & 71.4 & 69.7 & 62.1 & 68.2 & 62.0 & 81.4 \\
&ContraTTA~\cite{contrastivetta}  & {\scriptsize RF} & 68.6 & 72.3 & 70.4 & 70.7 & 60.0 & 65.1 & 61.6 & 80.1 \\
&          &                  {\scriptsize NBF}      & 69.2 & 72.8 & 70.9 & 71.1 & 60.2 & 66.5 & 62.1 & 80.7 \\
\grgr
\cellcolor{white}&w/\,ours &    {\scriptsize NBF}    & 71.6 & 74.6 & 72.1 & 75.3 & 64.1 & 69.7 & 62.7 & 82.7 \\
&GuidingSP~\cite{guidingps}  & {\scriptsize RF}      & 69.7 & 66.6 & 68.5 & 68.5 & 60.3 & 69.2 & 58.7 & 81.5 \\
&          &                  {\scriptsize NBF}      & 70.2 & 64.3 & 66.1 & 65.6 & 56.9 & 65.6 & 54.2 & 78.9 \\
\grgr
\multirow{-12}{*}{\cellcolor{white}\rotatebox[origin=c]{90}{ResNet-50~\cite{resnet}}} 
&w/\,ours &                    {\scriptsize NBF}     & 71.8 & 74.5 & 72.7 & 73.9 & 65.5 & 70.0 & 64.3 & 83.2 \\
\cdashline{2-11}
&Fully sup. &                    -                   & 83.6 & 85.6 & 81.4 & 85.6 & 80.4 & 81.4 & 80.4 & 90.1 \\

\shlinesmall

&Source model &                    -               & 64.5 & 63.6 & 70.2 & 61.6 & 56.7 & 65.5 & 53.5 & 80.5 \\
\cdashline{2-11}
&SHOT~\cite{shot}  & {\scriptsize RF}              & 73.4 & 73.9 & 74.9 & 73.2 & 66.8 & 74.8 & 65.4 & 84.7 \\
&          &                  {\scriptsize NBF}    & 73.7 & 74.6 & 75.6 & 74.2 & 67.0 & 74.4 & 65.4 & 84.6 \\
\grgr
\cellcolor{white}&w/\,ours &   {\scriptsize NBF}   & 74.1 & 75.1 & 75.7 & 74.9 & 67.6 & 74.6 & 66.0 & 84.7 \\
&NRC~\cite{nrc}  &             {\scriptsize RF}    & 72.2 & 73.0 & 73.9 & 72.3 & 65.6 & 73.6 & 63.8 & 83.0 \\
&          &                  {\scriptsize NBF}    & 71.9 & 73.1 & 73.8 & 72.1 & 65.2 & 73.0 & 64.1 & 82.3 \\
\grgr
\cellcolor{white}&w/\,ours &  {\scriptsize NBF}    & 72.9 & 73.9 & 74.9 & 73.9 & 65.5 & 73.4 & 64.5 & 84.3 \\
&ContraTTA~\cite{contrastivetta}& {\scriptsize RF} & 72.8 & 73.0 & 74.1 & 74.7 & 66.7 & 73.2 & 62.9 & 84.8 \\
&          &                  {\scriptsize NBF}    & 73.4 & 74.3 & 75.1 & 74.6 & 67.6 & 73.8 & 63.7 & 84.9 \\
\grgr
\cellcolor{white}&w/\,ours &    {\scriptsize NBF}  & 74.9 & 75.4 & 75.8 & 76.7 & 69.2 & 75.6 & 66.6 & 85.0 \\
&GuidingSP~\cite{guidingps}  & {\scriptsize RF}    & 73.3 & 73.9 & 74.5 & 75.0 & 66.9 & 73.7 & 63.4 & 85.1 \\
&          &                  {\scriptsize NBF}    & 73.7 & 74.8 & 75.5 & 74.6 & 67.8 & 73.9 & 63.9 & 85.1 \\
\grgr
\multirow{-12}{*}{\cellcolor{white}\rotatebox[origin=c]{90}{ViT-S~\cite{vit}}} 
&w/\,ours &                    {\scriptsize NBF}   & 75.0 & 75.6 & 75.8 & 76.9 & 69.1 & 75.6 & 66.5 & 85.2 \\
\cdashline{2-11}
&Fully sup. &                    -                 & 85.4 & 87.8 & 83.4 & 87.8 & 81.3 & 83.4 & 81.3 & 92.7 \\

\end{tabular}}
\caption{\textbf{Adaptation results with SFDA methods on DomainNet-126.} The adaptation performance on various domain shifts is reported, where the number of labeled data per class is 3. The details can be found in \tabrefs{tab:domainnet3}.}
\label{tab:ddd3} 
\end{table*}
% in seven domain-shift scenarios 

% ViT-S~\cite{vit}

%% file: table/domainnet_full_sfda_5.tex
\begin{table*}[t]
\tablestyle{3.5pt}{1.2}
\resizebox{\textwidth}{!}{
\begin{tabular}{x{5} y{50} x{35} x{50} ;{0.5pt/1.pt} x{35} x{35} x{35} x{35} x{35} x{35} x{35}} 
& method    & feedback & average & real{\scriptsize →}clip. & real{\scriptsize →}pain. & pain.{\scriptsize →}clip. & clip.{\scriptsize →}scat. & scat.{\scriptsize →}pain. & real{\scriptsize →}scat. & pain.{\scriptsize →}real \\
\shline

&Source model &                    -                 & 56.5 & 56.1 & 63.7 & 55.2 & 48.0 & 51.7 & 45.8 & 74.7 \\
\cdashline{2-11}
&SHOT~\cite{shot}  & {\scriptsize RF}                & 71.1 & 71.9 & 72.6 & 70.8 & 65.3 & 70.1 & 63.6 & 83.1 \\
&          &                  {\scriptsize NBF}      & 72.3 & 73.3 & 74.0 & 73.1 & 65.8 & 71.5 & 64.4 & 84.2 \\
\grgr
\cellcolor{white}&w/\,ours &   {\scriptsize NBF}     & 73.0 & 75.2 & 74.2 & 74.3 & 66.3 & 71.4 & 65.1 & 84.5 \\
&NRC~\cite{nrc}  &             {\scriptsize RF}      & 68.5 & 68.6 & 70.1 & 68.3 & 61.1 & 68.6 & 61.5 & 81.2 \\
&          &                  {\scriptsize NBF}      & 66.4 & 65.4 & 69.0 & 65.7 & 58.9 & 67.0 & 58.6 & 80.7 \\
\grgr
\cellcolor{white}&w/\,ours &  {\scriptsize NBF}      & 69.6 & 70.6 & 72.2 & 70.2 & 61.9 & 68.1 & 62.4 & 81.6 \\
&ContraTTA~\cite{contrastivetta}  & {\scriptsize RF} & 70.1 & 73.7 & 71.0 & 72.4 & 61.8 & 67.0 & 64.0 & 81.0 \\
&          &                  {\scriptsize NBF}      & 70.5 & 74.4 & 71.8 & 72.3 & 61.4 & 67.8 & 64.2 & 81.3 \\
\grgr
\cellcolor{white}&w/\,ours &    {\scriptsize NBF}    & 72.4 & 76.0 & 73.3 & 73.1 & 64.8 & 71.3 & 65.0 & 83.2 \\
&GuidingSP~\cite{guidingps}  & {\scriptsize RF}      & 70.5 & 70.9 & 70.6 & 70.4 & 72.7 & 65.3 & 70.8 & 63.7 \\
&          &                  {\scriptsize NBF}      & 71.0 & 67.7 & 69.0 & 68.7 & 69.7 & 59.5 & 67.6 & 58.8 \\
\grgr
\multirow{-12}{*}{\cellcolor{white}\rotatebox[origin=c]{90}{ResNet-50~\cite{resnet}}} 
&w/\,ours &                    {\scriptsize NBF}     & 72.8 & 74.3 & 76.7 & 74.4 & 76.8 & 68.8 & 72.8 & 66.2 \\
\cdashline{2-11}
&Fully sup. &                    -                   & 83.6 & 85.6 & 81.4 & 85.6 & 80.4 & 81.4 & 80.4 & 90.1 \\

\shlinesmall

&Source model &                    -               & 64.5 & 63.6 & 70.2 & 61.6 & 56.7 & 65.5 & 53.5 & 80.5 \\
\cdashline{2-11}
&SHOT~\cite{shot}  & {\scriptsize RF}              & 74.4 & 75.1 & 75.6 & 74.6 & 68.5 & 75.2 & 67.0 & 85.0 \\
&          &                  {\scriptsize NBF}    & 74.8 & 75.9 & 76.3 & 75.1 & 68.7 & 75.8 & 66.7 & 85.3 \\
\grgr
\cellcolor{white}&w/\,ours &   {\scriptsize NBF}   & 75.4 & 77.3 & 76.5 & 75.9 & 69.2 & 76.1 & 67.1 & 85.4 \\
&NRC~\cite{nrc}  &             {\scriptsize RF}    & 73.9 & 75.1 & 75.1 & 73.8 & 67.4 & 74.2 & 66.3 & 85.5 \\
&          &                  {\scriptsize NBF}    & 73.7 & 74.8 & 74.9 & 73.8 & 67.2 & 73.8 & 66.2 & 85.0 \\
\grgr
\cellcolor{white}&w/\,ours &  {\scriptsize NBF}    & 74.6 & 76.0 & 75.9 & 75.5 & 67.9 & 74.5 & 66.7 & 85.3 \\
&ContraTTA~\cite{contrastivetta}& {\scriptsize RF} & 73.9 & 74.3 & 74.9 & 76.2 & 68.5 & 74.1 & 64.7 & 84.9 \\
&          &                  {\scriptsize NBF}    & 74.8 & 74.9 & 75.7 & 76.2 & 69.2 & 75.3 & 66.7 & 85.5 \\
\grgr
\cellcolor{white}&w/\,ours &    {\scriptsize NBF}  & 76.4 & 77.2 & 76.4 & 79.0 & 70.9 & 76.8 & 67.8 & 86.5 \\
&GuidingSP~\cite{guidingps}  & {\scriptsize RF}    & 74.1 & 74.2 & 75.0 & 76.5 & 68.9 & 74.2 & 64.9 & 85.0 \\
&          &                  {\scriptsize NBF}    & 74.9 & 74.9 & 75.8 & 76.3 & 69.1 & 75.2 & 66.8 & 85.9 \\
\grgr
\multirow{-12}{*}{\cellcolor{white}\rotatebox[origin=c]{90}{ViT-S~\cite{vit}}} 
&w/\,ours &                    {\scriptsize NBF}   & 76.4 & 77.4 & 76.4 & 79.1 & 70.9 & 76.8 & 67.7 & 86.6 \\
\cdashline{2-11}
&Fully sup. &                    -                 & 85.4 & 87.8 & 83.4 & 87.8 & 81.3 & 83.4 & 81.3 & 92.7 \\

\end{tabular}}
\caption{\textbf{Adaptation results with SFDA methods on DomainNet-126.} The adaptation performance on various domain shifts is reported, where the number of labeled data per class is 5. The details can be found in \tabrefs{tab:domainnet3}.}
\label{tab:ddd4} 
\end{table*}
% in seven domain-shift scenarios 

% ViT-S~\cite{vit}